\documentclass[11pt,twoside]{book}
\pdfoutput=1
\usepackage[T1]{fontenc}
\usepackage[dvips]{graphicx} 
\usepackage{epsfig}
\usepackage{latexsym}
\usepackage{amssymb}
\usepackage{amsmath}
\usepackage[ansinew]{inputenc}
\usepackage{makeidx}
\usepackage{algorithmic}
\usepackage{algorithm}
\usepackage{booktabs}           
\usepackage{multirow}           
\usepackage{array,hhline}   
\usepackage{hyperref}
\usepackage{color}
\usepackage{subfigure}
\usepackage{lmodern}
\usepackage{fancyhdr}
\usepackage{multicol}
\usepackage{url}
\usepackage{indentfirst}
\usepackage{helvet}
\usepackage{mathrsfs}

\makeindex

\begin{document}
\sloppy
%
%
\newcommand{\paginalimpia}{\clearpage{\pagestyle{empty} \cleardoublepage}}
\makeatletter
\def\cleardoublepage{\clearpage\if@twoside \ifodd\c@page\else
\hbox{} \vspace*{\fill} \vspace{\fill} \thispagestyle{empty}
\newpage
\if@twocolumn\hbox{}\newpage\fi\fi\fi}
\makeatother
\pagestyle{empty}
%

\begin{figure}[t]
    \begin{center}
      \includegraphics[width=.4\textwidth]{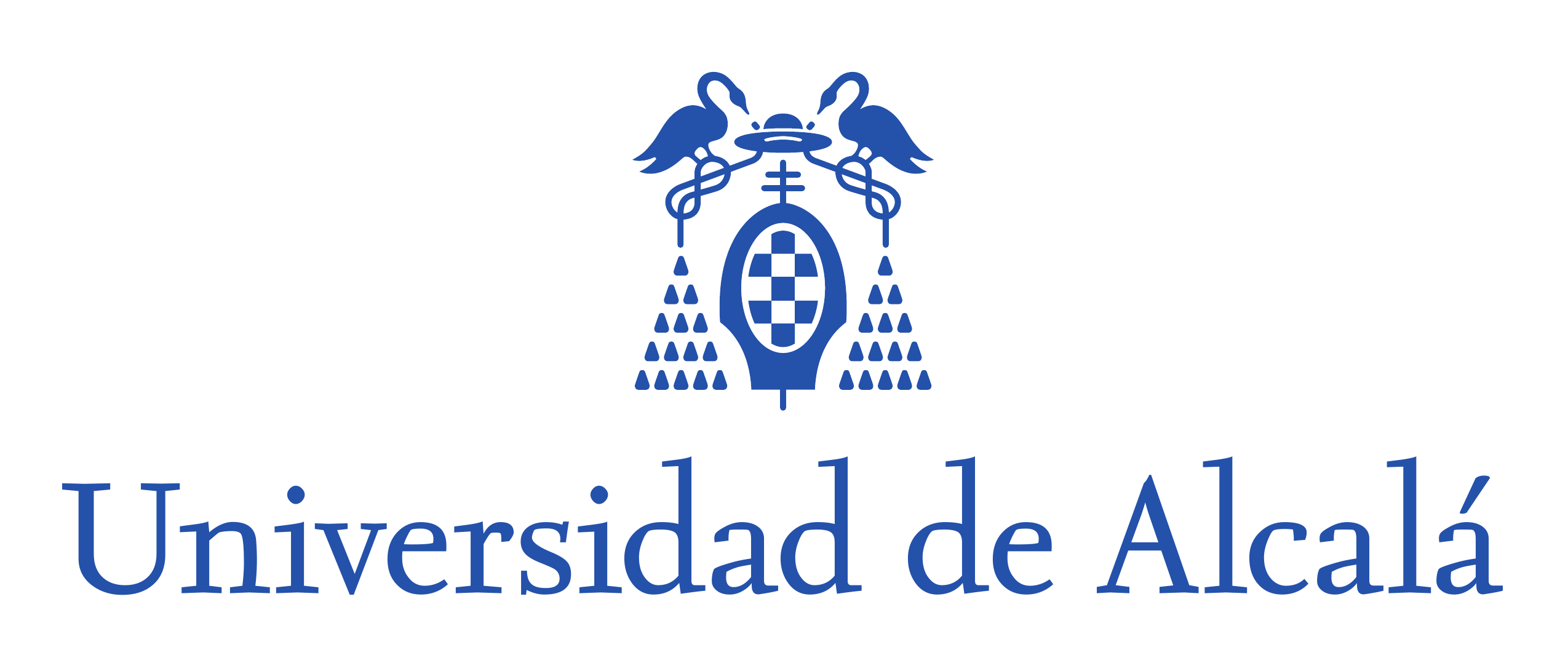}\\
    \end{center}
\end{figure}

\begin{center}
\LARGE \textsc{Escuela Politécnica Superior}\\
\vspace{1.0cm}
\LARGE \textsc{Dpto. de Teoría de la Señal y Comunicaciones}\\
\vspace{1.5cm}
\LARGE \textsc{\textcolor[rgb]{0.09,0.09,0.59}{Doctorado en Tecnologías de la Información y las Comunicaciones}}\\
\vspace{1.0cm}
\LARGE \textcolor[rgb]{0.09,0.09,0.59}{Memoria de tesis doctoral}\\
\vspace{2.0cm}
\textbf{\huge \textsc{\textcolor[rgb]{0.09,0.09,0.59}{Contributions to the development of the CRO-SL algorithm: Engineering applications problems}}}\\
\vspace{2.0cm}
\large Autor:  D. Carlos Camacho Gómez\\
\large Director: Dr. Sancho Salcedo Sanz\\
\large Codirectora: Dra. Silvia Jiménez Fernández\\
\end{center}


\newpage

\paginalimpia

\vspace*{8.0cm}
\begin{flushright}
\emph{A quienes siempre me han apoyado}\\
\end{flushright}

\paginalimpia

\chapter*{Abstract}\label{Abstract}
This Ph.D. thesis discusses advanced design issues of the evolutionary-based algorithm \textit{"Coral Reef Optimization"}, in its Substrate-Layer (CRO-SL) version, for optimization problems in Engineering Applications. Evolutionary Algorithms (EAs) have been widely applied to hard optimization problems when analytic approaches are not applicable. These problems usually have huge search spaces, a high number os constrains and unknown or discrete objective functions. While computational complexity of EAs algorithms is moderate, fitness function time executions may also be heavy. The increasing processing capacity in technology opens the door to tackling costly problems through meta-heuristics. One of the drawbacks of using this approaches is that it is not possible to know in advance which of them is the best for a specific problem (No-free lunch Theorem) and, whatever the choice, its application does not ensure to get the optimum solution, given the stochastic nature of the meta-heuristic. That is the reason why we have chosen CRO-SL algorithm, because it allows to combine the most powerful search procedures in a kind of co-evolution competitive approach, helping each other in order to attain the global optimum.

By applying the CRO-SL to these problems, we aim to satisfy two main objectives: first, to test the goodness of the CRO-SL algorithm in the selected applications. To do this, alternative state of the art objective meta-heuristics will be compared with the proposed CRO. The second is to promote the use of the CRO-SL as a tool for comparison between exploration methods. Some of the meta-heuristic algorithms are based on the iteration of a search method over a population of codified individuals, who represent solutions of the problem. The CRO-SL borrows the form in which other algorithms change their individuals, and forms new solutions in parallel. Among the best-known EAs included as substrates in the CRO-SL are: Harmony Search, Differential Evolution, Genetic Algorithms. In addition we propose and analyze the use of another type of mutations such as the Gaussian type, single mutation or multipoint crossover. Finally, during the development of this Thesis a new form of search based on strange attractors has also been tested.

The problems that can be tackled with meta-heuristic approaches is very wide and varied, and it is not exclusive of engineering. However we focus the Thesis on it area, one of the most prominent in our time. One of the proposed application is battery scheduling problem in Micro-Grids (MGs). Specifically, we consider an MG that includes renewable distributed generation and different loads, defined by its power profiles, and is equipped with an energy storage device (battery) to address its programming (duration of loading / discharging and occurrence) in a real scenario with variable electricity prices. Also, we discuss a problem of vibration cancellation over structures of two and four floors, using Tuned Mass Dampers (TMD's). The optimization algorithm will try to find the best solution by obtaining three physical parameters and the TMD location. As another related application, CRO-SL is used to design Multi-Input-Multi-Output Active Vibration Control (MIMO-AVC) via inertial-mass actuators, for structures subjected to human induced vibration. In this problem, we will optimize the location of each actuator and tune control gains. Finally, we tackle the optimization of a textile modified meander-line Inverted-F Antenna (IFA) with variable width and spacing meander, for RFID systems. Specifically, the CRO-SL is used to obtain an optimal antenna design, with a good bandwidth and radiation pattern, ideal for RFID readers. Radio Frequency Identification (RFID) has become one of the most numerous manufactured devices worldwide due to a reliable and inexpensive means of locating people. They are used in access and money cards and product labels and many other applications.

%

\chapter*{Resumen}\label{Resumen en Español}

Esta tesis doctoral aborda el diseño de un nuevo tipo de Algoritmo Evolutivo (EA) \textit{"Coral Reef Optimization"}, en su versión Substrate-Layer, para la optimización de problemas en diferentes ámbitos de la ingeniería. Los EAs han sido ampliamente aplicados a problemas de optimización difícilmente abordables de manera analítica, ya sea por tener espacios de búsqueda muy grandes o muchas restricciones, o por ser no lineales y de naturaleza discreta. Si bien la ejecución de estos algoritmos no supone un gran coste computacional hoy en día, sí lo suponen las funciones de coste que constantemente deben evaluar los algoritmos. La creciente capacidad de procesamiento en la tecnología abre las puertas al abordaje de problemas computacionalmente costosos por medio de la metaheurística. Uno de los inconvenientes de ésta, es que no hay forma de saber a priori cuál de ellos es mejor para un problema específico (No-free lunch Theorem), y sea cual sea la elección, la ejecución del mismo no asegura que se vaya a obtener el óptimo dada la naturaleza estocástica de estos algoritmos. Por este motivo se ha elegido el algoritmo CRO-SL, ya que permite combinar los procesos de búsqueda más potentes, potenciándose entre ellos para alcanzar el óptimo global del problema.

Mediante la aplicación del CRO-SL a estos problemas se pretende cumplir dos objetivos. El primero es comprobar la aptitud del propio algoritmo en las aplicaciones mencionadas. Para ello además se realizarán experimentos con los algoritmos más populares y los resultados podrán ser comparados entre sí. El segundo es promover el uso del CRO-SL como herramienta de comparación entre métodos de exploración. Algunos de los algoritmos metaheurísticos se basan en la iteración de un proceso de búsqueda sobre una población de individuos codificados, que encarnan la solución a un determinado problema. El CRO-SL toma prestado la forma en la que otros algoritmos cambian a sus individuos, y forma nuevas soluciones de manera paralela. Entre los algoritmos evolutivos más conocidos que vamos a ver durante el desarrollo de esta tesis se encuentran los algoritmos Harmony Search, Differential Evolution y Genetic Algorithm. Además se verán otro tipo de mutaciones como la de tipo Gaussiana, mutación simple o cruce multipunto. Por último, durante el desarrollo de esta tesis también se ha probado una nueva forma de búsqueda basada en atractores extraños. Gracias a la capacidad de comparación del CRO-SL podremos ver si esta nueva forma de búsqueda es útil o no.

La problemática a la que se puede aplicar la metaheurística es muy variada y no tiene por qué ser exclusiva de la ingeniería, sin embargo esta Tesis está centrada en este ámbito. La primera aplicación discutida en la Tesis
es un problema de optimización de planificación de las baterías en micro-redes (MG). Específicamente, consideramos una MG que incluye generación renovable y diferentes cargas, definidas por sus perfiles de potencia, y está equipada con un dispositivo de almacenamiento de energía (batería) para abordar su programación (duración de carga / descarga y ocurrencia) en un escenario real de precios variables de electricidad. La segunda apliucación abordada es un problema de control de vibración en estructuras de dos y cuatro pisos mediante el uso de elementos amortiguadores pasivos, TMD's (Tunned Mass Dampers). Esta aplicación viene motivada por la necesidad de cancelar vibraciones procedentes de la Tierra, como pudiera ser en un terremoto. En este caso el algoritmo no sólo intentará optimizar las características físicas de los TMD's sino también su colocación dentro del edificio. En tercer lugar, se abordará una aplicación relacionada, pero centrada en un control activo de las vibraciones que generamos los humanos al caminar en una estructura civil, mediante el uso de actuadores de masa inercial. En este problema se tratará de optimizar la localización de los actuadores así como sintonizar las ganancias de control. Por último se aborda el diseño de una antena de tipo F invertida (IFA), para sistemas de IDentificación por Radio-Frecuencia (RFID). Este tipo de dispositivos han sido muy utilizados en productos a lo largo de todo el mundo, tanto en tarjetas de crédito como en etiquetas de productos debido a su pequeño tamaño y a una fabricación sencilla y barata. En concreto, en este trabajo se usarán como conductores láminas de cobre y como dieléctrico, fieltro. Se pretende así, diseñar el ancho y el espaciamiento de estas tiras de cobre para que emita en un ancho de banda determinado con una calidad determinada.

\paginalimpia

\chapter*{Agradecimientos}
En este capítulo quiero agradecer a aquellas personas que me han acompañado en este camino el haber estado ahí. Es un camino porque ha sido una etapa de mi vida que me ha llevado de un estado de madurez a otro, tanto en lo profesional como en lo sentimental. Una etapa de mi vida que ha tenido momentos mejores y peores pero sólo sobresalen todo lo que he aprendido, las nuevas experiencias que he vivido y lo bien que me lo he pasado. Este trabajo ha sido posible gracias al apoyo, confianza y cariño de ciertas personas que mencionaré a continuación. Esas personas no son ni la mitad de las que me gustaría mencionar, y los agradecimientos son menos de la mitad de lo que la mitad de ellos merecen: 

\begin{itemize}
\item Al Dr. Sancho Salcedo Sanz, cuyo conocimiento ha resultado fundamental para poder encauzar adecuadamente los numerosos hitos alcanzados y a su hambre investigadora que me ha motivado durante estos casi tres años de trabajo. También tengo que agradecerle su apoyo y confianza en mí en todo momento. Una cosa que no se suele agradecer porque no se sabe a quién hay que hacerlo, es a que sea como es, de esa forma única, que hace que un jefe se vuelva tu amigo y el trabajo con él se vuelva fácil y divertido.

\item De los Drs. José Antonio Portilla y Silvia Jiménez he intentado aprender su metodología de trabajo y me he intentado empapar de su inteligencia y conocimiento. Me ha gustado mucho trabajar con ellos porque si bien hemos abordado ciertos trabajos de forma seria y profesional, siempre ha sido rodeados de un aire de buen rollo. Gracias.

\item Hacer antes de nada un agradecimiento especial al resto de profesores integrantes al grupo GHEODE, porque entre todos tienen un ambiente muy positivo, que hace el día a día muy ameno. Este grupo tiene la buena costumbre de colaborar en sus investigaciones con otros departamentos y otros centros y universidades. Es por ello que he tenido el gusto de trabajar con los Drs. Emiliano Pereira y Ricardo Mallol, entre otros, de los que he ampliado mis áreas de conocimiento enormemente.

\item Cierto es que esta estancia no ha sido sólo trabajar, también he hecho amistades de las que me siento más que orgulloso. Mis compis de laboratorio, Laura y Adrián, con los que compartía los tranchetes que nos pasaban por debajo de la puerta (es una broma del grupo). Ahora en serio, yendo por partes, Adrián es una de esas personas que se vuelve tu amigo sin casi saber cómo, y no me extraña, ya que ser amigo de Adri es muy fácil. Las risas que nos hemos echado son pocas comparadas con las que nos quedan. A Laura ya la conocía de antes, pero realmente no supe lo genial que era hasta que no pasé más tiempo con ella. Gracias por regalarme una sonrisa cada mañana. Para completar el grupo de los cuatro fantásticos está Freddy, un gran amigo que ha estado ahí estos años y con los que hemos compartido buenas ideas de negocio, buenas charlas y buenas tarde de gimnasio. Gracias a todos ellos, a su forma de ser, venir a la universidad ha sido muy agradable, en especial nuestras conversaciones en la comida, algo que sin duda echaré de menos en el futuro. Sé que todos triunfarán en la vida y yo estaré ahí para verlo. Por último no puedo olvidar a Pedro, un amigo para toda la vida, con el que he pasado tanto tiempo en la carrera y ahora en el doctorado. Gracias por ser tan trabajador y tan humilde, que sepas que tenemos pendiente el viaje a Ucrania.

\item Por último agradecer a mi familia ser tan buena como es. Mis padres, Esperanza y José Carlos, que siempre me apoyan en mis decisiones y confían en mí continuamente. Mis hermanas Paloma, Margarita y Esperanza, porque cada una tiene una forma de ser tan diferente y maravillosa a la vez, que estar con ellas me fascina. El resto de familia, abuelos, abuelas, tíos, tías, primas, etc, siempre han tenido buenas palabras para mí y me han mostrado siempre su cariño. Hay una persona a la que tengo que dar las gracias a la que más, por eso la dejo para el final. Sandra, gracias por aguantarme estos tres años y por quererme tanto. Siempre seremos felices con Thor y Loki. 
\end{itemize}

\paginalimpia

\pagenumbering{roman}

\tableofcontents
\listoffigures
\listoftables

\paginalimpia

\chapter*{List of Acronyms}

\begin{tabular}{ll}
\hline
{\em General}& \\
\hline
LSGO & Large Scale Global Optimization\\
SME & Small and Medium-size Enterprises\\

\hline
{\em Battery Scheduling Application}& \\
\hline
BSOP & Battery Scheduling Optimization Problem\\
DG & Diesel Generator\\
MG & Micro-Grid\\
PV & Photo-Voltaic\\
SOC & State of Charge\\

\hline
{\em Passive Vibration Control Application}& \\
\hline
FRF & Frequency Response Function\\
TMD & Tuned Mass Dampers\\

\hline
{\em Active Vibration Control Application}& \\
\hline
AVC & Active Vibration Control\\
DVF & Direct Velocity Feedback\\
MIMO & Multi-Input Multi-Output\\
PI & Performance Index\\
SCI & Steel Construction Institute\\
SISO & Single-Input Single-Output\\
VDV & Vibration Dose Value\\

\hline
{\em Antenna Design Application}& \\
\hline
IFA & Inverted-F Antenna\\
PCB & Printed Circuit Board\\
RFID & Radio Frequency IDentification\\

\hline
\end{tabular}

\begin{tabular}{ll}

\hline
{\em Algorithms}& \\
\hline
2Px & 2-Points crossover\\
ACO & Ant Colony Optimization\\
CRO & Coral Reefs Optimization algorithm\\
CRO-SL & Coral Reefs Optimization with Substrate Layer algorithm\\
DE & Differential Evolution\\
DECC-G & Differential Evolution Cooperative Coevolution Grouping\\
EA & Evolutionary Algorithm\\
ECBO & Enhanced Colliding Bodies Optimization\\
EP & Evolutionary Programming\\
ES & Evolutionary Strategies\\
GA & Genetic Algorithm\\
GM & Gaussian Mutation\\
HMCR& Harmony Memory Considering rate\\
HS & Harmony Search\\
MPx & Multi-Point Crossover\\
PAR & Pitch Adjusting Rate\\
PSO& Particle Swarm Optimization\\
SA & Simulated Annealing\\
SAbM & Strange Attractors based Mutation\\
\hline

\end{tabular}

\paginalimpia

\pagenumbering{arabic}
\pagestyle{fancy}
\fancyhf{}
\fancyhf{}
\fancyhead[LO]{\rightmark} 
\fancyhead[RE]{\leftmark} 
\fancyhead[RO,LE]{\thepage} 
\renewcommand{\chaptermark}[1]{\markboth{{Chapter \thechapter. #1}}{}} 
\renewcommand{\sectionmark}[1]{\markright{{\thesection. #1}}} 
\renewcommand{\headrulewidth}{0.0pt}
\setlength{\headheight}{15pt}

\part{Introduction and algorithmics state of the art}\label{part:estadodelarte}

\chapter{Introduction and algorithm definition}\label{cap:introestadodelarte}

\section{Motivation}
This Thesis elaborates on the application of the CRO-SL algorithm to several engineering optimization problems. However, its application and its development was not evident. During the initial phase of this research, the basic CRO was developed in order to tackle one of the problems discussed later on the Thesis. As many other optimization problems, it presented a large dimension search space and high number of constrains, so it was necessary a powerful search tool. At this point, we decided to build the first version of CRO with Substrate Layer to try to explore the search space more efficiently. After evaluating the results obtained, two objectives were immediately bring into play: The first was that the algorithm should be able to be adapted to any alternative optimization problem with the least possible effort, in order to be able to evaluate its results. That is the reason why it was programmed in the most general way possible, and allowed an easy and quick adaptation to different optimization problems. Another of the keys of this adaptability of the CRO-SL is that the substrates were programmed as external functions. The second objective we pursued was that the algorithm should allow the incorporation and exclusion of substrates without hard efforts. In this way, it was possible to perform an algorithm execution with all the substrates and based on the results, or choose the ones that worked better for a definitive execution. Once obtained the final results, it would be very interesting to see which of them have behaved better and for this, several parameters are extracted for an a posteriori comparison. Those parameters are the number of larvaes enter to the coral per substrate and per iteration, and the percentage of times that every substrate pulls out the best larvae in every iteration. In order to better analyze this information, graphic paths will be taken, which would close the third objective: make decisions about which substrates enter the process. Therefore, the motivation of this work does not only come from the desire to develop an algorithm able to adapt to many optimization problems but also lies in doing it in a simple way.

\section{Introduction}\label{sec:Introduction}
In the last years, engineers and scientist around the world have dedicated their efforts to tackle optimization problems through meta-heuristic algorithms. This is due to the good quality of the solutions these approaches produce, and the light run-times they employ. Furthermore, they show a great performance when solving problems with special restrictions, of high dimensions, or objective functions with non-linear or discrete search spaces. Classical approaches do not provide, in general, good solutions in these cases or they are just unable to be applied to them. In this context, modern optimization heuristics and meta-heuristics have been lately the core of research, aimed at solving the aforementioned lack of efficient methods. A good number of such algorithms are bio-inspired techniques such as Evolutionary Algorithms (EA), which includes a whole family of techniques such as Genetic Algorithms \cite{Ei03}, Evolutionary Strategies \cite{Be99}, Evolutionary Programming \cite{Ya99}, and Differential Evolution \cite{St97} among others. All of them are based on darwinian concepts as survival of the fittest and natural evolution. Likewise, some of the most famous algorithms between the bio-inspired ones are, Ant Colony Optimization \cite{Do96}, which is based on social behaviour of ants, Cuckoo Optimization Algorithm \cite{Ra11}, which is inspired by the egg laying and breeding of this bird family and Particle Swarm Optimization \cite{Ke95} that imitates bird flocking and fish schooling as a swarm, how they move, change their positions, and trajectories to find their destination. There are many other bio-inspired meta-heuristics, with approaches such as Immune Systems Algorithm \cite{Ke94} focus on imitating the behavior of the immune system in animals, Artificial Bee Colony \cite{Ka08} which is based on the intelligent foraging behaviour of honey bee swarm in the hive, as well as Invasive Weed Optimization \cite{Me06} based on weed growth and their invasive properties and Hunting Search Algorithm \cite{Of10} inspired by group hunting of animals such as lions, wolves, and dolphins, etc.

Other kind of meta-heuristics are inspired by physics phenomena that occurs in nature, as Ray Optimization Algorithm \cite{Kh12} or Gravitational Search Algorithm \cite{Ra09}. The first imitate the light refraction law when light travels from a lighter medium to a darker medium and the second is inspired by the gravitational law. Other methods inspired by Physics process that has been widely used \cite{Salcedo16b}. Among others Artificial Chemical Reaction Optimization Algorithm \cite{Al11}, Electromagnetism-like Algorithm \cite{Il03} and Big-Bang Big-Crunch Algorithm \cite{Er06}. In fact a Physics inspired algorithm has been used in this Thesis, the Simulated Annealing Algorithm (SA) \cite{Ki83} which has been adapted as a substrate of the coral reef algorithm.  The SA is a probabilistic single-solution-based search method inspired by the annealing process in metallurgy. The Harmony Search Algorithm \cite{Ge01} mimics the improvisation of music players for finding a best harmony all together which is another search substrate. There are other methods based on the geographical distribution of some living organisms as Virus-Evolutionary Genetic Algorithm \cite{Cu09}, Bacterial Colony Foraging optimization algorithm \cite{Ch14}, Amoeba Algorithm \cite{Xi17} and Colliding Bodies Optimization \cite{Kaveh14}. Also, in optimization domain, researchers have developed many effective stochastic techniques that mimic the specific behavior of human beings. In this context can be highlighted Teaching-Learning-Based Optimization Algorithm \cite{Do18}, Society and Civilization-based Algorithm \cite{Ra03} and Imperialist Competitive Algorithm \cite{Ah17}.

Due to there is a large number and very different meta-heuristics algorithms, and there is no way to know, a priori, which one would be the best (No-Free-Lunch Theorem) \cite{Wolpert97}, in the last years there have been appeared approaches that combines two or more of these methods. In this context can be founded \cite{Fa18}, where the authors applies a new hybrid method by a combination of three population base algorithms such as Genetic Algorithm, Particle Swarm Optimization and Symbiotic Organisms Search in order to tackle continuous optimization problems, and \cite{Ma17} where Simulated Annealing has been embedded to the Whale Optimization Algorithm \cite{Mi16} for feature selection. One of the main characteristic of the approaches mentioned is that they use to execute the different in serial mode, it means that one method can not be applied till another one had finished. Here is where CRO-SL highlights, because its search procedures are executed in parallel, resulting in a fully competitive co-evolution. This characteristic, among others, will be fully described in the Thesis.

In the following sections of this Chapter the algorithms and search procedures used in this Thesis will be described. Thus, the basic CRO and CRO-SL will be fully detailed. thus, in the Application Sections there would not be necessary to explain the algorithm details in depth. The objective of this theoretical review is not to elaborate an exhaustive study about the current state of art regarding the algorithms treated, but rather to establish an elementary theoretical framework that allows the correct understanding and evaluation of the different techniques and alternatives that are propose throughout the rest of this work.

The organization of the rest of this Thesis is the following. We have devoted one chapter per engineering application tackled. Each chapter starts with a small state of the art about the application and also, a deep explanation on the problem will be given. As it can be assumed, each problem has different characteristics and specific needs, so the CRO-SL will adapt its encoding and even the type of the substrates implemented. Therefore, a brief analysis on the characteristics of the algorithm will be presented for each application. At the end of each chapter, the results of the application of CRO-SL to the problem will be shown and some conclusions will be extracted. Thus, in Chapter \ref{cap:scheduling} it is proposed a CRO-SL algorithm to tackle the BSOP in MGs. In chapter \ref{cap:tmds} it is showed how CRO-SL is able to tackle the TMD design and location problem for passive control of ground motions. Chapter \ref{cap:avcs} deals with an active vibration control systems to real complex structures (with a large number of vibration modes and/or with a large number of test points) by achieving global optimum designs with affordable computation time via CRO-SL. In chapter \ref{cap:antena} an optimal textile antenna design for RFID is presented using the proposed algorithm. To finalize this Thesis, a number of final conclusions will be extracted as well as future research lines that would be interesting to follow will be given.

\section{Corals and Coral Reefs} \label{sec:Corales}
This section describes some important properties of corals and coral reefs, that will be simulated by the basic CRO approach. First it will be described some characteristics of corals and reefs, and then we focus on corals reproduction.\\

\subsection{Corals and Reef Formation}\label{subsec:Formation}

A coral is an invertebrate animal belonging to the group Phylum Cnidaria, which also includes sea anemones, hydras, or jellyfishes \cite{Burk08}. In fact, a more detailed classification includes corals in the Anthozoa class, together with sea anemones, sea pens, or sea pansies. These animals are characterized by their ability to subsist either as individuals or in colonies of polyps, living attached to a substrate. There are more than 2500 different species of corals, living in shallow and deep waters, and each year new species are found and described. An important subclass of corals are reef-building corals, also known as hermatypic or simply "hard corals". Hard corals are usually shallow-water animals that produce a rigid skeleton of calcium carbonate, segregated from their base. A coral reef is formed by hundred of hard corals, cemented together by the calcium carbonate they produce. Periodically, the polyp lifts off its basal plate of calcium carbonate and secrete a new one, forming a tiny chamber that will contribute to the coral’s skeleton. Polyps continuously build these chambers in the reef, so finally the complete reef grows upwards. Living corals grow on top of the skeletons of calcium carbonate of their dead predecessors. A coral reef is usually formed by corals living in colonies or on its own. A colony is composed of a single specie of coral, but a reef’s structure can comprise multiple types of species. In fact, a coral reef finally ends up as a truly ecosystem, in which a diverse collection of animals and plants interact with each other, as well as with their environment. In addition to corals, many other animals and plants live in and from the reef, such as algae, sponges, sea anemones, bryozoans, sea stars, crustaceans (e.g., shrimps, crabs, lobsters), octopuses, squids, clams, snails, and other mollusca. And, of course, a huge variety of fish that find shelter and food in the reef.

In general, hard coral species require free space to settle and grow. Although a-priori the implementation of this settlement procedure might be easy for a potential new member of the reef, in practice free space is an extremely limited resource in the reef environment \cite{Gen94}. As a result, species often compete with each other or exhibit aggressive behavior to secure or maintain a given plot of substrate \cite{Ates89}. Different strategies used by corals to compete for the space have been thoroughly described in the literature \cite{Chad87}, \cite{Ates89}. Among them, fast growing is deemed as the most used and simple strategy since it grounds on the fact that there are corals that have evolved to yield a faster growth rate than others. When a fast-growing coral sets near a slow-growing one, the former attacks the latter by overtopping it. The underlying coral suffers from light deficiency, thus affecting its ability to conduct photosynthesis and to get into contact with food particles. As time evolves, overtopping by fastgrowing species kills the slower-growing species underneath. Other aggressive strategies carried out by some species of corals include sweeper tentacles (i.e., detect and damage adjacent coral colonies), mesenterial filaments (namely, enabling external digestion of neighboring colonies), and terpenoid compounds (coral chemical warfare).

\subsection{Coral reproduction}\label{subsec:Reproduction}
Corals can reproduce in two different modes: sexual or asexual. In fact, an individual polyp may use both modes within its life time. Furthermore, sexual reproduction can be either external or internal, depending on the coral species.

\begin{enumerate}
\item Sexual External Reproduction: \textit{Broadcast Spawning}.\\
The majority of hard corals species resort to a sexual external reproduction method known as broadcast spawning \cite{Mol12}: every coral produces male and/or female (some species of corals are hermaphrodites) gametes that are massively released out to the water. Once the egg and sperm meet together, a larva (also called planula) is produced. Planulae float in the water until they find a proper space to attach and start growing a polyp \cite{Tay11}. In the majority of reefs, the phenomenon of coral spawning occurs as a synchronized event. This timing is crucial for successful reproduction, since corals cannot move to force reproductive encounters. There are different natural aspects that affect the timing of the corals’ spawning, such as temperature, day length, or temperature change rate.
\item Sexual Internal Reproduction: \textit{Brooding}.\\
Brooding is a method of internal reproduction used by some species of corals. In this reproduction mode, some female polyps contain eggs that are not released to the water. Instead, sperm released by other male corals of the same species gets inside the polyp and fertilizes the eggs, producing small planulae. These planulae are released later through the mouth of the coral in an advanced stage of development, so it becomes easier for these planulae to set onto hard substrate without being attacked or depredated. There has also been described a type of brooding reproduction in hermaphrodite corals \cite{Braz98}.
\item Asexual Reproduction: \textit{Budding or Fragmentation}.\\
Budding is a form of asexual reproduction in corals: basically, new polyps bud off from parent polyps to expand or begin new coral colonies \cite{Yamashi98}. Budding occurs when the coral has grown enough to produce budding. Fragmentation is a process similar to budding, but it is caused by external phenomena (e.g., storms or boats’ grounding), and usually a larger part of the coral is divided in comparison to budding \cite{Lirman00}. As such, in fragmentation apart of a coral colony is separated from the parent polyps. Individuals broken off this way from the main colony are able to keep growing and finally establishing a new colony far way from the parent one if conditions are favorable. It is important to note that both budding and fragmentation processes produce polyps that are genetically identical to the parent polyp/colony.\\
\end{enumerate}

\subsection{Reef Longevity and Causes of Death}\label{subsec:CauseDeath}
There are not reliable statistics on corals’ lifespan. However, it is well known that coral colonies can live for several centuries. Corals and coral reefs must face different hazards during their life. In larva state, corals are massively depredated by fishes and other predators. However, the huge number of larvae produced in broadcast spawning reproduction ensures that enough polyps settle in favorable ground and start forming a colony. On the other hand, coral polyps encounter many types of predators including sea stars, parrot-fishes, or butterfly-fishes. Human activities (e.g., fishing activities or industrial processes that increase ocean pollution) and climate changes (increase of the oceans’ temperature, among others) also contribute to the loss of living corals \cite{Lesser04}.

\section{The Coral Reef Optimization Algorithm}\label{sec:CROAlgorithm}
The CRO algorithm \cite{Sancho14} is an evolutionary-type meta-heuristic, which artificially simulates a coral reef, where different corals (namely, solutions to the optimization problem considered) grow and reproduce in coral colonies, fighting for space in the reef. This fight for space, along with the specific characteristics of the corals reproduction, produces a robust metaheuristic algorithm shown to be powerful for solving hard optimization problems. The proposed CRO approach can be regarded as a cellular-type evolutionary scheme, with superior exploration-exploitation properties thanks to the particularities of the emulated reef structure and coral reproduction. This algorithm has been applied in different kind of problems as \cite{Sal14}, where basic CRO is applied for offshore wind farm design and layout optimization, or \cite{SalPolu14} where the authors tackle the Mobile Network Deployment Problem (MNDP),  in which the control of the electromagnetic pollution plays an important role. In \cite{SalFeature14} the research is based on a Feature Selection Problem carried out with the CRO, that must obtains a reduced number of predictive variables out of the total available, so this set finally provides the wind speed prediction over an Extreme Learning Machine. The CRO was also used for feature selection in \cite{SalHS15}, where the search procedure was taken of the Harmony Search algorithm, providing very good results. In \cite{Sal16} the CRO was proposed as a grouping algorithm to tackle the Mobile Network Deployment Problem mentioned above. Finally, one of the latest research is \cite{Antonio17} where the main parameters of the CRO algorithm are adjusted based on the mean and standard deviation associated with the fitness distribution and was combined with two Local Search methods for time series segmentation. Thus, this section is focused on explaining the basic CRO algorithm.

Having these fundamentals on the coral's reproduction and formation in mind, the CRO algorithm tackles optimization problems by modeling and simulating all the distinct processes explained in Section \ref{sec:Corales}. Let $\Lambda$ be a model of reef, consisting of a $N \times M$ square grid. We assume that each square $(i, j)$ of $\Lambda$ is able to allocate a coral (or colony of corals) $\Xi$, representing different solutions to our problem, encoded as strings of numbers in a given alphabet  $\mathcal{F}$. The CRO algorithm is first initialized at random by assigning some squares in $\Lambda$ to be occupied by corals and some other squares in the grid to be empty; that is, holes in the reef where new corals can freely settle and grow. The rate between free/occupied squares in $\Lambda$ at the beginning of the algorithm is an important parameter of the CRO algorithm, which will be denoted in what follows as $0<\rho_0<1$. Each coral is labeled with an associated health function $f(\Xi) : \mathcal{F} \rightarrow \mathbb{R}$ that represents the problem’s fitness function. Note that the reef will progress as long as healthier corals (which represent better solutions to the problem at hand) survive, while less healthy corals perish.

After the reef initialization described above, a second phase of reef formation is carried out by the CRO algorithm. To this end, a simulation of the corals’ reproduction in the reef is done by sequentially applying different operators. This sequential set of operators is then applied iteratively until a given stop criterion is met. Thus, we define different operators for modeling sexual reproduction (broadcast spawning and brooding), asexual reproduction (budding), and polyps depredation. In both sexual and asexual reproduction we give the conditions under which new corals effectively get attached to the reef or are depredated while at the larvae phase, it is as follows:
\begin{itemize}
\item Broadcast Spawning (External Sexual Reproduction). The modeling of coral reproduction by broadcast spawning consists of the following steps.
\begin{itemize}
    \item In a given step $k$ of the reef formation phase, select uniformly at random a fraction of the existing corals $\rho_{k}$ in the reef to be broadcast spawners. The fraction of broadcast spawners with respect to the overall amount of existing corals in the reef will be denoted as $F_b$ . Corals that are not selected to be broadcast spawners (i.e.,1-$F_b$) will reproduce by brooding later on, in the algorithm.\\
    \item Select couples out of the pool of broadcast spawner corals in step $k$. Each of such couples will form a coral larva by sexual crossover, which is then released out to the water. Note that, once two corals have been selected to be the parents of a larva, they are not chosen anymore in step $k$ (i.e., two corals are parents only once in a given step). These couple selection can be done uniformly at random or by resorting to any fitness proportionate selection approach (e.g., roulette wheel).\\
\end{itemize}

\item Brooding (Internal Sexual Reproduction).As previously mentioned, at each step $k$ of the reef formation phase in the CRO algorithm, the fraction of corals that will reproduce by brooding is 1-$F_b$. The brooding modeling consists of the formation of a coral larva by means of a random mutation of the brooding-reproductive coral (self-fertilization considering hermaphrodite corals). The produced larva is then released out to the water in a similar fashion than that of the larvae generated in step (4) in algorithm \ref{alg:CRO}.\\
\item Larvae Setting. Once all the larvae are formed at step $k$ either through broadcast spawning or by brooding, they will try to set and grow in the reef. First, the fitness function of each coral larva is computed. Second, each larva will randomly try to set in a square ($i$, $j$)of the reef. If the square is empty (free space in the reef), the coral grows therein no matter the value of its health function. By contrast, if a coral is already occupying the square at hand, the new larva will set only if its fitness function is better than that of the existing coral. We define a number $\kappa$ of attempts for a larva to set in the reef: after $\kappa$ unsuccessful tries, it will be depredated by animals in the reef.\\
\item Asexual Reproduction. In the modeling of asexual reproduction (budding or fragmentation), the overall set of existing corals in the reef are sorted as a function of their level of healthiness (given by $f(\Xi_{ij})$), from which a fraction $F_a$ duplicates itself and tries to settle in a different part of the reef by following the setting process described before.\\
\item Depredation. Corals may die during the reef formation phase of the CRO algorithm. At the end of each reproduction step $k$, a small number of corals in the reef can be depredated, thus liberating space in the reef for next coral generation. The depredation operator is applied with a very small probability $P_d$ at each step $k$, and exclusively to a fraction $F_d$ of the worse health corals in $\Lambda$. For the sake of simplicity in the parameter setting of the CRO algorithm, the value of this fraction may be set to $F_d$ =$F_a$. Any other assignment may also apply provided that $F_d$ + $F_a$ $\leq$ 1 (i.e., no overlap between the asexually reproduced and the depredated coral sets).\\
\end{itemize}

\section{Coral Reef Optimization Algorithm with Substrate Layer}\label{sec:CRO-SLAlgorithm}
The CRO-SL algorithm is a modification of the original CRO approach, based on the fact that there are many more interactions in real reef ecosystems which can be also modelled and incorporated to the CRO approach to improve it. For example, different studies have shown that successful recruitment in coral reefs (i.e. successful settlement and subsequent survival of larvae) depends on the type of substrate on which they fall after the reproduction process \cite{Vermeij05}. This specific characteristic of the coral reefs was first included in the CRO in \cite{Sancho_IJBIC}, in order to solve different instances of the Model Type Selection Problem for energy applications. In \cite{Sancho_IJBIC}, different substrate layers were defined in the CRO, in such a way that each layer represents a different model to evaluate the energy demand estimation in Spain, from macro-economic variables. This algorithm has been tested in LSGO problems \cite{PacoHe}, specifically to a specially designed test suite. The performance of the CRO-SL was combined with a Local Search procedure and was compared to the reference algorithm DECC-CG, proving that the proposal, not been specifically designed for LSGO, obtained better results, specially in the most complex functions: non-separable, and overlapping functions. The CRO-SL is a much more general approach: it can be defined as an algorithm for competitive co-evolution, where each substrate layer represents different processes (different models, operators, parameters, constraints, repairing functions, etc.). Taking into account the fact that there is not a form of mutation that a-priori is better in a certain problem, the CRO-SL can be used to compare these forms of mutation. Also the purpose of this Thesis is to use the algorithm CRO-SL in the resolution of different engineering optimization problems.

\begin{figure}[!ht]
\centering
\includegraphics[width=0.5\textwidth]{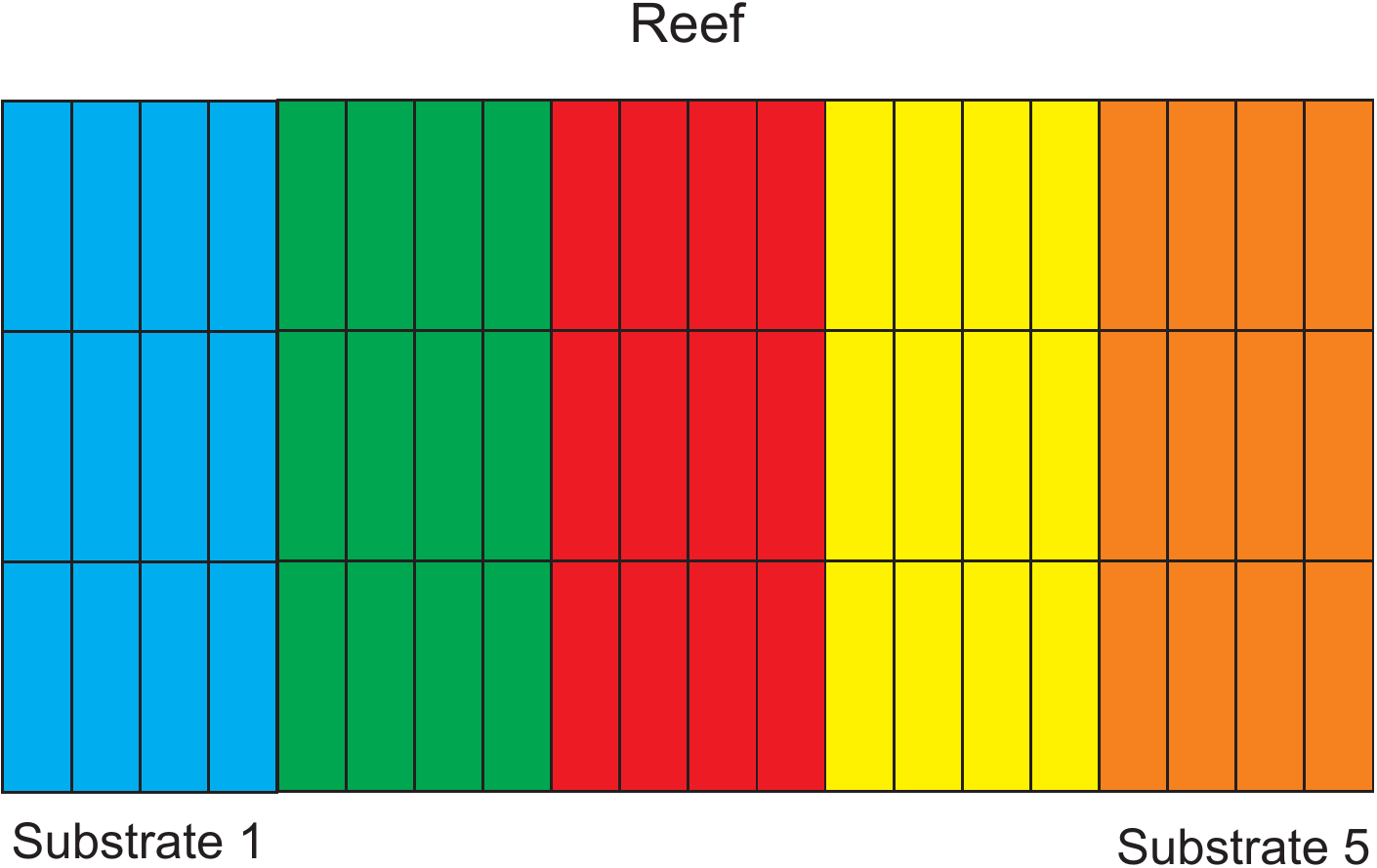}
\caption{CRO-SL reef.}
\label{fig:CROSL}
\end{figure}

The inclusion of substrate layers in the CRO can be done, in a general way, in a straightforward manner: we redefine the artificial reef considered in the CRO in such a way that each cell of the square grid $\Psi$ representing the reef is now defined by 3 indexes $(i,j,t)$, where $i$ and $j$ stand for the cell location in the grid, and index $t \in T$ defines the substrate layer, by indicating which structure (model, operator, parameter, etc.) is associated with the cell $(i,j)$. Each coral in the reef is then processed in a different way depending on the specific substrate layer in which it falls after the reproduction process. Note that this modification of the basic algorithm does not imply any change in the corals' encoding. When the CRO-LS is focused on improving the searching capabilities of the classical CRO approach, each substrate layer is defined as a different implementation of an exploration procedure. Thus, each coral will be processed in a different way in the reproduction step of the algorithm. Figure \ref{fig:CROSL} shows an example of the CRO-SL, with four different substrate layers. Each one is assigned to a different exploration process, Harmony Search based, Differential Evolution, 1-point crossover, Gaussian mutation, etc. Of course this is only an example and any other distribution of search procedures can be defined in the algorithm. In this Thesis, each substrate layer only affects to the calculation of the larvae coming from the broadcast spawning process, whereas we have considered the same brooding procedure for all the corals in the reef.

\begin{algorithm}[!ht]
   \caption{Pseudo-code for the CRO-SL algorithm}
   \label{alg:CRO}
   \begin{algorithmic}[1]
   \REQUIRE Valid values for the parameters controlling the CRO algorithm
   \ENSURE A single feasible individual with optimal value of its \emph{fitness}
   \STATE Initialize the algorithm
   \FOR{each iteration of the simulation}
   \STATE Update values of influential variables: predation probability, etc.
   \STATE Sexual reproduction processes (broadcast spawning and brooding). In the broadcast spawning phase, larvae are generated from different substrates.
   \STATE Settlement of new corals. Note that any larva can settle in any substrate independently of which substrate came from.
   \STATE Asexual reproduction process
   \STATE Predation process
   \STATE Evaluate the new population in the coral reef
   \ENDFOR
   \STATE Return the best individual (final solution) from the reef
   \end{algorithmic}
\end{algorithm}

There are some important remarks that can be done regarding the CRO-SL approach. First, note that the original CRO is a meta-heuristic based on exploitation of solutions, and leaves the specific exploration open (in the same manner as, for example, Simulated Annealing \cite{Ki83}). This way, the CRO-SL can be seen as a generalization of the original CRO, that does not modify the dynamics of the algorithm, with can be still outlined following Algorithm \ref{alg:CRO}. The only difference is the specific implementation of the broadcast spawning procedure, which now depends on the specific substrate to which the coral is associated. Second, as has been previously mentioned, the CRO-SL can be seen as a competitive co-evolution procedure. The CRO-SL is a general procedure to co-evolve different models, operators, parameter values, etc., with the only requisite that there is only one health function defined in the algorithm. Since the CRO is based on a procedure of larvae settlement which involves competition among corals, the substrate layer of the CRO-LS promotes this co-evolution process between corals, without the necessity of defining different populations. In this sense, we say that the CRO-LS makes competitive co-evolution of different searching models or patterns within one population of solutions.

\begin{figure}[!ht]
\centering
\includegraphics[width=0.70\columnwidth]{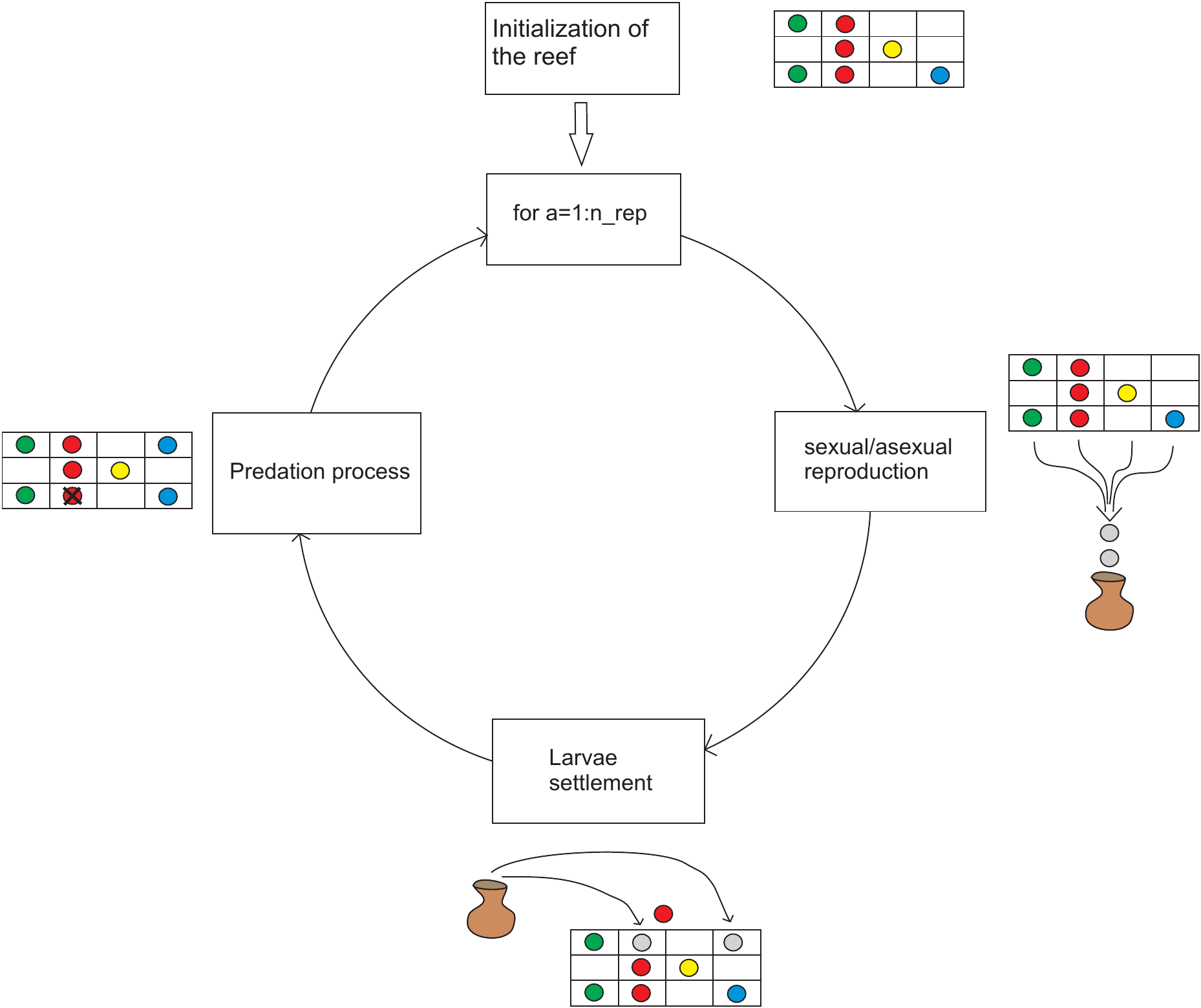}
\caption{Execution diagram about CRO-SL algorithm.}
\label{fig:CROSLDiagram}
\end{figure}

\begin{figure}[!ht]
\centering
\includegraphics[width=0.5\textwidth]{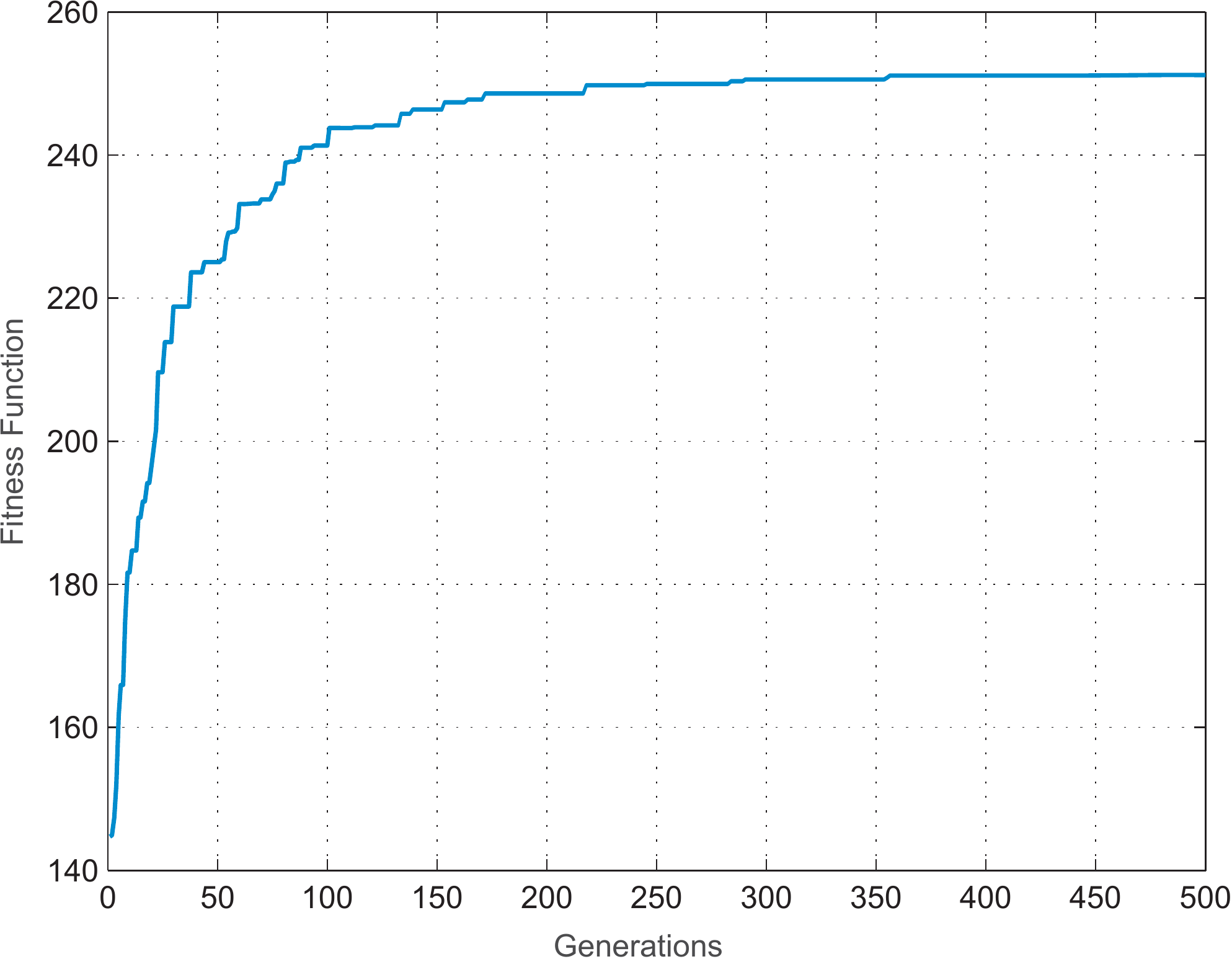}
\caption{Example of evolution of the better(maximum) fitness function of every iteration.}
\label{fig:evoMAX_ejemplo}
\end{figure}

\begin{figure}[!ht]
\centering
\includegraphics[width=0.5\textwidth]{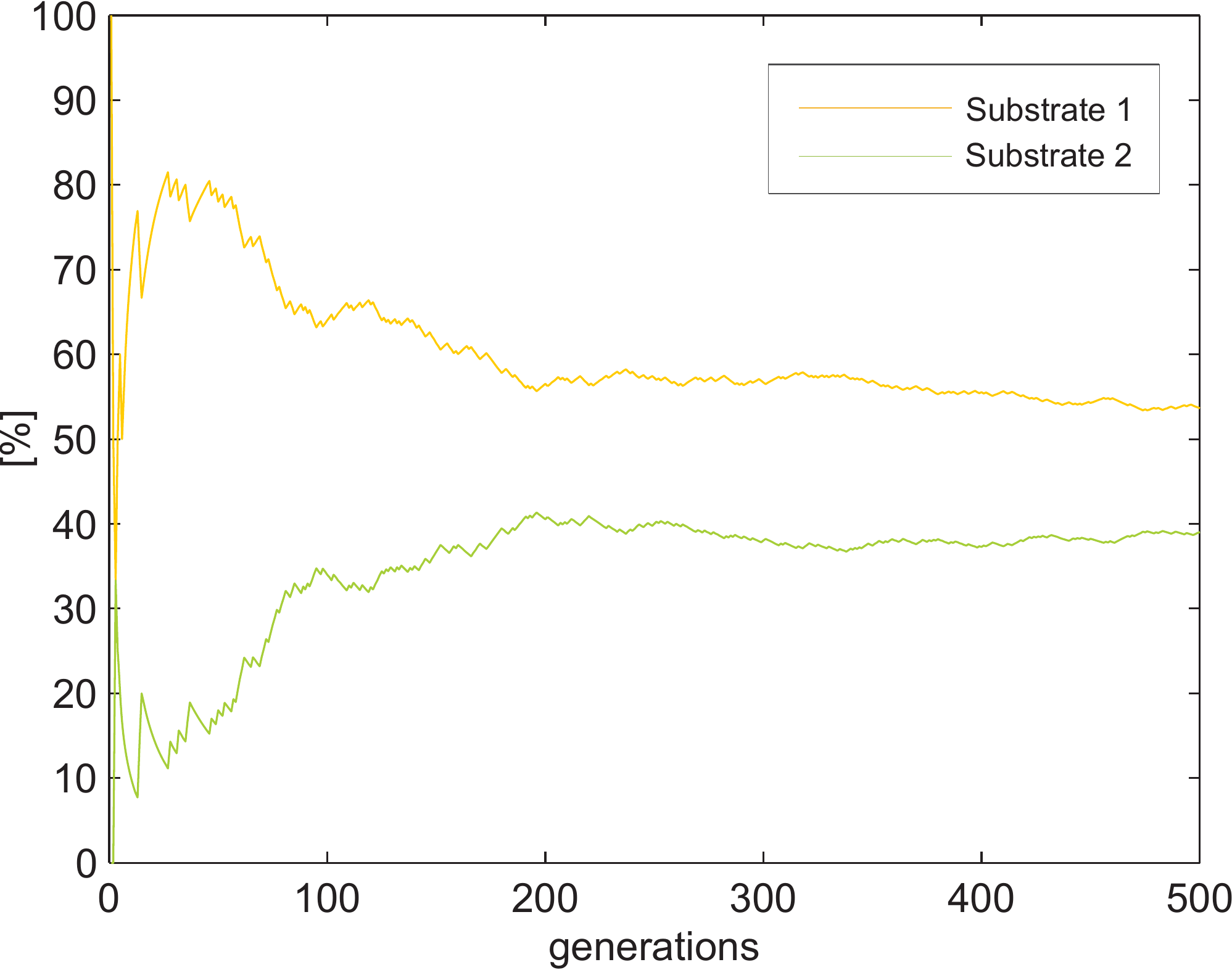}
\caption{Example of percentage about the best production of the substrates.}
\label{fig:ratioMejor_ejemplo}
\end{figure}

\begin{figure}[!ht]
\centering
\includegraphics[width=0.5\textwidth]{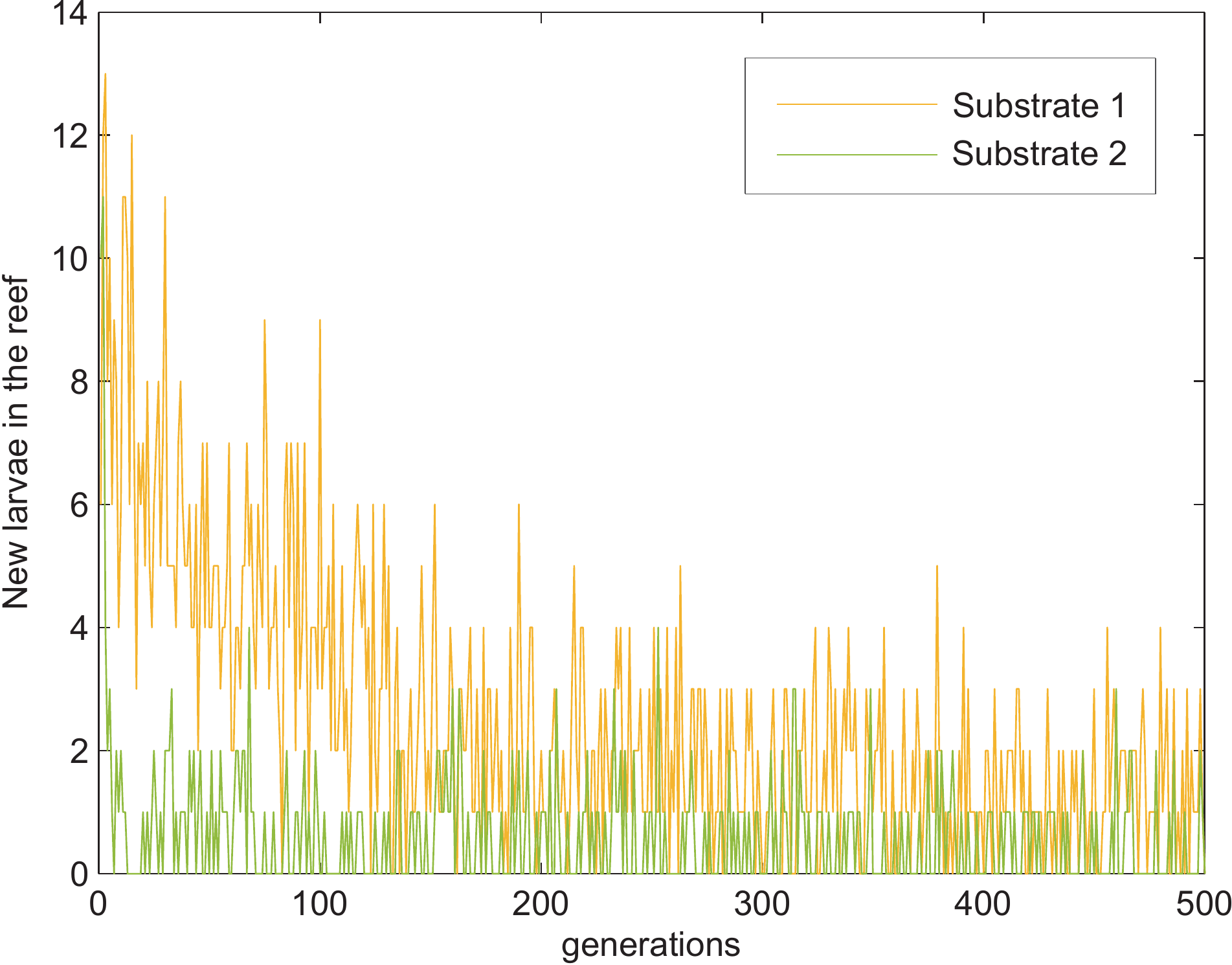}
\caption{Example of the number of larvae that enter to the reef per substrate.}
\label{fig:larvasentran_ejemplo}
\end{figure}

In the different problems that we are addressing, a series of constrains are presented and therefore, the CRO-SL algorithm will need to vary some of its characteristics as the substrates, substrates' parameters or the algorithm, the encoding or even hybridizing it with another local search or technique. Therefore, the explanation of these changes and their justifications will be made in the following chapters where the applications and the results are treated. As mentioned before, one of the main objectives is to find out which search procedures are working better in a given problem, so some graphics will be shown in each application. The basic graphic we are going to show later is the minimum fitness value evolution, which is, the minimum fitness value attained by the hole reef in every iteration (example Figure \ref{fig:evoMAX_ejemplo}). We will also use the evolution of the percentage of times that any substrate gives the best larva. Note that this larva does not have to be better than any existing coral of the reef (example Figure \ref{fig:ratioMejor_ejemplo}). This graphic does not tell us which substrate is adding larvae to the reef and that is the reason why we print the third graphical. It is a counter about how many larvae of the substrates are entering to the coral (example Figure \ref{fig:larvasentran_ejemplo}). In this example substrate 1 allocates more larvae than the second, and can be seen how in the first iterations both introduce more new corals due to in the beginning there is more holes in the reef and the population fitness is easier to be beaten.

There are some differences between the algorithm developed in \cite{Sancho_IJBIC} and the algorithm that it is presented in this Thesis. This is due to the implementation of this Thesis has been based on the experiments, so the algorithm has been adapted:

\begin{enumerate}
\item First, the reef is defined as a $M \times N$ matrix in which the corals are allocate, thus, $M$ is the maximum number of individuals and $N$ is the length of each one. So it is necessary to define a binary vector $check$ where 1 in the position $i$ means that there is a coral, and 0 that there is a hole. The number $M$ divided by the number of substrates gives the number of corals per substrate and, the position $i$ is linked to one substrate.\\
\item In the process of creating new individuals, the original CRO-SL selects a percentage of the population $F_b$ for carrying out broadcast spawning while the rest $1-F_b$ perform an asexual brooding. In this version of the CRO each individual has a probability $P_b$ of its reproduction is sexual, and a probability $1-P_b$ of it is asexual. Supposing that $F_b=97\%$, the $97\%$ of the population would be selected randomly for broadcast spawning, which is not the same if $P_b=97\%$.\\
\item Due to in the first phases of the algorithm it may converge to local minimum very fast, one of the actions taken has been avoiding a larva entering the reef if there was already a coral equal to it. This characteristic is exclusive to this version of the CRO-SL, and the idea is to maintain the diversity of the reef population.\\
\item Finally, in the original CRO the selection of the weakest individuals for depredation was made exclusively to a fraction $F_d$ of the worse health corals fraction $F_a$. So in this version $F_d$ is equal to $F_a$.\\
\end{enumerate}

\section{Substrates used in this Thesis}
In this section different mutation procedures proposed in this work will be fully described. In future application chapters the related substrates with the CRO-SL will be mentioned. Below is the definition of each substrate as well as a pseudo-code of its operation.
\begin{enumerate}
\item HS substrate: This substrate borrows the mutation procedure of Harmony Search algorithm which has been proved in many researches, resulting in a good exploratory algorithm. Harmony Search algorithm is a population-based method where every solution (musician) search the best notes (values of the codification) in order to attain the best solution (harmony). Although only the corals of the substrate mutate, the election of a new note is made on the whole population.
\begin{algorithm}[!ht]
   \caption{Pseudo-code for the HS substrate}
   \label{alg:61HS}
   \begin{algorithmic}[1]
   \FOR{Each individual of the substrate}
   \FOR{Each position of the individual}
   \STATE With probability HMCR(Harmony Memory Considering Rate) its value is replaced by that of the randomly selected individual among the entire population.
   \STATE With probability (1-HMCR) the value is selected randomly between the search space.
   \STATE With probability PAR (Pitch Adjusting Rate) the previous value is added to +/-$\delta$.
   \ENDFOR
   \ENDFOR
   \end{algorithmic}
\end{algorithm}\\
\item DE substrate: In the Differential Evolution algorithm the population mutates mixing their individuals each other to form new ones. Each coral of the substrate uses two other reef solutions to modify it self. This substrate uses the function $F(a)$ which determines the evolution factor weighting the perturbation amplitude, and where $a$ stands for the number of iteration of the algorithm. Thus, this is a linear decreasing function in which their values are included in [0.1,0.4].
    \begin{algorithm}[!ht]
   \caption{Pseudo-code for the DE substrate}
   \label{alg:61DE}
   \begin{algorithmic}[2]
   \FOR{Each individual of the substrate}
   \STATE Randomly two individuals are selected from the reef.
   \STATE The main individual is modified following the next expression: $x'=x_1+F(a)(x_2-x_3)$ forming a new larva.
   \ENDFOR
   \end{algorithmic}
\end{algorithm}\\
\item 2Px: In genetic algorithms this is the classic mutation procedure. Deep researches has shown that this mutation is the simplest but also one of the best. Two-point crossover shows a higher performance than one-point crossover, however, increase the number points doesn't ensure a better results. Obviously, the goodness of this substrate will depends on the type of the encoding and the restrictions of the search space.
    \begin{algorithm}[!ht]
   \caption{Pseudo-code for the 2Px substrate}
   \label{alg:612Px}
   \begin{algorithmic}[3]
   \FOR{Each individual of the substrate}
   \STATE Randomly one individual is selected from the reef.
   \STATE Two points of the encoding are selected also randomly. If this points were the same, one-point crossover would be performed.
   \STATE The parents mixes each other to form two new individuals.
   \ENDFOR
   \end{algorithmic}
\end{algorithm}\\
\item MPx: As mentioned, some studies appoint that the higher points, the lower performance, and can be truth but one of the advantages is that can explore the search space efficiently. This, together with the fact that other substrates can attain better individuals, it can be say that the function of this substrate is give to the algorithm some diversity, and prevent stagnancy.
    \begin{algorithm}[!ht]
   \caption{Pseudo-code for the MPx substrate}
   \label{alg:61MPx}
   \begin{algorithmic}[4]
   \FOR{Each individual of the substrate}
   \STATE Randomly one individual is selected from the reef.
   \STATE $M=10$ points of the encoding are selected also randomly.
   \STATE The parents mixes each other to form two new individuals.
   \ENDFOR
   \end{algorithmic}
\end{algorithm}\\
\item GM: In evolutionary and genetic algorithms is the most used mutation.It consists of generating random numbers with a given mean and variance and adding each one in a different position of the individual. Its average value tends to become smaller throughout the iterations in order to reduce the search space and obtain more refined solutions. However, this can induce a stagnation of the population. In this problem, the CRO will vary linearly the value of the average $\sigma$ as the number of executions increases, so it goes decreasing from 0.4 to 0.1.
    \begin{algorithm}[!ht]
   \caption{Pseudo-code for the Gaussian Mutation substrate}
   \label{alg:61GM}
   \begin{algorithmic}[5]
   \FOR{Each individual of the substrate}
   \STATE Generate the gaussian vector.
   \STATE Add it to the individual.
   \ENDFOR
   \end{algorithmic}
\end{algorithm}\\
\item SAbM: SAbM is a search operator proposed in \cite{Salcedo16b}, specifically designed to improve the searching capabilities of MHs by using fractal geometric patterns. Specifically, it is designed to generate structures of non-linear dynamical systems with chaotic behavior~\cite{Grassberger1983}. Overall, these kinds of fractal structures can be generated by means of the general two-dimensional quadratic map. This quadratic map allows generating very different chaotic-behavior attractors with a reduced number of input parameters. In order to introduce a chaotic mutation in a solution using SA, we followed the procedure introduced in \cite{Salcedo16b}. Figure~\ref{fig:Strange_Attractors} shows some examples of strange attractors used in this operator for the proposed application. Figure \ref{fig:Strange_Attractors} shows some examples of strange attractors, regarding a simulation with values of parameters $a_1$-$a_{12}$ in the range $[-1.2,1.2]$. This range allows the generation of an extremely large variety of attractors (over $25^{10}$) with different characteristics: chaotic, intermittent or convergent to a periodic orbit.
    \begin{figure}[!ht]
    \centering
        \subfigure[]{\includegraphics[width=0.4\columnwidth]{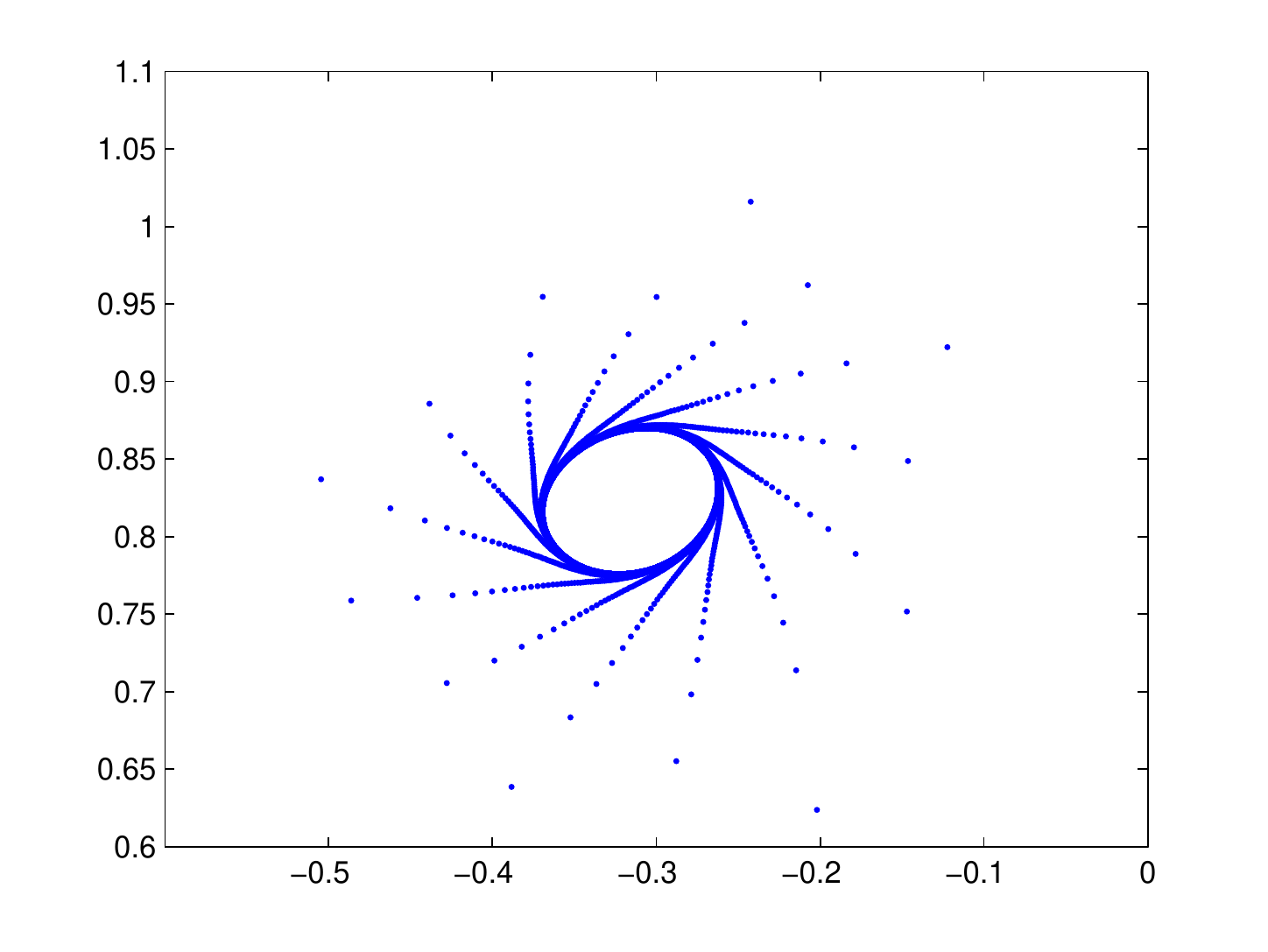}}
        \subfigure[]{\includegraphics[width=0.4\columnwidth]{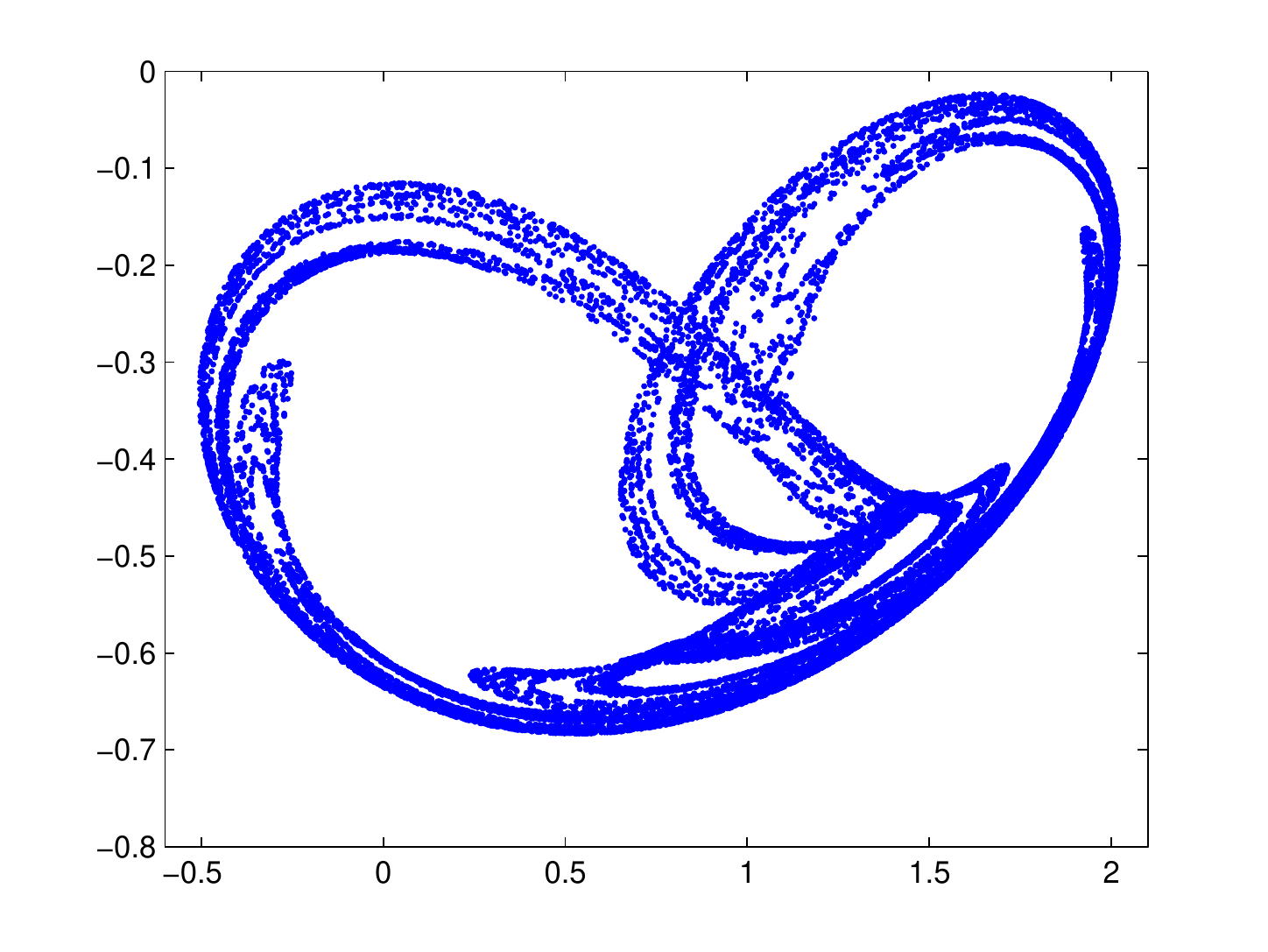}}\\
        \subfigure[]{\includegraphics[width=0.4\columnwidth]{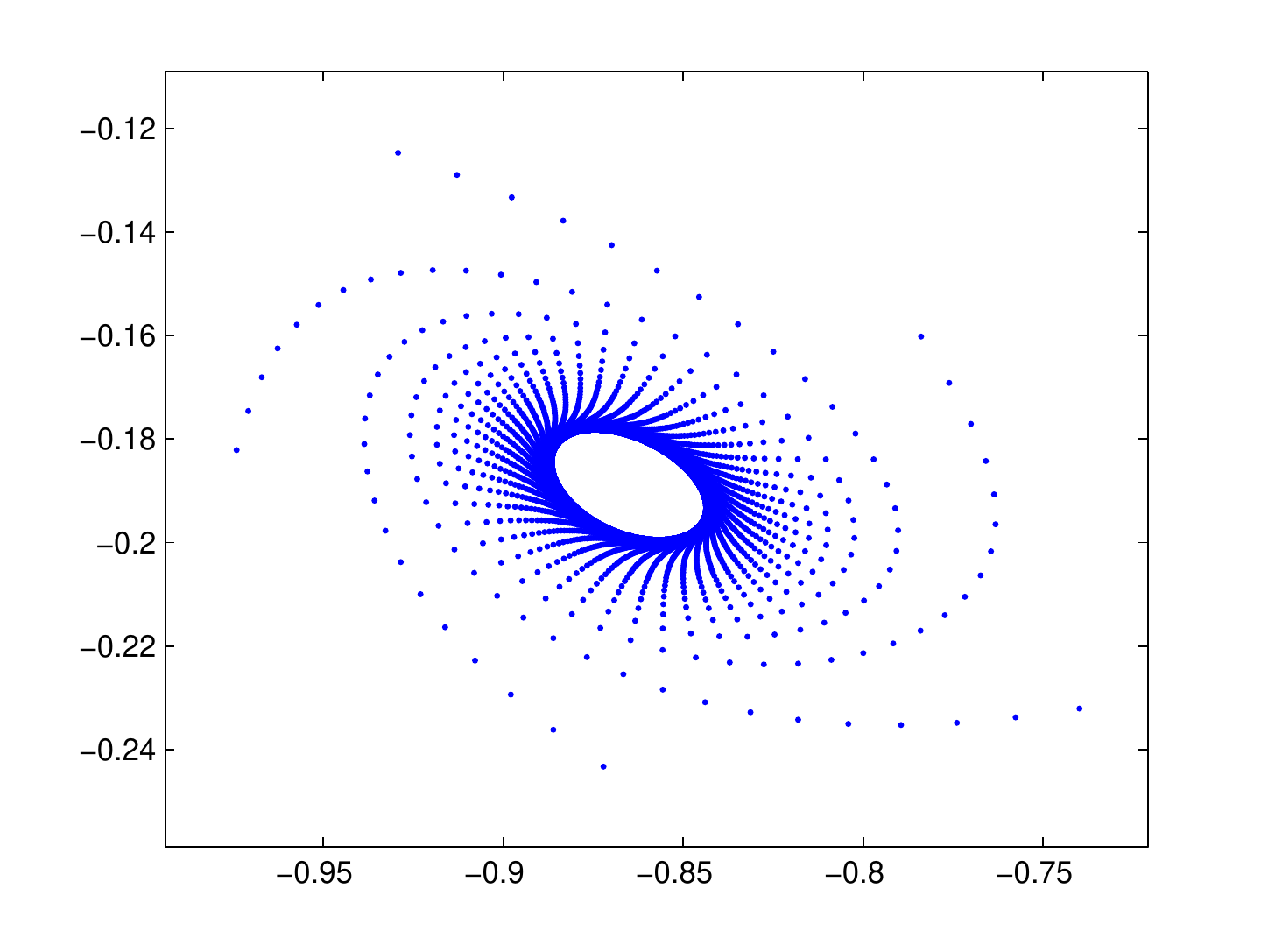}}
        \subfigure[]{\includegraphics[width=0.4\columnwidth]{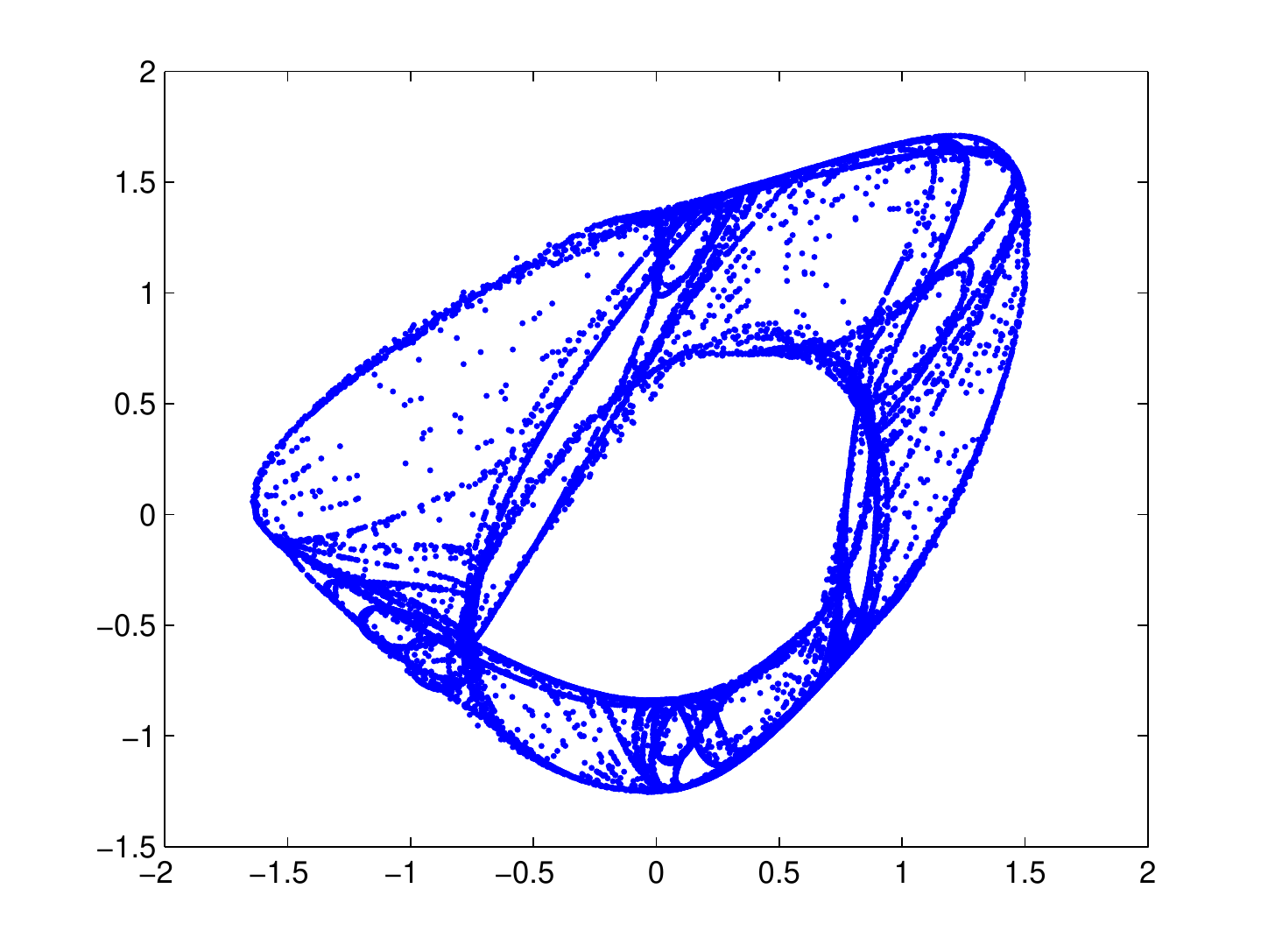}}\\
    \caption{Examples of four Strange Attractors in the phase space (x vs. y), used in the SA substrate.}
    \label{fig:Strange_Attractors}
    \end{figure}
    \begin{algorithm}[!ht]
   \caption{Pseudo-code for the SAbM substrate}
   \label{alg:61SAbM}
   \begin{algorithmic}[6]
   \FOR{Each individual of the substrate}
   \STATE Get a random strange attractor.
   \STATE Select randomly a number of iterations $S$ for the Strange Attractor function. $S$ have a range $[2,5000]$.
   \STATE Obtain $S$ pair of points in $X$ and $Y$ coordinates from the Strange Attractor function and save them in the matrix $Z$.
   \STATE Permute randomly the $S$ value and get as many points as the length of the individual $N$ in vector $S_{per}$.
   \STATE Get a random vector $XorY$ for selecting $X$ or $Y$ coordinates.
   \STATE Get a random vector $NorP$ for selecting the sign of each position.
   \STATE Get the new individual as $x'=x+Z(XorY,S_{per}(1:N)).*NorP$.
   \ENDFOR
   \end{algorithmic}
\end{algorithm}\\
\end{enumerate}

\section{Objectives and main contributions of this work}
There are many types of meta-heuristics that are able to effectively address complex problems such as those this Thesis is focused on. The proposed CRO-SL algorithm expands the limits of the basic CRO and makes it more efficient at the same time as versatile. The CRO-SL algorithm gives the possibility of jointly exploring the performance of different meta-heuristics within a single population. The application of the proposed algorithm to the following problems already presents a remarkable contribution, partly due to the novelty of the algorithm itself, but also due to certain factors that are stated below:

\begin{itemize}
\item Battery Scheduling Optimization Problem: In spite of there are many researches that address optimization problem in MG's with heuristics, this work takes into account Spanish regulation for energy prices and different load and generation profiles through the year.\\
\item TMD Design and Location Problem: In this work, the possibility of addressing a design problem for canceling vibrations in structures has been provided, which allows freedom in the positioning of TMD's. In addition, the proposed algorithm has managed to address a problem with high computational time, on an objective function that is non-linear.\\
\item MIMO-AVC Design: In terms of structural design, the main contribution of this work is to have the possibility of designing a multi-input multi-output (MIMO) AVC for complex floor structures with several closely frequency space vibration modes, where the number of test points and sensor/actuator pairs is not a problem to obtain a global optimum solution in a affordable computation time. This Thesis also shows that a MIMO-AVC improves substantially the vibration reduction compared with a single-input and single output AVC for the proposed application example, which is a real complex floor structure.\\
\item Meander-line IFA Antenna design: The novelty lies in the physical implementation on felt, fulfilling the requirements of design widely and obtaining two bands of transmission in a single antenna. In this case the fitness value of each coral (representing a given antenna design) has been carried out by hybridizing the CRO-SL with a simulated package: The CST Microwave Studio.\\
\end{itemize}

\section{Structure of the thesis}
This chapter will be finished with the structure of the following parts of this Thesis. The structure of this Thesis is mainly focused on applications, so the optimization problems discussed in this work represent the guiding thread of the Thesis. At the same time, the algorithmic part of the work is focused on the CRO-SL algorithm and how to improve it for the different optimization problems tackled. Having these points in mind we have elaborated this document with the following structure: in a first introductive chapter we will fully describe the main ideas behind the CRO-SL approach, and how this algorithm was obtained from the original CRO meta-heuristic. Then, we give details on the specific implementation of the algorithm in each chapter, devoted to specific applications. Also, we will describe the state-of-the-art referent to each problem at the beginning of each application chapter. Note, however, that we have opted by a common bibliography for the whole document. The applications chapters are structured as follows:

\begin{itemize}
\item In Chapter \ref{cap:scheduling}, the CRO-SL algorithm will be applied to battery scheduling problem. So in the first part will be provided the problem definition and a state of the art of the application. Then will be introduced the specific characteristics of the algorithm for this problem, which cover the substrates and their parameters. Finally the results and and its analysis will close this first application chapter.\\

\item In Chapter \ref{cap:tmds}, the CRO-SL will be used to optimize the solutions in a problem of control of vibrations in structures. First the state of the art of the application and the definition of the problem will be given. Then the substrates and characteristic parameters of the algorithm proposed for this application will be detailed, and finally the results and their corresponding conclusions will be gathered. \\

\item Chapter \ref{cap:avcs} analogously collects the application of the CRO-SL algorithm to a problem of active vibration cancellation. Specifically, two active devices must be placed over a floor to try to minimize the vibrations that humans produce when we walk or run in a building. Results in a real-world case will be shown and discussed.\\

\item In Chapter \ref{cap:antena} an antenna design for radio-frequency identification will be tackled. First the state of the art and the definition of the problem will be given. Then the characteristics of the CRO-SL for this problem will be detailed. The simulation of the reflection coefficient of the antenna is made by CST Studio Software. Finally the physical implementation of the antenna on felt and his radiation pattern will be shown and discussed.\\

\item In the final chapter, the main conclusions of this research will be summarized, and future lines of research will be outlined.\\

\end{itemize}

\part{Engineering Applications}\label{part:aplicacion1}
\chapter{Optimal Battery Scheduling Optimization in Micro-Grids}\label{cap:scheduling}

\section{Introduction and state of the art}
MicroGrids (MGs) are defined as the coordinated operation and control of distributed energy resources, involving different technologies, together with controllable or non-controllable loads and energy storage systems, operating connected to the utility grid and capable of islanding \cite{Jiayi08,Berry10}. MGs are often considered as the next evolution of the current electricity distribution systems, since they allow a high penetration of low emissions generation, a reduction of electricity transportation lines losses, the reuse of waste heat to service thermal loads, etc. MGs offer other important advantages such as a larger robustness against extreme weather events or attacks, an improved reactive power support and a save in deployment time of the systems \cite{Jiayi08}.

In its basic configuration, a MG is a medium or low voltage network with distributed energy generators, multiple electrical loads and, optionally, energy storage devices. MGs are connected to the utility grid via the Point of Common Coupling, however, one of the MGs' characteristics is that they have the  capability to disconnect from the utility grid and operate in islanding mode in case of faults in the upstream network. Constraints about this islanded operation in regulations, on the basis of security and control of the grid, have limited the deployment of MGs so far. However, the development of fast and safe power electronic inverters and the consequent regulation revision may increase the number of MGs in operation. Like Super Grids at transmission level or Virtual Power Plants at software and communication one, MGs can be related with the global and opened concept of Smart Grids as one of the multiple and complementary options to develop the electricity network of the future \cite{Asmus10}. As stated by the EU SmartGrid Platform, Smart Grid is an ``electricity network that can intelligently integrate the actions of all users --generators, consumers and those that do both-- in order to efficiently deliver sustainable, economic and secure electricity supply''. Deployment of decentralized architectures, improve management and control techniques, efficiently integrate intermittent generation systems and enhance the role of the demand side are mentioned as key challenges for Smart Grids. MGs have to cope with these issues at the distribution level.

The research on different aspects of MGs, both theoretical (MG design and planning, control, renewable energies integration, etc.), and practical (rural electrification and stand alone systems, energy management, real application in smart cities) has been massive in the last few years \cite{Asmus10}. Some of these works apply advanced computational methodologies, such as neural computation \cite{Xu10}, evolutionary computation \cite{Bajpai12} or computational intelligence techniques in order to obtain good quality solutions when tackling difficult problems related to MGs. For example, one of the key issues concerning MG technology deployment is related to the integration of distributed energy resources into distribution networks. In the last few years there have been many different works dealing with optimal location and sizing of distributed generation in MG \cite{Moradi12}-\cite{Ghosh10}. Other aspects of MG design such as topology design \cite{Zeng13}, control \cite{Planas13,AlSaedi13}, load prediction \cite{Haesen05}, energy storage systems \cite{Tan13} or hybrid fuel/battery back-up systems \cite{Moghaddam11} have been recently studied.

Regarding the use of meta-heuristic optimization techniques, such as GA, EA, PSO, etc., these technique have been profusely used to improve different aspects related to the design or operation. In \cite{Yang15}, the design of a Distributed Energy Resource system coupled with cooling, heating, and power distribution networks in Guangzhou City (China) is studied. Several scenarios are analyzed considering that electricity requirements are served directly from the main utility grid or the operation is determined following various constraints, and a mixed integer linear programming model is constructed to minimize total annual cost. In \cite{Zhao14} a stand-alone MG is sized using a GA-based method with multiple objectives such as life-cycle cost, renewable energy source penetration and pollutant emissions. In \cite{Zhao16} a multi-objective {\em fruit fly} optimization algorithm as used in another stand-alone MG design, including PV-wind and diesel generation, and batteries as back-up systems. Optimal MG structure design problems using EAs have also been analyzed recently \cite{Moradi12,Doagou13,Taher11}. Other works have dealt with the heuristic optimization of the different operational problems within MGs \cite{Moghaddam11,Mallol15,Kusakana15}. For example, an expert multi-objective adaptive modified Particle Swarm Optimization algorithm is presented in \cite{Moghaddam11} for optimal operation of a MG with renewable energy sources together with some back-up elements. The authors simultaneously minimize the total operating cost and the net emission improving the results obtained with a genetic algorithm or a PSO. In \cite{Severini13} an approach based on different soft-computing techniques for energy demand scheduling at home in MGs is proposed, mixing different approaches such as genetic algorithms for optimization and neural networks for prediction. In \cite{Mallol15} a hyper-heuristic approach is proposed for a problem of optimal EES scheduling in MGs, taking into account renewable energy sources and different types of loads. In \cite{Mallol16} an EA has been proposed for the joint design of a MG structure and operation (energy storage systems scheduling), in a context of renewable energy generation (photovoltaic and wind) within the MG. Specifically, a two-step approach was proposed, in such a way that in the first step, the structure of the MG was designed, and in the second stage of the algorithm, the operation of the MG (with the structure set in the first step), was tackled. Good results of this approach were reported. In \cite{Kusakana15} a photovoltaic (PV)-Diesel-Battery hybrid system is considered, and the optimal scheduling of the energy production at any given time that minimizes the diesel generator (DG) fuel expenses is addressed. For this purpose, the author proposes two strategies: ``on/off'' control of the DG (leading to a smooth control of the the PV and battery bank to feed the load) and continuous operation (when the DG is ``on'' most of the time and its output is continuously controlled, depending on the demand, to minimize fuel usage), concluding that the latter achieves more fuel saving.

This chapter is focused on the management of an energy storage device in MGs within an environment of variable electricity prices, taking into account MG with renewable generation. Specifically, we consider the problem of energy storage system (battery) scheduling in a MG for modifying the main grid consumption profile, in such a way that the electricity cost imported to the MG is minimized. We refer to this problem as BSOP and the CRO-SL algorithm will be considered. A case study based on the Spanish regulation for energy prices is considered and discussed in the experimental part of the paper, where we will show the good performance of the CRO-SL in the BSOP tacked, taking into account different load and generation profiles through the year. \\

\section{Problem definition}\label{61ProblemDefinition}

\begin{figure}[!ht]
\begin{center}
\includegraphics[draft=false,angle=0,width=8cm]{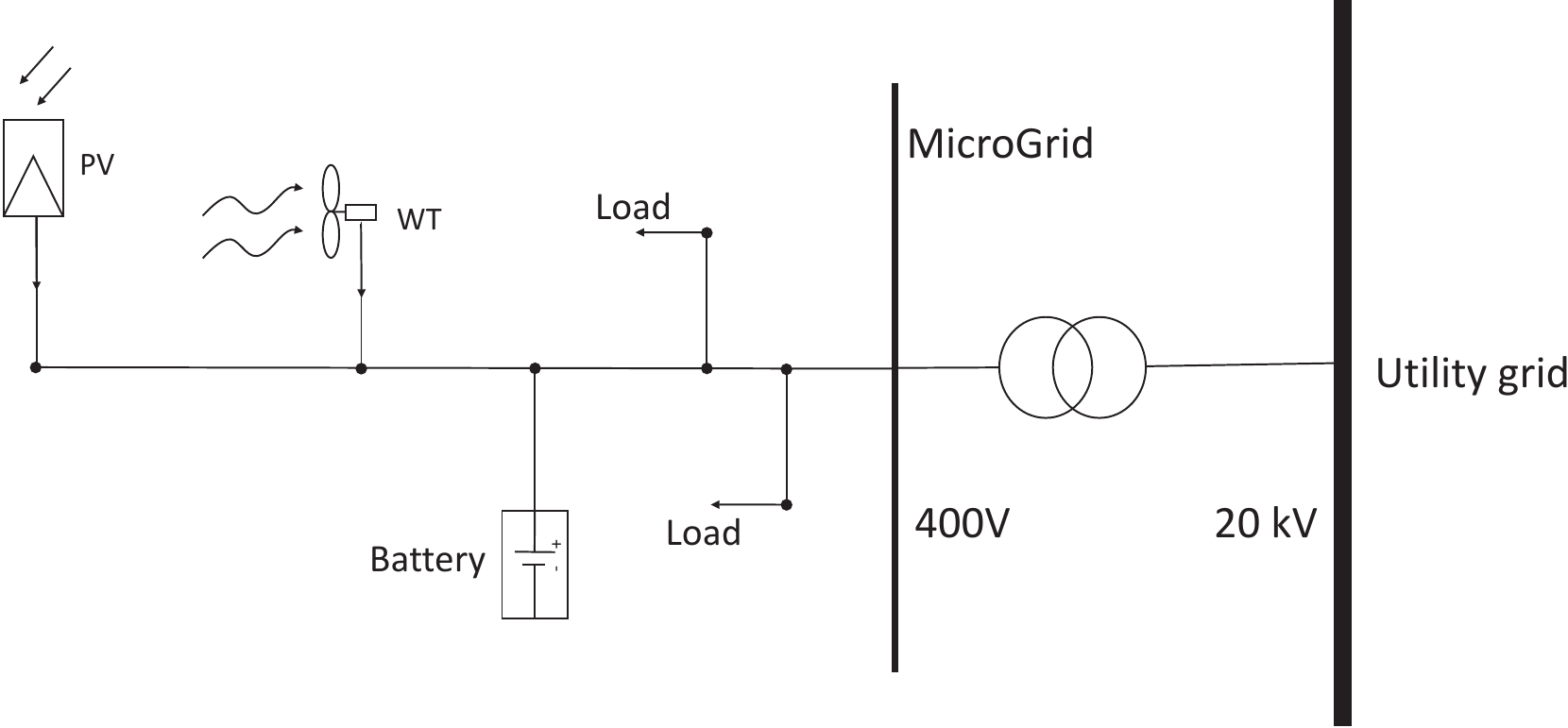}
\caption{\label{61Estructura_MG} MicroGrid structure used in this work.}
\end{center}
\end{figure}

Let it be considered, without loss of generality, a model of MG consisting of different loads, renewable energy generators and a battery (energy storage device) connected together to the main utility grid at the same point (see Figure \ref{61Estructura_MG} for reference). We will focus on establishing the optimal battery scheduling for a given period ($T$ hours). For that, it is considered typical (or predicted) wind generation $\mathcal{W}=\{\mathcal{W}_1, \ldots, \mathcal{W}_T\}$ and photovoltaic generation $\mathcal{F}=\{\mathcal{F}_1, \ldots, \mathcal{F}_T\}$ annual profiles. It is also considered two different load profiles $\mathcal{L}^1=\{\mathcal{L}^1_1, \ldots, \mathcal{L}^1_T\}$ corresponding to a residential consumption, and $\mathcal{L}^2=\{\mathcal{L}^2_1, \ldots, \mathcal{L}^2_T\}$, that corresponds to an industrial consumption. The battery scheduling is defined by a vector $\mathcal{B}$ that stands for the battery charging ($\mathcal{B} > 0$) or discharging ($\mathcal{B} < 0$) power. Note that if an annual scheduling is considered, then $\mathcal{B}$ is a $8760$-length vector, whereas in a weekly profile the encoding of $\mathcal{B}$ would be a $168$-length vector. A vector $\mathcal{P}$ is defined as the power exchanged between the MG and the main grid, considering the effect of all profiles described above acting together at the same node, as follows:

 \begin{equation}
 \mathcal{P}=\mathcal{L}^1+\mathcal{L}^2-\mathcal{F}-\mathcal{W}+\mathcal{B}
 \end{equation}

In this work, battery power values are limited, and minimum \% SOC (State of Charge) value is set to 20\% to achieve longer battery life time (according to a simplified battery model). The simplest way for handling the battery is named here as {\em Deterministic} battery use. In that case, the battery is charged (with the maximum possible power) every period of time in which the generation is larger than the load's demand. If this power is within battery limits, there is no energy exported to the main grid. On the other hand, for the periods of time in which the load's demand is larger than the generation, the battery is discharged with the maximum possible power, in such a way that it avoids power exportation to the main grid.

Finally, an objective function $g(\mathcal{B},\mathcal{W},\mathcal{F},\mathcal{L}^1,\mathcal{L}^2)$ is considered, where $g$ stands for the total price paid for the electricity that the MG consumes from the main grid. Specifically, we have defined $g$ as the electricity cost. This cost adds up two terms: an energy term (ET, the price for the energy consumption), and a power term (PT, the price for the availability of electric energy at our site):

 \begin{equation}\label{61objective_function}
 g=ET+PT
 \end{equation}

In this work it has been recreated the Spanish scenario for SMEs \cite{RD01}. Precisely, it has been implemented access tariff 3.1 for high voltage and power supplies up to 450kW. This scenario specifies three access tariff periods: $P_1$ (corresponding to high-priced hours, and with a duration of 4 hours), $P_2$ (mean-priced hours, and a duration of 12 hours), and $P_3$ (low-priced hours, and a duration of 8 hours). This means that the energy term is obtained as $ET=\sum_{j=1}^3 \beta_j\cdot E_j$, where $\beta_j$ is the access tariff price for the $P_j$ period of time, and $E_j$ is the energy consumption during the $P_j$ period. Regarding the power term, and also for the Spanish scenario, an estimation of the maximum's power consumption for each of the three access tariff periods, $HP_j$ (Hired Power in period $P_j$), has to be specified when signing the contract. Clients have to be meticulous when specifying and agreeing the $HP_j$, as the Electric Company will penalize them when their power consumption exceeds the agreed $HP_j$, and benefit them when their power consumption is limited to a certain percentage of that $HP_j$.
$PT$ is obtained as $PT=\sum_{j=1}^3 \alpha_j\cdot IP_j$, where $\alpha_j$ is the power term price, and $IP_j$ is the invoiced power during the $P_j$ period. According to the above-mentioned benefit/penalty policy, $IP_j$ may have three different values, depending on the maximum value of the power consumed in period $j$, $\mathcal{M}$:

\begin{equation}
IP_{j}=\left\{
\begin{array}{lc}
HP_j&if~~0.85\cdot HP_j < \mathcal{M} < 1.05\cdot HP_j\\
0.85\cdot HP_j&if ~~\mathcal{M} < 0.85\cdot HP_j\\
\mathcal{M} + 2\cdot \lbrack \mathcal{M} - HP_j \rbrack&if ~~\mathcal{M} > 1.05\cdot HP_j\\
\end{array}
\right.
\end{equation}\\
Where $\mathcal{M}=\max (PC_j)$ and $PC_j$ is the Power Consumption during the $P_j$ period. In this scenario, it is obvious that the client needs to avoid the penalty region, and to restrain power consumption within the other two regions.

Mathematically the problem consists of obtaining the optimal battery scheduling $\mathcal{B}^*$ that produces the lowest value of $g(\mathcal{P})$ given $\mathcal{L}^1$, $\mathcal{L}^2$, $\mathcal{W}$ and $\mathcal{F}$.

 \vspace{0.3cm}
 \noindent find $\mathcal{B}^*$ such that

 \begin{equation}
 \min g({\mathcal{P}})
 \end{equation}
 given $\mathcal{L}^1$, $\mathcal{L}^2$, $\mathcal{W}$ and $\mathcal{F}$.\\

\section{The CRO substrates definition and main varieties}
For the implementation of the algorithm, five substrates were developed based on the HS \cite{Ge01} and DE \cite{St97} algorithms, and on the mutation forms two-points crossover, multi-point crossover \cite{Ei03} and GM \cite{Ya99}. Each position of the coral has associated one of these forms of mutation, therefore each coral that is found in it will mutate in that way. Each coral remains in its place and the larva that it produces is collected along with the rest, until it reaches the phase of settlement. The mutation of new larvae will depend on where they fall dawn and if their competitors are weaker.
\begin{enumerate}
\item HS: Mutation from the Harmony Search algorithm with a $\delta$ value linearly decreasing during the run,
      from $20$ to $5$.
\item DE: Mutation from Differential Evolution algorithm with a $F=0.6$ value linearly decreasing during the run,
      from $0.4$ to $0.1$.
\item 2Px: Classical 2-points crossover.
\item GM: Gaussian Mutation, with a $\sigma$ value linearly decreasing during the run,
      from $20$ to $5$.
\item MPx: Multi-points crossover ($M=10$).
\end{enumerate}

AS it will be shown later, the population starts with a good solution provided by a deterministic method. The population is composed by the deterministic solution, some individuals got by add to this solution a random gaussian vector, and finally a set of random solutions. This solution is so hard to beat that the algorithm tends to stagnate into it. This is why one process is added to avoid the situation. It consist on a re-generation of the population when its minimum fitness value doesn't get better for a hundred iterations. Thus the new population is generated in the same way as the initial one, but substituting the deterministic solution by the local optimum solution. For keep avoiding this stagnation, a larva will not be able to enter to the reef if there is already that solution inside.
Another characteristic os this CRO-SL version is that it does not perform the asexual reproduction. This is due to experimental results have proved that it can be inefficient under stagnation conditions.\\

\section{Experiments and Results}\label{61Experiments}

In order to show the performance of the CRO-SL proposed for the BSOP, we have carried out a number of experiments in a MG equipped with micro-wind and micro-photovoltaic generation. We consider two different loads, one of them stands for a residential profile (standard profile published by the Spanish system operator, REE, \cite{REE_perfil}) characterized by an energy demand of 162500 $kWh/year$ \cite{BOE_casa}, that corresponds to 50 typical homes in Spain, with 3250 $kWh/year$ each ($\mathcal{L}^{1}$). Second, we consider an industrial consumption profile ($\mathcal{L}^{2}$), normalized in such a way that the annual energy consumption is 200000 $kWh$. Regarding the generation profiles, we have considered a 100 $kW$ photovoltaic generator ($\mathcal{F}$) which provides 165000 $kWh/year$ and a 100 $kW$ wind power generator ($\mathcal{W}$) providing 140000 $kWh/year$. According to \cite{Velik13} we have considered a 300 $kWh$ capacity battery that, in average, represents 1/3 of the daily energy demand. However, note that if we consider the energy provided by the generators, those 300 $kWh$ would represent the energy consumed from the main grid during two days. Regarding $\alpha$ parameters of the objective function, we have considered $\alpha=[59.1735,36.4907,8.3677]$ \emph{euro/kW}, values periodically published by the Spanish Ministry of Industry, Energy and Tourism (\cite{ref_Alfas}). The values of the $\beta$ parameters contemplated are $\beta=[0.1044496,0.089868,0.065655]$~\emph{euro/kW}, published on a yearly basis by the electric companies (e.g. values valid for Endesa Energ\'ia's clients can be found at \cite{ref_Betas}). Finally, the values considered for $HP_j$ are $HP_j=[72,66,58]$ $kW$, corresponding to the maximum power demanded in each period $PC_j$ obtained with the deterministic solution. In this work we have considered hourly-defined generation and consumption profiles for the whole year (52 weeks). However, the length of any individual would be 8760 and the performance of the algorithm would be very slow. This is why one week per season has been selected, leading to 168-length individuals. These data have been applied to three different scenarios: 1) No battery use, 2) Deterministic battery use (as explained in Section \ref{61ProblemDefinition}), and 3) Battery scheduling optimization (proposed CRO-SL). The first two are baseline cases for comparison purposes, while the latter represents our proposed methodology. Regarding the CRO-SL implementation, we have considered a complete reef of size 120, divided into 5 substrates: DE, HS, GM, 2Px and MPx. The specific parameter's values for each search algorithm are given in Table \ref{tab:61parameters}.

\begin{table}[!ht]
    \centering
\caption{Parameters values used in the implementation of the CRO-SL.}
\label{tab:61parameters}
\begin{tabular}{llr}
\hline
Parameter & Description & value\\
\hline
Reef & Reef size & 200\\
$n_T$& Max number of iterations& 50000\\
$F_b$ & Frequency of broadcast spawning & 97\%\\
$\mathcal{N}_{att}$ & Number of tries of larvae settlement & 3\\
$F_d$ & Fraction of corals for depredation & 40\%\\
$P_d$ & Probability of depredation & 1\%\\
$T$ & Number of substrates & 5\\
$F$ & Parameter $F$ for DE substrate & $0.1-0.4$ linear\\
$HMCR$ & HMCR probability in the HS substrate & 0.9\\
$PAR$ & PAR probability in the HS substrate & 0.2\\
\hline
\end{tabular}
\end{table}

\begin{figure}[!ht]
\begin{center}
\includegraphics[draft=false,angle=0,width=12cm]{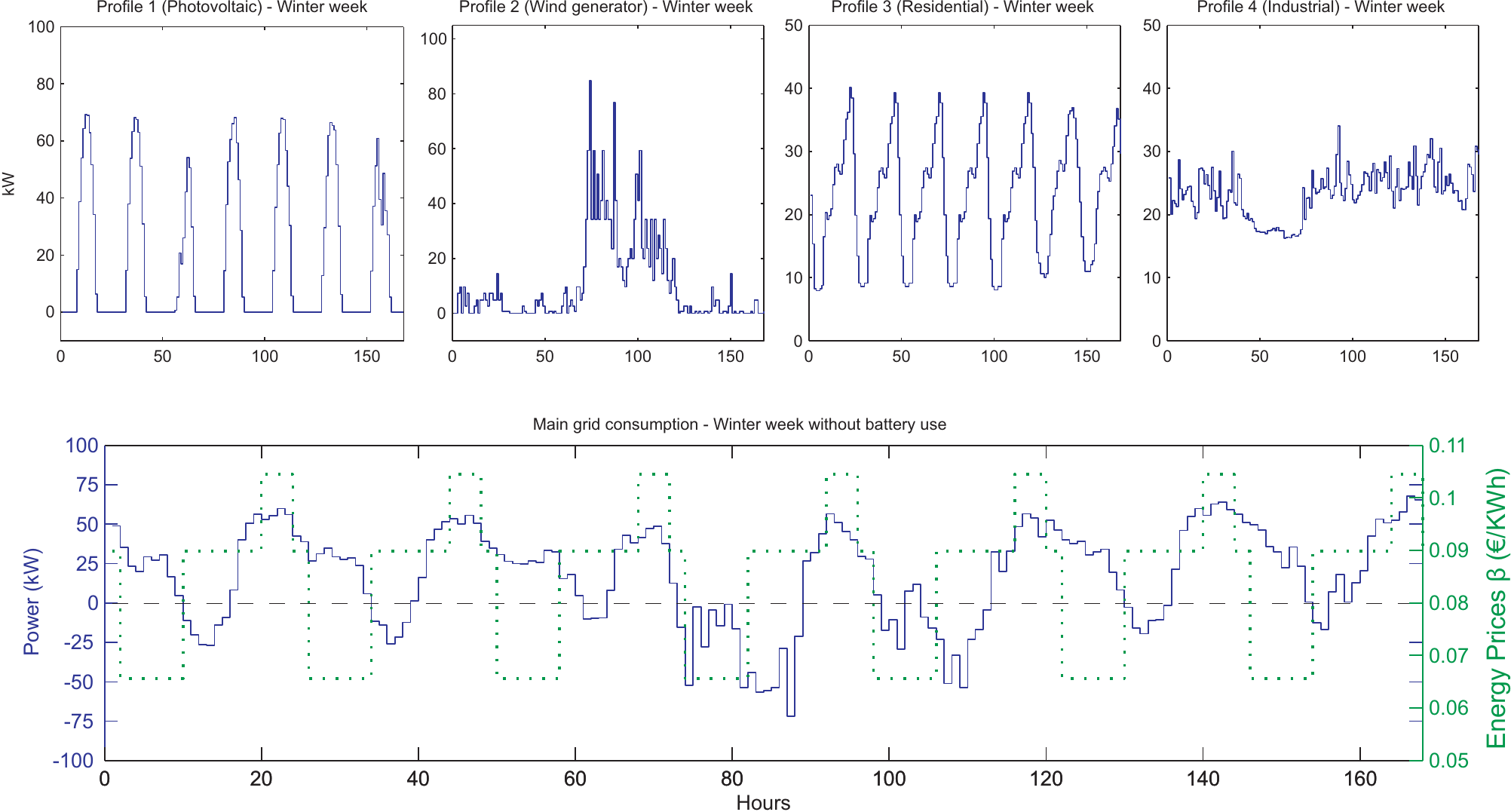}
\caption{\label{61Perfiles_invierno} Generation and demand profiles considered and consumption from the main grid ($\mathcal{P}$) occurred in the MG without the use of any battery in the system (baseline algorithm for comparison purposes, $\mathcal{B}=0$). Dotted green line stands for the values of $\beta$.}
\end{center}
\end{figure}

Figure \ref{61Perfiles_invierno} shows the generation profiles considered during winter week (wind and photovoltaic), as well as the load profiles (residential and industrial). Finally, the main grid consumption ($\mathcal{P}$) without battery, obtained by the addition of the above mentioned profiles, can be seen in this figure. Table \ref{tab:61Tabla_resultados} presents the results obtained for a randomly selected week out of every season (winter, spring, summer, and autumn) to consider different consumptions and generations. In all cases, improvements obtained with the CRO-SL over the case of non-battery use and deterministic scheduling for the battery are also displayed in the table. Note how the CRO-SL is able to obtain excellent optimization results of the battery scheduling, when comparing with the deterministic use of the battery, and of course with the case of not installing a battery in the MG.

\begin{table}[!ht]
\renewcommand{\arraystretch}{1.3}
\caption{BSOP results for four randomly selected weeks, each corresponding to one different season.}
\vspace{5mm}
\label{tab:61Tabla_resultados}
\centering
\begin{tabular}{lccccccc}
\hline
$ $ & $PT$ & $ET$ & Cost/\em{fitness} & \multicolumn{2} {c} {Improvement over}  \\
 $ $ & (euro) & (euro) & (euro)  &  No battery $ $ $ $  & Deterministic\\
\hline
\multicolumn{6} {l}{\em{Winter week.}} \\
\hline			
No battery $ $ $ $  &137.20 $ $ $ $  & 366.88 & 504.09 & 0.00\% & --\\
Deterministic $ $ $ $  &137.20$ $ $ $  & 290.48 & 427.68 & 15.16\% & 0.00\%\\
Proposed CRO-SL $ $ $ $  &118.02$ $ $ $  & 261.29 & 379.31 & 24.75\% &11.31\%\\
\hline
\multicolumn{6} {l}{\em{Spring week.}} \\
\hline	
No battery $ $ $ $  &116.62$ $ $ $  &140.05&256.67&0.00\%&--\\
Deterministic $ $ $ $  &116.62$ $ $ $  &50.76& 167.39 &34.79\%&0.00\%\\
Proposed CRO-SL $ $ $ $  &116.62$ $ $ $  &49.04& 165.66 &35.46\%&1.03\%\\
\hline
\multicolumn{6} {l}{\em{Summer week.}} \\
\hline
No battery $ $ $ $  &124.95 $ $ $ $  &313.52&438.46&0.00\%&--\\
Deterministic $ $ $ $  &124.95$ $ $ $  &234.38&359.33&18.05\%&0.00\%\\
Proposed CRO-SL $ $ $ $  &118.01$ $ $ $  &217.31 & 335.32 &23.52\%&6.68\%\\
\hline
\multicolumn{6} {l}{\em{Autumn week.}} \\
\hline
No battery $ $ $ $  &128.88$ $ $ $  &316.98&445.86&0.00\%&--\\
Deterministic $ $ $ $  &116.62$ $ $ $  &211.44&328.06&26.42\%&0.00\%\\
Proposed CRO-SL $ $ $ $  &116.62$ $ $ $  & 191.93 & 308.55 &30.8\%&5.95\%\\
\hline
\end{tabular}
\end{table}

\begin{figure}[!ht]
\begin{center}
\includegraphics[draft=false,angle=0,width=10cm]{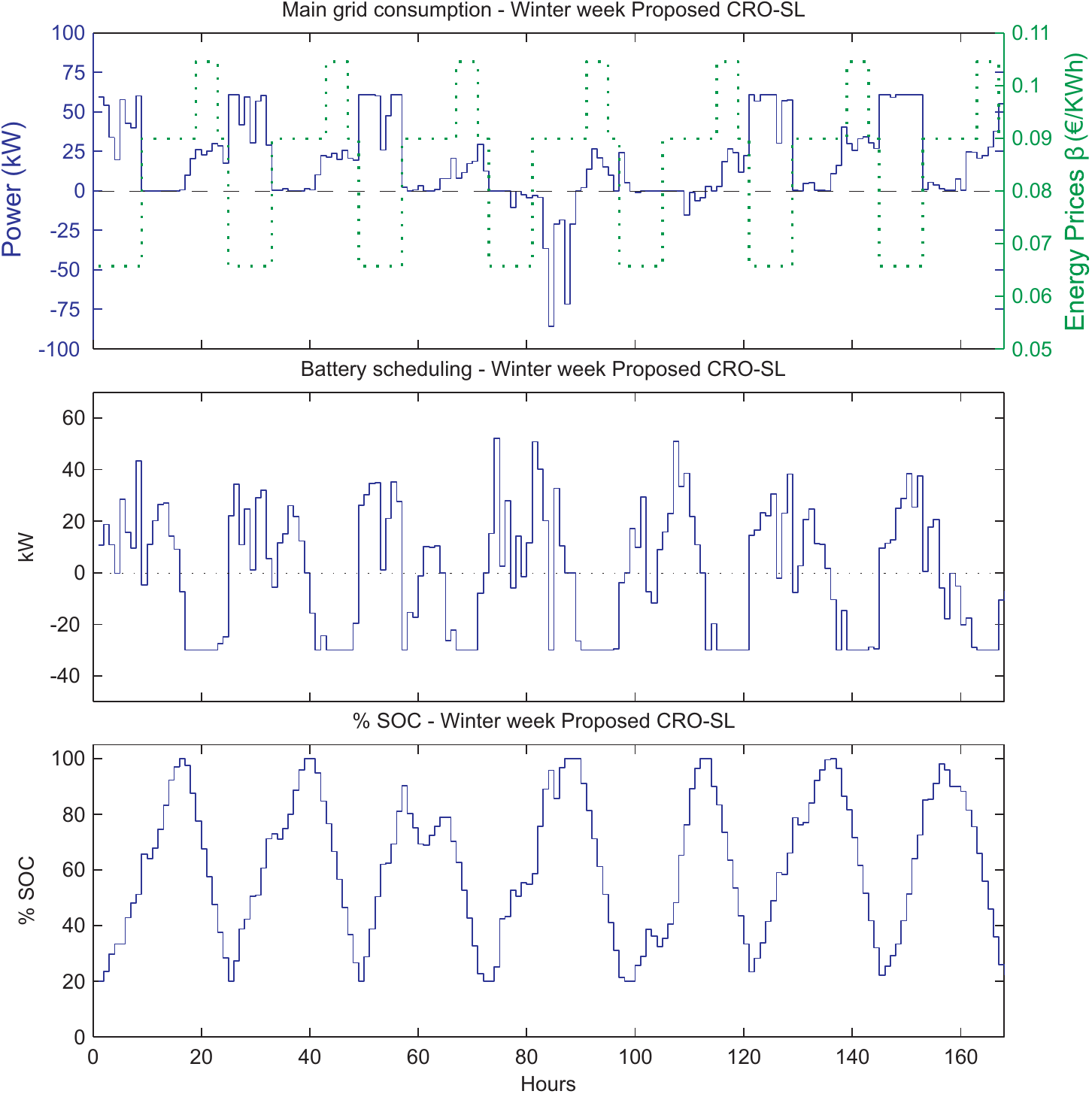}
\caption{\label{61Resultados_inviernoCRO} Proposed CRO-SL results: Consumption from the main grid, Battery power scheduling and \% SOC.}
\end{center}
\end{figure}

\begin{figure}[!ht]
\begin{center}
\includegraphics[draft=false,angle=0,width=10cm]{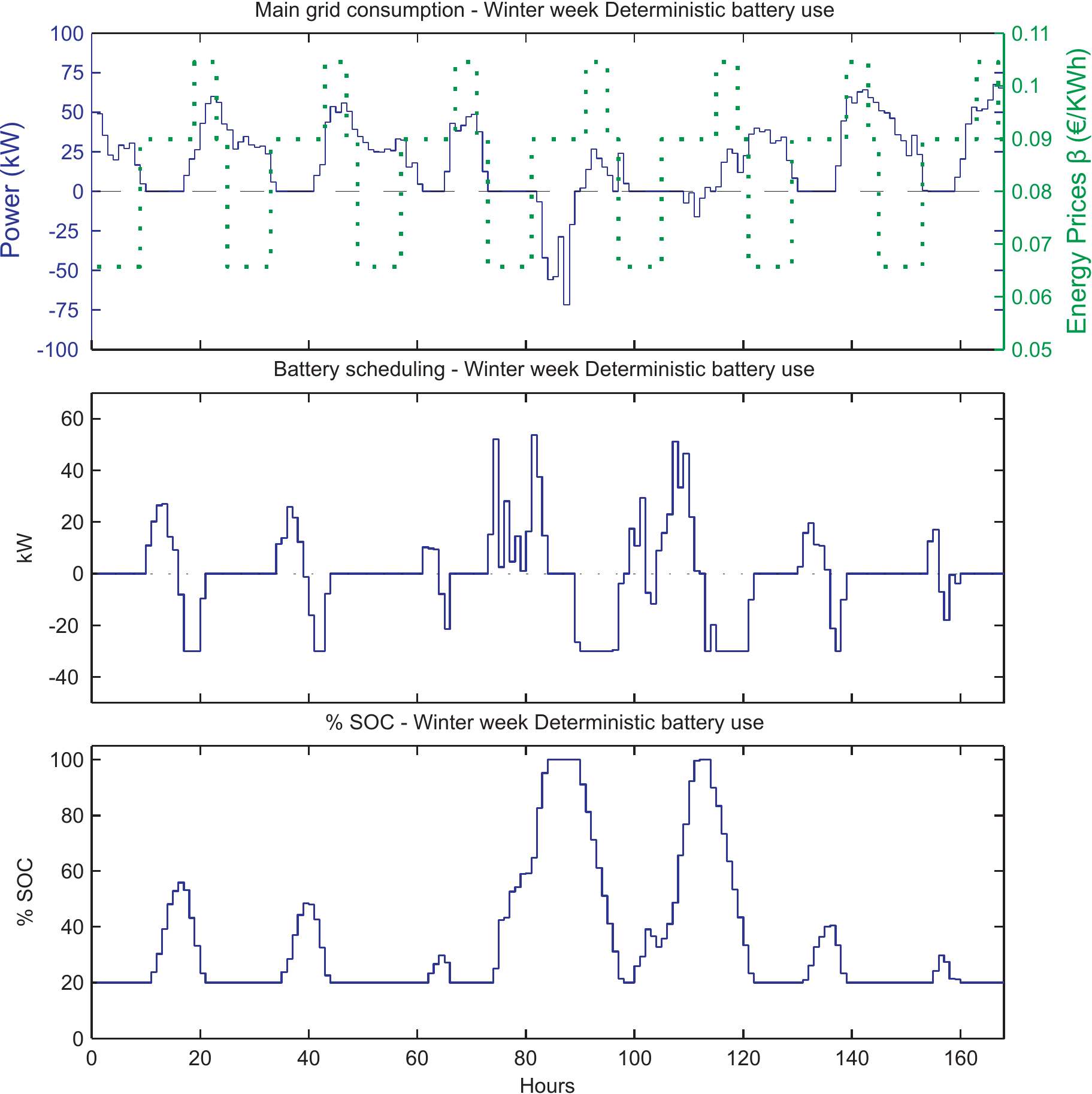}
\caption{\label{61Resultados_invierno} Deterministic approach results: Consumption from the main grid, Battery power scheduling and \% SOC.}
\end{center}
\end{figure}

We can graphically analyze the result obtained with the CRO-SL in Figure \ref{61Resultados_inviernoCRO}, in terms of consumption from the main grid, battery power scheduling and \% SOC of the battery, in the winter week considered. A comparison with the Deterministic approach in Figure \ref{61Resultados_invierno} shows the effect of a good battery scheduling given in this case by the CRO-SL algorithm. First, note that an optimal power scheduling for an energy storage device on the MG provides a significant cost reduction over the Deterministic scheduling, providing a shorter recovery time of investment. The main goal of using an energy storage device is to avoid energy waste at the moments when the energy produced by the generators is larger than the energy demanded by the loads attached to the MG. However, we have shown that the use of the battery in a Deterministic way, taken only into account one single instant of load and generation (and ignoring the future instants predicted on its weekly profiles) is not an optimal procedure for performing the scheduling. Instead, the use of a meta-heuristic approach such as the proposed CRO-SL is able to provide a solid battery scheduling, with a reduced consumption from the main grid. An important characteristic of the scheduling obtained using the proposed CRO-SL, is that the periods of time in which the battery is charged may not be those in which an excess of generation occurs (see Figure \ref{61Resultados_invierno}, main grid consumption). In this case, battery scheduling allows reducing the electricity consumption in the most expensive periods of time, shifting it towards the cheapest ones, which means a better battery use.

\begin{figure}[!ht]
\begin{center}
\includegraphics[draft=false, angle=0,width=10cm]{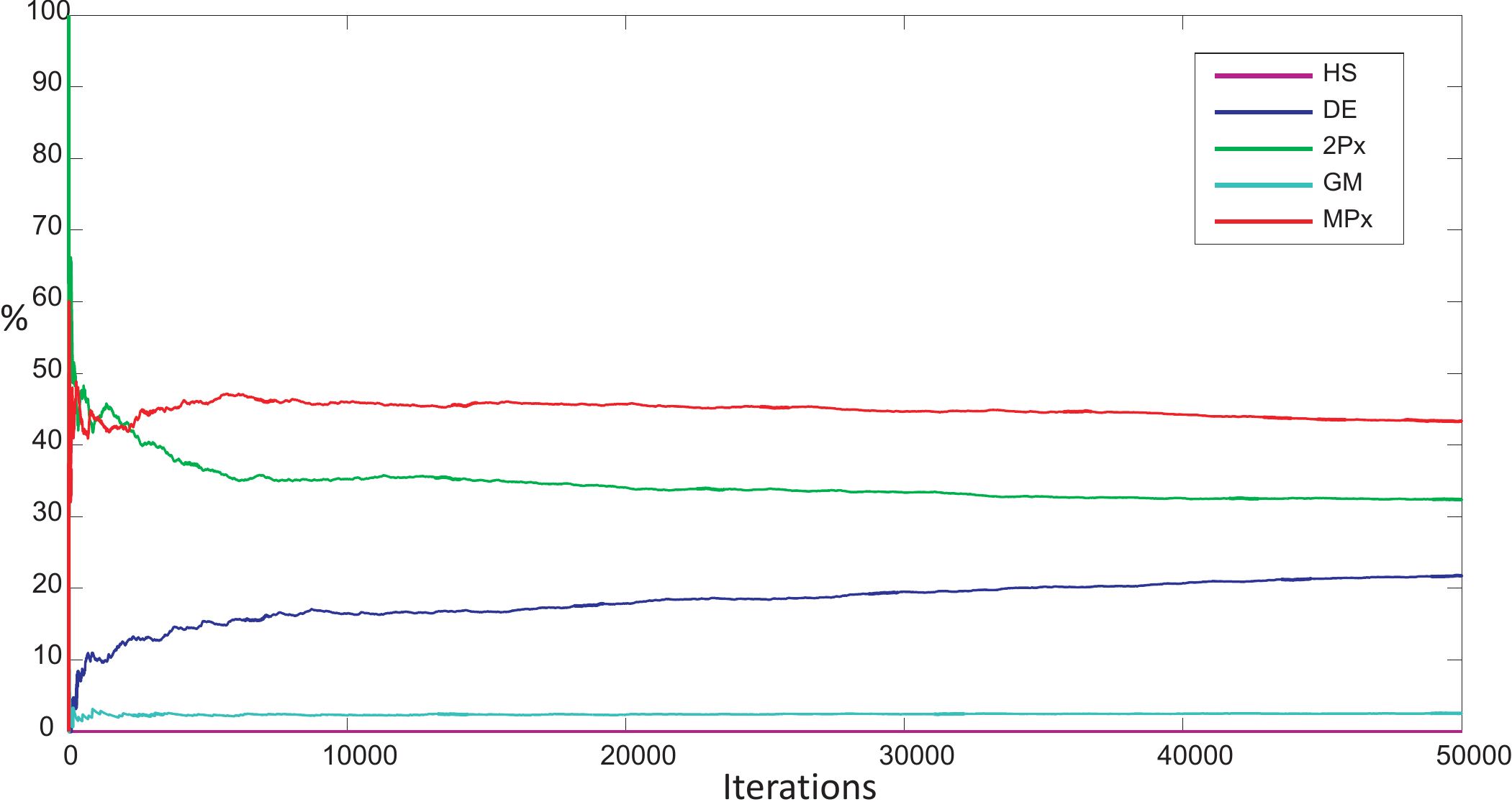}
\end{center}
\caption{ \label{61Comp_Search1} Comparison of the effect of the different substrates (exploration procedures) in the Winter week considered (percentage of times in which a given substrate gives the best larva (new solution) in each generation)}
\end{figure}

\begin{figure}[!ht]
\begin{center}
\includegraphics[draft=false, angle=0,width=10cm]{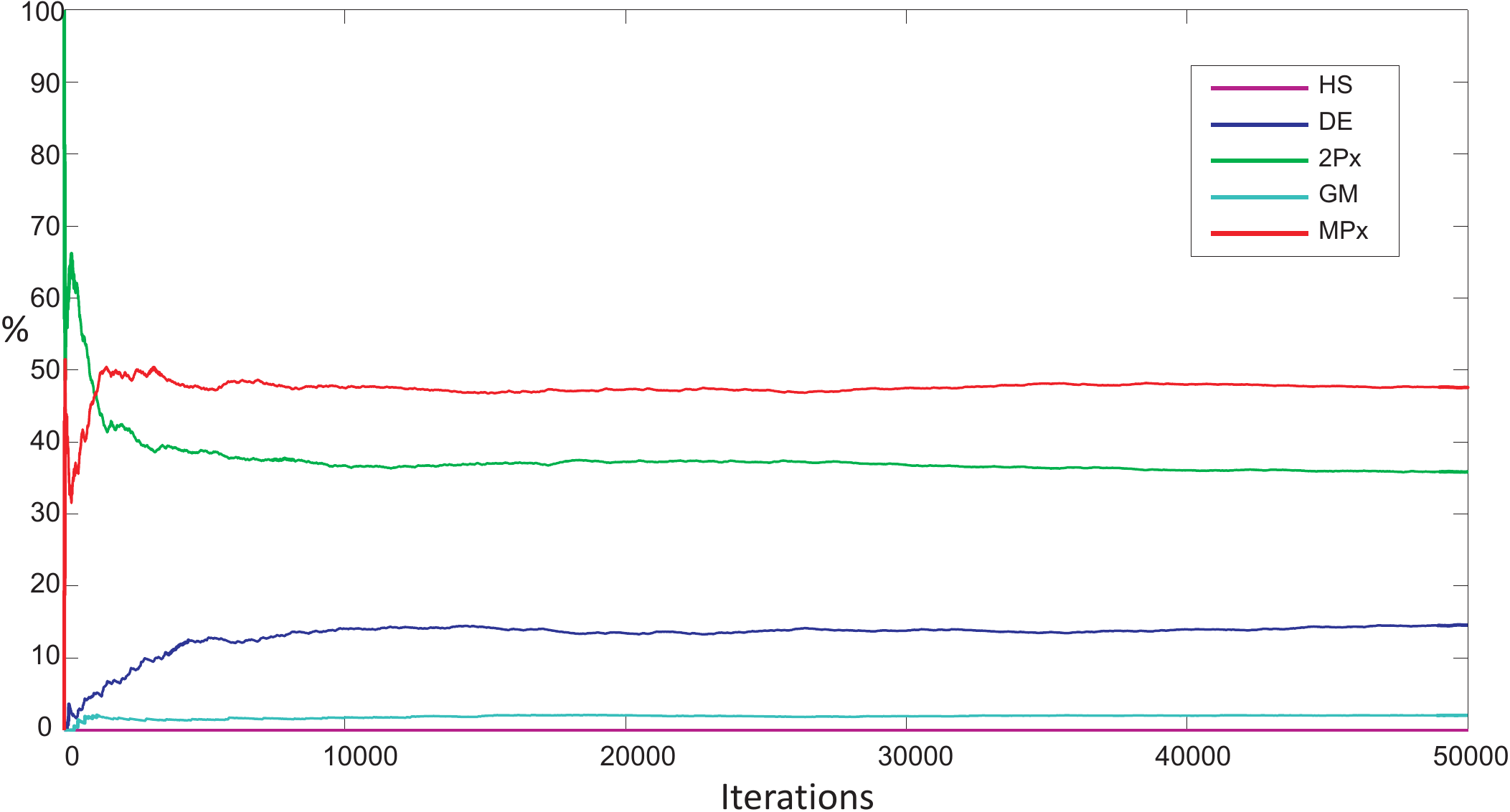}
\end{center}
\caption{ \label{61Comp_Search2} Comparison of the effect of the different substrates (exploration procedures) in the Spring week considered (percentage of times in which a given substrate gives the best larva (new solution) in each generation)}
\end{figure}

\begin{figure}[!ht]
\begin{center}
\includegraphics[draft=false, angle=0,width=10cm]{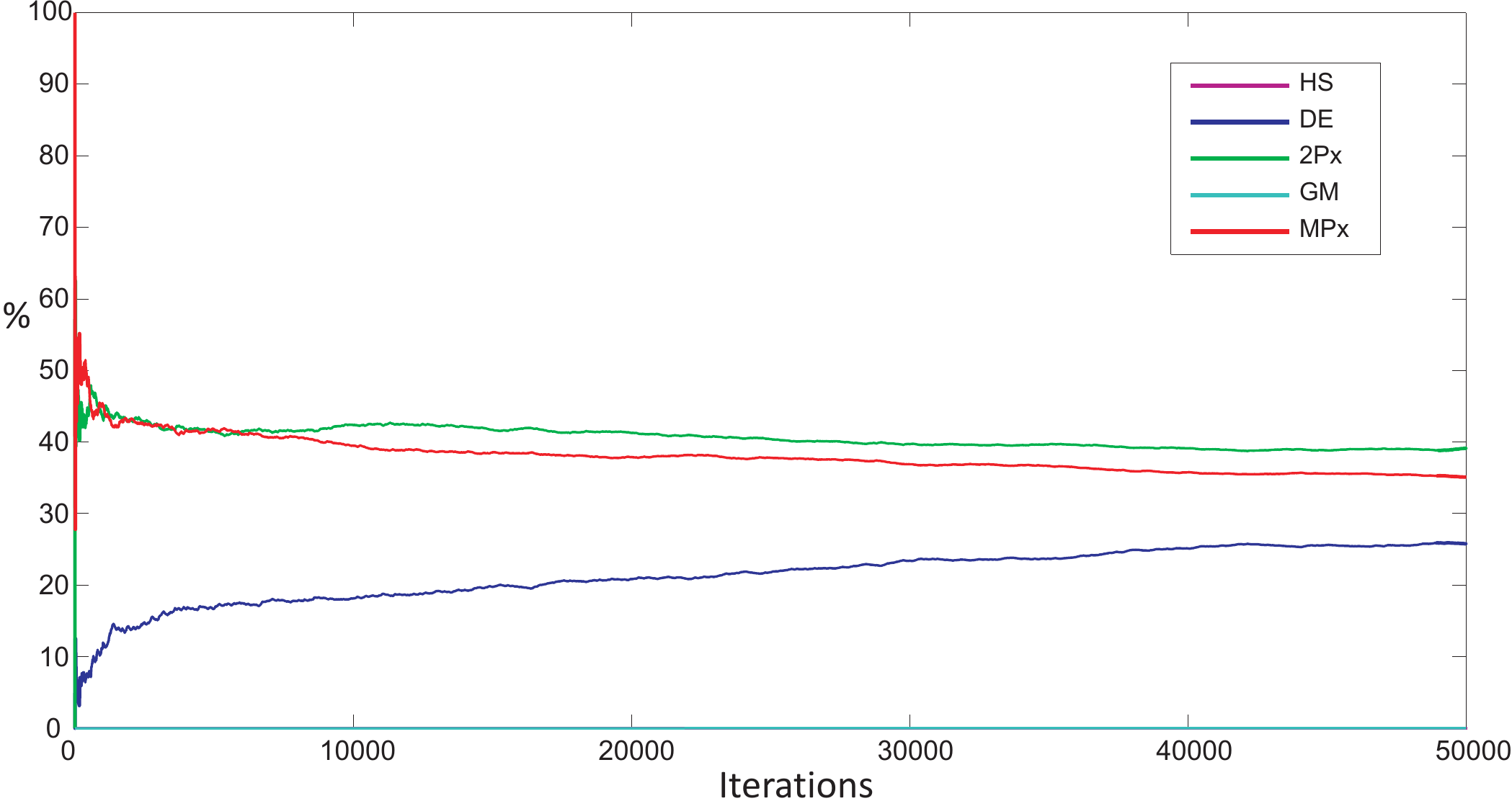}
\end{center}
\caption{ \label{61Comp_Search3} Comparison of the effect of the different substrates (exploration procedures) in the Summer week considered (percentage of times in which a given substrate gives the best larva (new solution) in each generation)}
\end{figure}

\begin{figure}[!ht]
\begin{center}
\includegraphics[draft=false, angle=0,width=10cm]{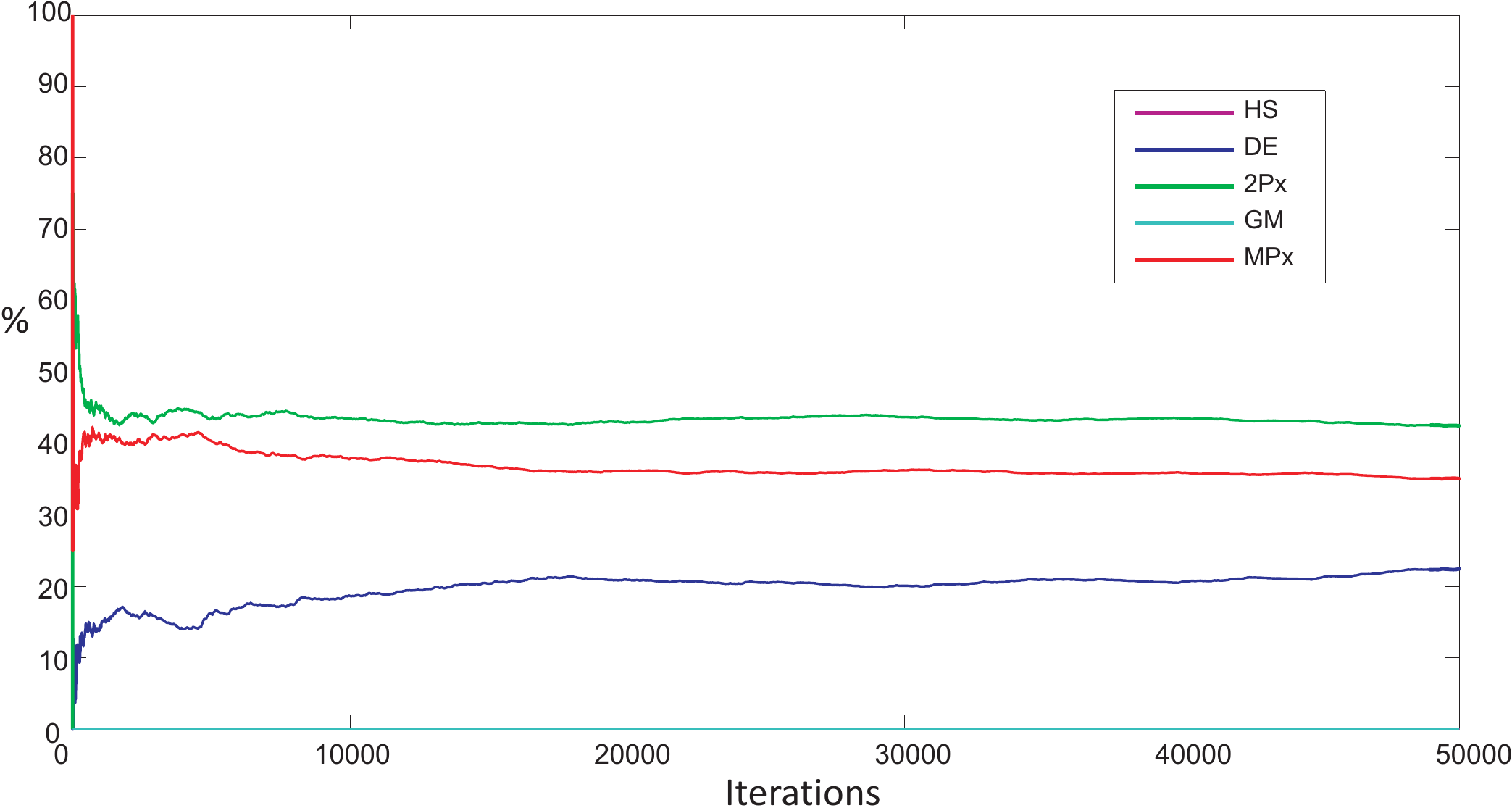}
\end{center}
\caption{ \label{61Comp_Search4} Comparison of the effect of the different substrates (exploration procedures) in the Autumn week considered (percentage of times in which a given substrate gives the best larva (new solution) in each generation)}
\end{figure}

A detailed  analysis on the computational performance of the CRO-SL approach can be carried out. First, Figures \ref{61Comp_Search1}-\ref{61Comp_Search4} give an idea of the importance of each substrate in the overall performance of the CRO-SL. They show the percentage of times in which a given substrate gives the best larva (new solution) in each iteration of the algorithm, for the different weeks considered in the work. These figures are really informative about the contribution of the different exploration approaches implemented in each substrate to the global CRO-SL performance. As can be seen, the two-points crossover and the multi-point crossover are consistently the searching procedures which contribute the most to obtain good quality solutions for the problem. DE based exploration also contributes in a significant way to the algorithm's performance in all the weeks analyzed. On the other hand, it seems that contribution of the GM and the HS exploration procedures is rather low in this particular optimization problem. Note that the CRO-SL is an excellent framework to evaluate the performance of different exploration mechanism in any optimization problem, in this case in the BSOP.

An alternative analysis can be carried out by comparing the proposed CRO-SL results with that of the CRO with a single substrate layer, in which the different approaches implemented (DE, HS, GM, 2Px and MPx) are implemented on its own. Table \ref{tab:61Tabla_comparativa} shows the results obtained, to be compared with the complete CRO-SL approach. As can be seen, the co-evolution of different exploration procedures produces better results, improving the performance of the CRO with single substrate layer. These results show how the competitive co-evolution of the different searching procedures in one population helps improve the searching capabilities of the algorithm. Note that the performance of the CRO with DE, 2Px and MPx is worse than the CRO-SL, though better than the one obtained by the deterministic approach (base line). The GM search on its own only improves the performance of the Deterministic procedure in the Summer week, whereas the HS is not able to improve the Deterministic procedure. This shows the importance of using a co-evolution procedure such as the CRO-SL to improve the performance of the meta-heuristic search in the BSOP problem tackled.

\begin{table}[!ht]
\renewcommand{\arraystretch}{1.3}
\caption{Comparison between CRO results with different substrates.}
\vspace{5mm}
\label{tab:61Tabla_comparativa}
\centering
\begin{tabular}{ccccc}
\hline
\hline
CRO & Winter week & Spring week & Summer week & Autumn week\\
\hline
 CRO-SL (5 substrates) & 379.31 & 165.66  &  335.32   & 308.55\\

HS & 427.68 & 167.39 & 359.33 & 328.06\\

DE & 389.33 & 167.39 & 351.54 & 324.97\\

2Px  & 399.03 & 167.11 & 338.08 & 315.83\\

GM & 427.68 & 169.39 & 351.24 & 328.06\\

MPx  & 419.32 & 167.03 & 338.09 & 324.07\\
\hline
\end{tabular}
\end{table}

\section{Conclusions}\label{61Conclusions}
In this chapter we have presented the application of the CRO-SL to a problem of battery scheduling optimization in Micro-Grids, in an environment of variable electricity prices. First, the problem of battery scheduling in MGs has been detailed, including a complete mathematical formulation of the problem. Then, the CRO-SL proposed to solve the problem has been described, including the specific problem encoding, initialization and different operators applied to evolve the candidate solutions. In the experimental part of the work, we have shown the excellent performance of the proposed algorithm in a realistic BSOP, considering a MG with photovoltaic and wind generations and two loads type profiles (residential and industrial). Savings up to 35.5\% in the spring week (outperforming in 1.03\% the Deterministic use of the battery), around 24.7\% in winter (11.3\%), 23.7\% in summer (6.7\%) and 30.8\% in autumn week (6\%), are obtained. The effect of co-evolve different meta-heuristics exploration approaches have also been analyzed and compared to the case of application of a single search procedure, showing the goodness of the proposed CRO-SL for competitive co-evolution of exploration procedures.

\chapter{Structures Vibration Control via Tuned Mass Dampers Optimization}\label{cap:tmds}

\section{Introduction and state of the art}
Problems in structural optimization are often characterized by search spaces of extremely high dimensionality and nonlinear objective functions \cite{Saka16}. In these optimization problems, classical approaches do not lead, in general, to good solutions, or in many occasions they are just not applicable, due to the unmanageable search space structure or its huge size, which implies an extremely high computation cost. In this context, modern optimization meta-heuristics have been successfully applied to an important number of structural optimization problems \cite{Glover03}. Meta-heuristics algorithms have been shown as a possibility to obtain a good enough solution to a given problem which cannot be tackled with exact algorithms.

There are different meta-heuristics that have been applied to structural engineering problems. Genetic and evolutionary algorithms have been applied to the optimization of discrete structures in \cite{Rajeev92}. There have been other works that applied GAs in structural optimization problems such as shape optimization \cite{Soh96}, optimization of 3D trusses \cite{Togan06}, impact load characterization of concrete structure \cite{Yan09}, the plane stress problem \cite{Simonetti14} or welded beam optimization problems \cite{Deb91}. The PSO algorithm is another important meta-heuristic which has been successfully applied to structural optimization problems, such as truss layout \cite{Kaveh14b} or truss structures optimization \cite{Schutte03}. The HS approach \cite{Ge01,Saka16} and the teaching-based learning algorithm \cite{Rao11b,Rao11,Degertekin13} have also been used to solve mechanical design optimization problems. In the last few years, alternative modern meta-heuristics based on physics process have been applied to structural optimization problems, such as the Big-Bang Big-Crunch algorithm \cite{Kaveh13}, the colliding bodies optimization algorithm \cite{Kaveh14}, the Ray optimization \cite{Kaveh12} or the charged system search algorithm \cite{Kaveh10}.

In this chapter, the CRO-SL is applied to the design and location of Tuned Mass Dampers for structures subjected to earthquake ground motions. A TMD, which can be used for passive and semi-active control strategies, improves the vibration response of a structure by increasing its damping (i.e. energy dissipation) and/or stiffness (i.e. energy storage) through the application of forces generated in response to the movement of the structure \cite{Symans99}. In the case of structures with spatially distributed and closely spaced natural frequencies, the TMD design may not be obvious, because Den Hartog's theory \cite{Hartog56} cannot be applied due to the existence of a coupling between the motions of the vibration modes of the structures and the used TMDs \cite{Abe95}. Multi-storey buildings are good examples of structures with spatially distributed and closely spaced natural frequencies. For example, Greco et al. \cite{Greco2015} proposes a robust optimum design of tuned mass dampers installed on multi-degree-of-freedom systems subjected to stochastic seismic actions. Other similar examples can be found in \cite{Fleck-Fadel2014} and \cite{Debnath2013}. In this work, the generalized framework presented in \cite{Mohtat2011} is used to formulate a $N$ floor building where $M$ TMDs must be installed. Unlike \cite{Mohtat2011}, where the position of each TMD is fixed (p.e., $M$ TMDs in one floor or one TMD for each floor), this work proposes a modification that allows the optimization algorithm deciding the position of each TMD (i.e., a TMD can be placed at any floor to damp any vibration mode). In addition, an interval for the mass, damping and stiffness is defined for each TMD. Thus, the optimization algorithm will try to find the best solution by obtaining the $4\times M$ parameters (3 physical parameters and the TMD location).

\section{Problem definition} \label{sec:62Model}

\begin{figure}[!ht]
\centering
\includegraphics[width=0.8\linewidth]{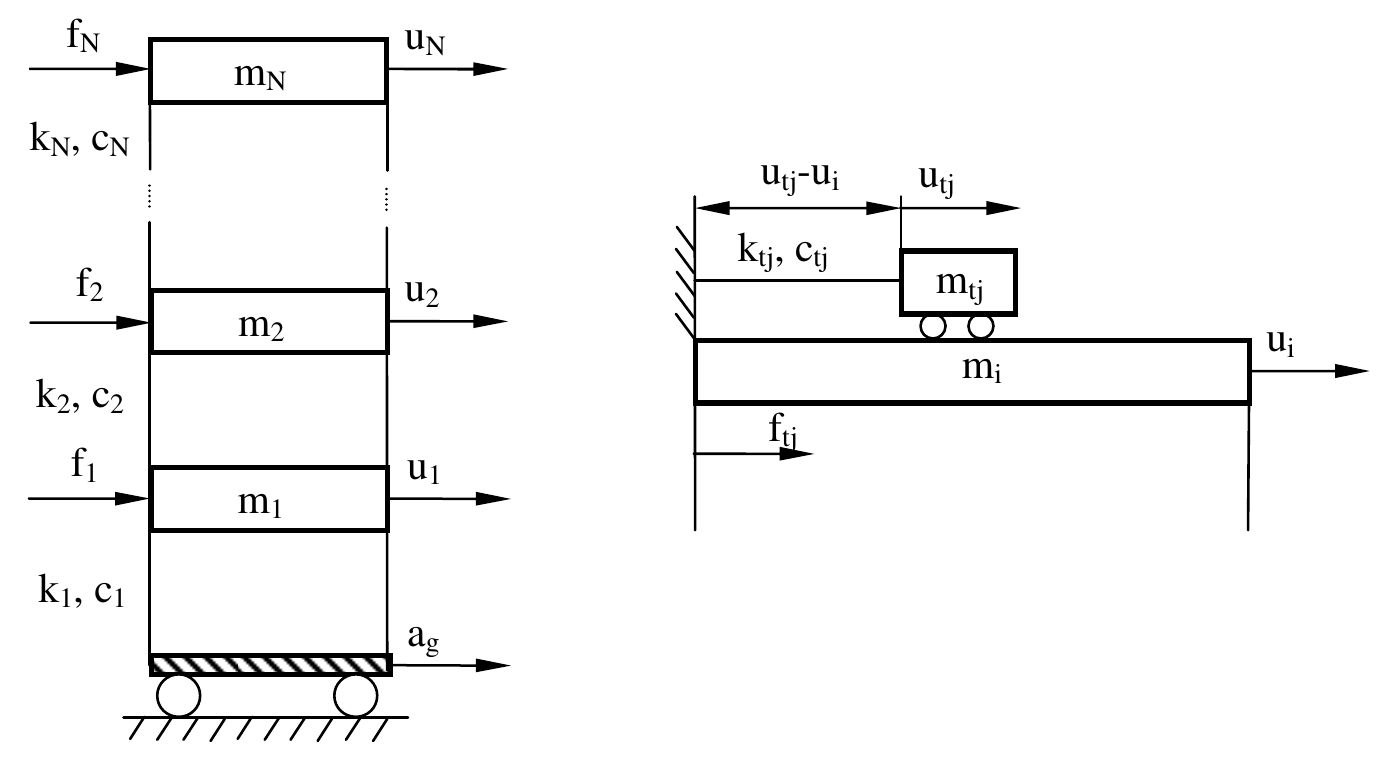}
\caption{(a) N storey building and (b) TMD models.}
\label{62Building}
\end{figure}

The $N$ storey building can be modelled as a $N$ degree of freedom system (see Figure \ref{62Building} (a)), where the mass is concentrated at each floor ($m_1$, $m_2$, \ldots,$m_N$), $k_i$ and $c_i$ are, respectively, the $i^{th}$ floor stiffness and damping coefficient (relative to $(i-1)^{th}$ floor or to the ground if $i=1$).

If the applied forces in each floor ($\mathbf{f}=\left[f_1,f_2,\cdots,f_N \right]^{T}$) and the acceleration of the ground ($a_g$) are considered as inputs, the differential equation of the building can be represented as follows:
\begin{equation} \label{Buid1}
\mathbf{M}\ddot{\mathbf{u}}+\mathbf{C}\dot{\mathbf{u}}+\mathbf{K}\mathbf{u}=-\mathbf{M}\mathbf{r}_g a_g+\mathbf{f},
\end{equation}
where $\mathbf{M}=diag(m_1, m_2, \ldots,m_N)$ is the structural mass matrix,
\begin{eqnarray} \label{Buid2}
\mathbf{K}=\begin{bmatrix}
k_1+k_2& -k_2   & 0      & \cdots      & 0       \\
-k_2   &k_2+k_3 &-k_3     & \cdots      & 0       \\
\vdots & \vdots & \ddots & \cdots      & \vdots  \\
0      &   0    &-k_{N-1}&k_{N-1}+k_{N}& -k_N    \\
0      &   0    &0       &-k_{N}       &  k_N
\end{bmatrix}
\end{eqnarray}
is the structural stiffness matrix, $\mathbf{C}$ is the proportional damping matrix, $\mathbf{u}=\left[u_1,u_2,\cdots,u_N\right]^{T}$ are the floor displacements relative to the base and $\mathbf{r}_g=\left[1\cdots1\right]^{T}$ is the influence vector of the ground acceleration. In order to simplify the model, the damping matrix  is assumed as $\mathbf{C}=\frac{2\xi_s\omega_1\omega_2}{\omega_1+\omega_2}\mathbf{M}+\frac{2\xi_s}{\omega_1+\omega_2}\mathbf{K}$. Note that (like at reference \cite{Mohtat2011}), the dissipation matrix assuming that damping forces depend only on generalized velocities is not the only linear model of vibration damping (see reference \cite{Woodhouse1998} for a detailed discussion). However, for simplification of the illustration, a dissipation matrix damping is assumed for the shear building model (this does not affect the generality of the framework, because the performance measures are formulated using the plant without any assumptions rather than linearity).

The state space state model can be deduced from Equation (\ref{Buid1})
\begin{eqnarray} \label{Buid6}
\begin{bmatrix}
\dot{\mathbf{u}}  \\
\ddot{\mathbf{u}}
\end{bmatrix} = \begin{bmatrix}
\boldsymbol{0}_{N \times N}     &  \mathbf{I}_{N \times N} \\
-{\mathbf{M}}^{-1}{\mathbf{K}}  &  -{\mathbf{M}}^{-1}{\mathbf{C}}
\end{bmatrix} \begin{bmatrix}
\mathbf{u}  \\
\dot{\mathbf{u}}
\end{bmatrix} + \begin{bmatrix}
\boldsymbol{0}_{N \times 1}    &   \boldsymbol{0}_{N \times N}      \\
-\mathbf{r}_g      &   \mathbf{M}^{-1}
\end{bmatrix}    \begin{bmatrix}
a_g             \\
\mathbf{f}
\end{bmatrix}, \\ \nonumber
\mathbf{y}= \ddot{\mathbf{u}} + \mathbf{r}_ga_g = \left[ -{\mathbf{M}}^{-1}{\mathbf{K}}  \  -{\mathbf{M}}^{-1}{\mathbf{C}} \right]\begin{bmatrix}
\mathbf{u}  \\
\dot{\mathbf{u}}
\end{bmatrix} + \mathbf{M}^{-1} \mathbf{f},
\end{eqnarray}
where $\mathbf{y}$ is the vector formed by the absolute accelerations (i.e., the accelerations measured with the accelerometers installed at each floor).

\begin{figure}[!ht]
\centering
\includegraphics[width=0.4\linewidth]{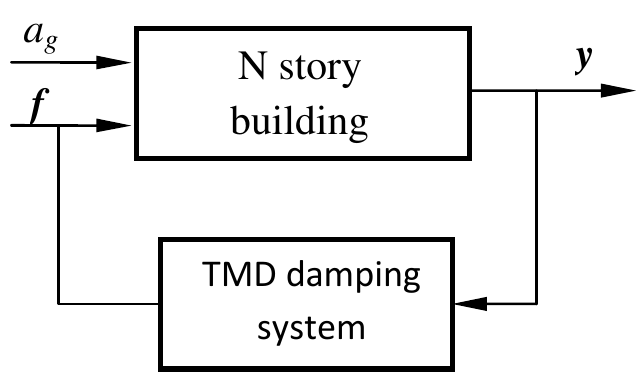}
\caption{General framework.}
\label{62CL}
\end{figure}

The TMD can be modelled as a one degree of freedom system (see Figure \ref{62Building} (b)), where $m_{tj}$ is the mass, $k_{tj}$ and $c_{tj}$ are the TMD linear stiffness and damping coefficient of the $j^{th}$ TMD relative to the $i^{th}$ floor. The differential equation of the TMD relates the accelerations of the ground floor ($a_g$) the mass of the $i^{th}$ floor ($u_i$) and the mass of the $j^{th}$ TMD as follows:
\begin{equation} \label{TMD1}
m_{t,j}\ddot{u}_{tj}+c_{t,j}\dot{u}_{t,j}+k_{t,j}u_{t,j}-c_{t,j}u_{i}-k_{t,j}u_{i}=-m_{t,j}a_g,
\end{equation}
where the force exerted by the $j^{th}$ TMD on the $i^{th}$ floor is:
\begin{equation} \label{TMD2}
f_{t,j}=k_{t,j}\left(u_{t,j}-u_{i}\right)+c_{t,j}\left(\dot{u}_{t,j}-\dot{u}_{i}\right)=k_{t,j}u_{r,ij}+c_{t,j}\dot{u}_{r,ij},
\end{equation}
and the relative displacement between $j^{th}$ TMD and $i^{th}$ floor is defined as $u_{r,ij}=u_{t,j}-u_{i}$. If the variable $u_{r,ij}$ is considered, the Equation (\ref{TMD1}) can be arranged as the following:
\begin{equation} \label{TMD1a}
m_{t,j}\ddot{u}_{r,ij}+c_{t,j}\dot{u}_{r,ij}+k_{t,j}u_{r,ij}=-m_{t,j}\left(a_g+\ddot{u}_i\right)=-m_{t,j}y_i.
\end{equation}
Note that Equations (\ref{Buid6}), (\ref{TMD2}) and (\ref{TMD1a}) define the system formed by the $N$ floor building and the $M$ TMDs, which can be represented as in Figure \ref{62CL}. Note also that the values of the applied forces in each floor ($f_i$) are equal to the sum of the forces of all TMD located in this floor according to Equation (\ref{TMD2}).
The optimization problem proposed in this work consists of minimizing the maximum of the Frequency Response Functions (FRFs) defined between each output ($y_i$) and the ground acceleration ($a_g$). Thus, the Equations (\ref{Buid6}), (\ref{TMD2}) and (\ref{TMD1a}) must be defined at frequency domain. First of all, the state space model of Equation (\ref{Buid6}) can be defined as the following $N \times (N+1)$ matrix of transfer functions \cite{Gawronski2004}:
\begin{eqnarray} \label{CL1}
\begin{bmatrix}
Y_1(s) \\
Y_2(s) \\
\vdots \\
Y_N(s)
\end{bmatrix} = \begin{bmatrix}
G_{11}(s) & G_{12}(s) & \cdots & G_{1N}(s) & G_{1g}(s) \\
G_{21}(s) & G_{22}(s) & \cdots & G_{2N}(s) & G_{2g}(s) \\
\vdots    & \vdots    & \ddots & \vdots    & \vdots    \\
G_{N1}(s) & G_{N2}(s) & \cdots & G_{NN}(s) & G_{Ng}(s) \end{bmatrix} \begin{bmatrix}
F_1(s) \\
F_2(s) \\
\vdots \\
F_N(s) \\
A_g(s), \end{bmatrix}
\end{eqnarray}
where $s$ is the (complex) frequency variable and the capital letters $Y_i(s)$, $F_i(s)$ and $A_g(s)$ denote the Laplace transform of $y_i$, $f_i$ and $a_g$, respectively. Finally, $G_{i_yi_f}(s)$ is the transfer function between the acceleration measured at $i_y^{th}$ floor and the force applied to the $i_f^{th}$ floor and $G_{i_yg}(s)$ is the transfer function between the acceleration measured at $i_y^{th}$ and the ground acceleration ($a_g$).

From Equations (\ref{TMD2}) and (\ref{TMD1a}) the following transfer function ($H_j(s)$) for the TMD system can be deduced:

\begin{equation} \label{TMD4}
\frac{F_{t,j}(s)}{Y_i(s)}=-\frac{m_{t,j}\left( c_{t,j}s+k_{t,j} \right)}{m_{t,j}s^2+c_{t,j}s+k_{t,j}} = -m_{t,j}\frac{2\xi_{t,j}\omega_{t,j}s+\omega^{2}_{t,j}}{s^2+2\xi_{t,j}\omega_{t,j}s+\omega^{2}_{t,j}}=H_j(s),
\end{equation}

where $\omega_{t,j}=\sqrt{(k_{t,j}/m_{t,j})}$ and $\xi_{t,j}=c_{t,j}/2\sqrt{k_{t,j}m_{t,j})}$ are the natural frequency and damping ratio of the $j^{th}$ TMD as an isolated system, respectively, and $F_{t,j}(s$) and $Y_i(s)$ are the Laplace transform of $f_{t,j}$ and $y_i$.

The Equations (\ref{CL1}) and (\ref{TMD4}) can be connected as in the general framework of Figure \ref{62CL} with the following equation:

\begin{equation} \label{TMD5}
F_{i}(s)=\sum^{M}_{j=1}K_{j,i}H_j(s)Y_{i}(s),
\end{equation}

where $K_{ji}=1$ if the TMD $j$ is placed on $i^{th}$ floor. Once the general framework is defined in the Laplace domain, the optimization problem consist of minimizing the following functional:

\begin{equation} \label{min11}
g(\bf{x})=\max \left\{   \left\|\frac{Y_1(j\omega)}{A_g(j\omega)}\right\|_{\infty},\left\|\frac{Y_2(j\omega)}{A_g(j\omega)}\right\|_{\infty}, \cdots,\left\|\frac{Y_N(j\omega)}{A_g(j\omega)}\right\|_{\infty} \right\},
\end{equation}

by finding the optimal parameters of ${\bf x}=\left[\boldsymbol{\Omega}_t,\boldsymbol{\Xi}_t,\mathbf{M}_t,\mathbf{FB}\right]$, where $\left\|·\right\|_{\infty}$ is the infinity norm, $\boldsymbol{\Omega}_t=\left[\omega_{t,1},\omega_{t,2},\cdots,\omega_{t,M} \right]$, $\boldsymbol{\Xi}_t=\left[\xi_{t,1},\xi_{t,2},\cdots,\xi_{t,M}\right]$, $\mathbf{M}_t=\left[m_{t,1},m_{t,2},\cdots,m_{t,M} \right]$ and $\mathbf{FB}=\left[fb_1,fb_2,\cdots,fb_M \right]$. Note that the parameter $fb_j=i$ if the TMD $j$ is placed on the $i^{th}$ floor (i.e., $K_{ji}=1$). This problem can be formulated as follows:

\begin{equation} \label{min2}
\min_{{\bf x}}\left(g({\bf x})\right).
\end{equation}

\section{CRO substrates definition and main varieties}
In this version neither regeneration of the population nor asexual reproduction have been carried out. The substrates carried out in the experiments by the algorithm are the same as shown in the chapter \ref{cap:scheduling}:
\begin{enumerate}
\item HS: Mutation from the Harmony Search algorithm with four $\delta$ values for every type of optimization variable (mass, damping coefficient, natural frequency and positioning) : $\delta=[0.01 0.02 0.3 0.5]$.
\item DE: Mutation from Differential Evolution algorithm with $F$ value linearly decreasing during the run,
      from $2$ to $0.5$.
\item 2Px: Classical 2-points crossover.
\item GM: Gaussian Mutation, with a $\delta$ value linearly decreasing during the run,
      from $10$ to $1$.
\item MPx: Multi-points crossover ($M=3$).
\end{enumerate}

Note that in this case the encoding in this optimization problem mixes integer numbers, such as the positioning of the TMD's, with real ones, and within the latter, it can be found some with values of several tens, for the value of resonance frequency, and others that do not exceed a few decimals, such as the damping coefficient or the mass. This is why mutation cannot be done with large parameters, as some would only explore their extreme values, or with small parameters, because others would only explore a very prudent part of their search space. The solution has been to define the mutation parameters as a function of the search spaces of each variable to be optimized.

\section{Computational evaluation, comparisons and results}\label{62Experimentos}

The examples carried out to evaluate the proposed CRO-SL in this context consist of designing and locating $M$ TMDs on a $N$ floor building. The CRO-SL parameters used in the experiments are shown in Table \ref{tab:62parameters}.

\begin{table}[!ht]
    \centering
\caption{Parameters values used in the hybridization of the CRO-SL.}
\label{tab:62parameters}
\begin{tabular}{llr}
\hline
Parameter & Description & value\\
\hline
Reef & Reef size & 120\\
$F_b$ & Frequency of broadcast spawning & 97\%\\
$\mathcal{N}_{att}$ & Number of tries for larvae settlement & 3\\
$F_d$ & Fraction of corals for depredation & 15\%\\
$P_d$ & Probability of depredation & 10\%\\
$n_T$ & Maximum number of iterations & 1000\\

\hline
\end{tabular}
\end{table}

\begin{figure}[!ht]
\centering
\includegraphics[width=0.7\linewidth]{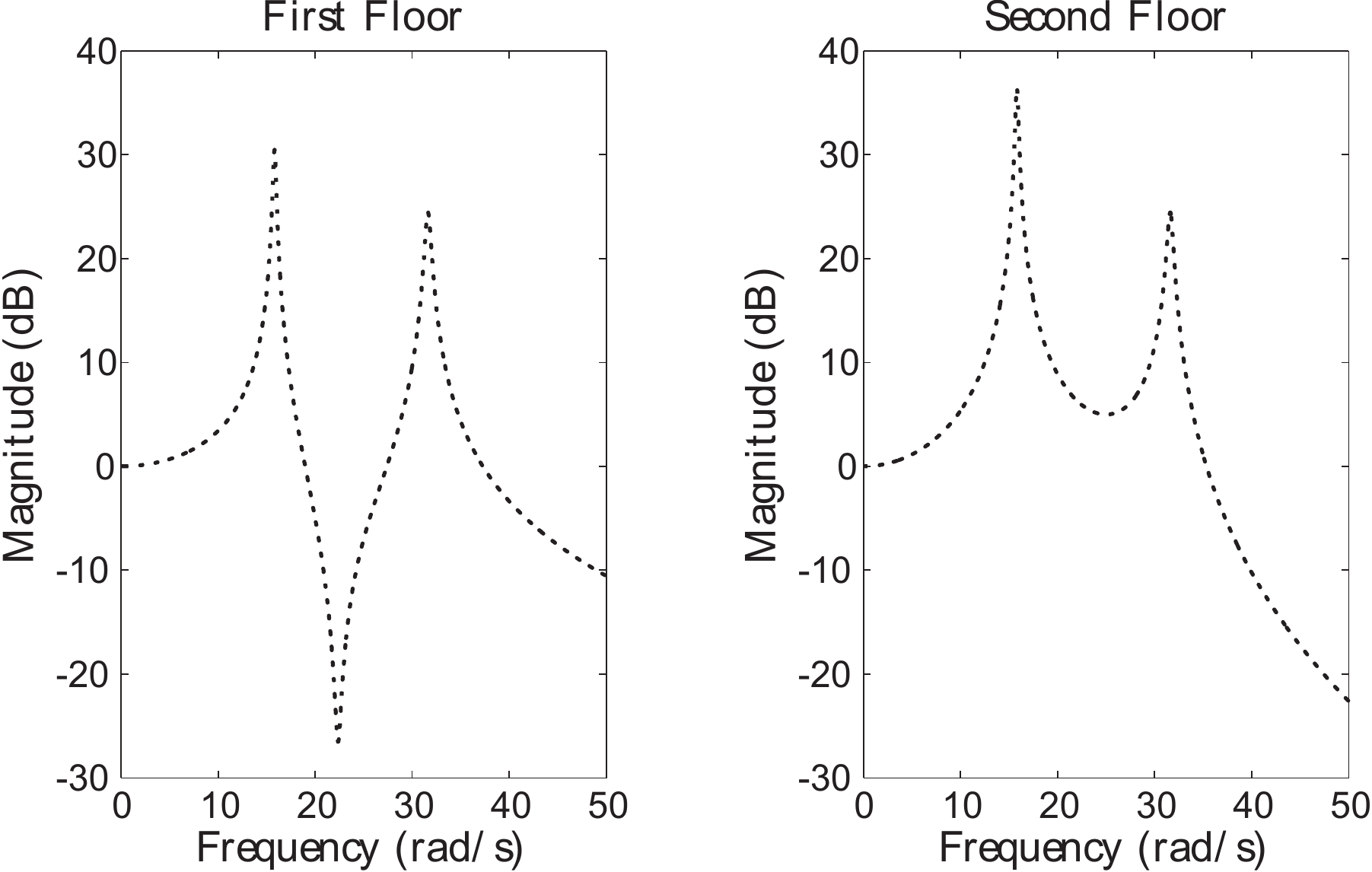}
\caption{FRF for the $N=2$ floors case without any TMD installed.}
\label{62Results2floorswithoutTMD}
\end{figure}

The CRO-SL with the previously defined parameters has been applied to solve two different application examples, consisting of designing and locating $M=N$ TMDs for a $N=2$ and $N=4$ storey building. The TMDs can be placed on any floor to damp any vibration mode. The FRF between the acceleration of each floor ($y_j$) and the acceleration of the ground ($a_g$) will be used to show the performance of the optimal design. The parameters for the $N=2$ floor building are the following: i) $k_1=1000$ N/m and $k_2=500$  N/m  ii) $m_1=2$ kg and $m_2=1$ kg and iii) $\xi_s=0.01$. With these parameters, the natural frequencies and damping are $\omega_1=15.811$ rad/s, $\omega_2=31.623$ rad/s and $\xi_1=\xi_2=0.010$. The FRF between the acceleration of each floor ($y_j$) and the acceleration of the ground ($a_g$) is shown in Figure \ref{62Results2floorswithoutTMD} (without any TMD). The constraints for $\bf{x}$ are $\omega_{t,j}\in[0,50]$ rad/s, $\xi_{t,j}\in[0,0.3]$, $m_{t,j}\in[0,0.05]$ kg and $fb_j\in\left\{1,2\right\}$.

\begin{figure}[!ht]
\centering
\includegraphics[width=0.7\linewidth]{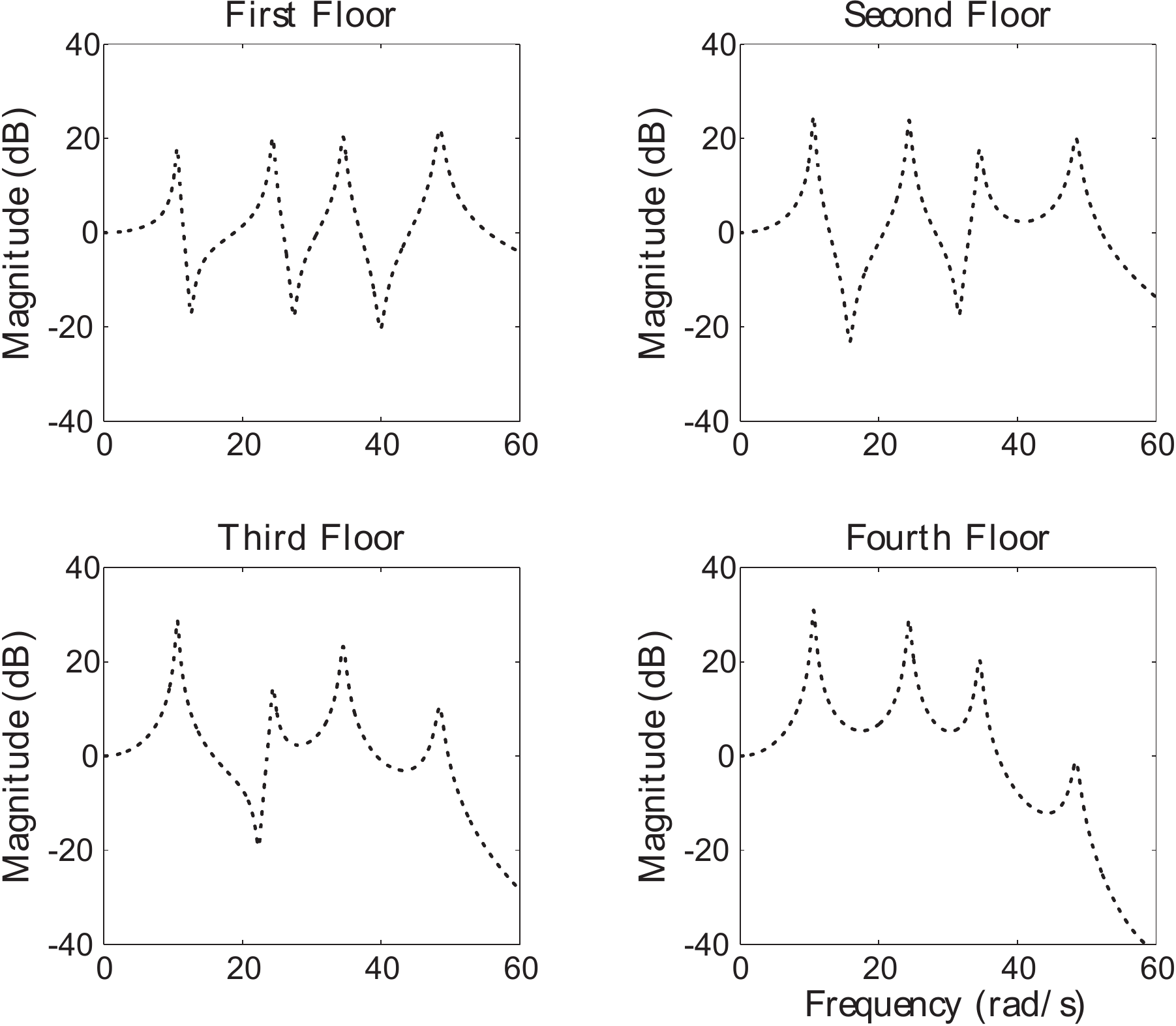}
\caption{FRF for the $N=4$ floors case without any TMD installed.}
\label{62Results4floorswithoutTMD}
\end{figure}

On the other hand, the parameters for the $N=4$ floor building problem are the following: i) $k_1=2000$ N/m, $k_2=1500$ N/m, $k_3=1000$ N/m and $k_4=500$ N/m, ii) $m_1=m_2=m_3=2$ kg and $m_4=1$ kg and iii) $\xi_s=0.01$. With these parameters, the natural frequencies and damping are $\omega_1=10.608$ rad/s, $\omega_2=24.380$ rad/s, $\omega_3=34.538$ rad/s, $\omega_4=48.479$ rad/s, $\xi_1=0.020$, $\xi_2=0.011$ and $\xi_3=\xi_4=0.010$. The FRF between the acceleration of each floor ($y_j$) and the acceleration of the ground ($a_g$) is shown in Figure \ref{62Results4floorswithoutTMD} (without any TMD). In this case, the restrictions for $\bf{x}$ are $\omega_{t,j}\in[0,50]$ rad/s, $\xi_{t,j}\in[0,0.3]$, $m_{t,j}\in[0,0.05]$ kg and $fb_j\in\left\{1,2,3,4\right\}$.

Table \ref{tab:62Comparacion_Resultados} shows the results obtained by the proposed CRO-SL, compared to different alternative algorithms. Specifically, all the algorithms that form the substrate layers in the CRO-SL approach has been tried on their own, with the same number of function evaluations, in order to show how the competitive co-evolution process promoted by the CRO-SL is positive to obtain better solutions for the TMD design and location problem. In Table \ref{tab:62Comparacion_Resultados} it can be seen how the CRO-SL obtains the best performance, both in the $N=2$ and $N=4$ cases. In the $N=2$ case, the differences among different methods are small, with the HS as the second best approach, and the 2Px and MPx crossover quite close behind. In this case, it seems that the DE and GM exploration patterns works worse than the other search procedures considered. In the case of $N=4$, the differences are larger. The proposed CRO-SL approach obtains the best result, and in this case DE operator also obtains a very good solution, close to the best obtained by the CROSL. The Mpx is the third best approach in this instance, whereas the GM and HS operators seem to work worse in this harder problem.

\begin{table}[!ht]
\begin{center}
\caption{\label{tab:62Comparacion_Resultados} Comparison of the results obtained in the two case-studies taken into account ($N=2$ and $N=4$) with different algorithms, in terms of the fitness function considered (Equation \ref{min11}).}
\begin{tabular}[t1]{ccccc}
\hline
& \multicolumn{2}{c}{2 floors}&\multicolumn{2}{c}{4 floors} \\
\hline
& Min & Mean & Min & Mean \\
\hline
\hline
CRO-SL & 8.4348 & 8.5773 & 7.7746 & 7.8747 \\
\hline
HS & 8.4728 & 8.5786 & 8.848 & 9.4393 \\
\hline
DE & 9.5405 & 10.1129 & 7.8831 & 7.9833\\
\hline
2Px & 8.5306 & 8.627 & 8.9341 & 8.9897\\
\hline
GM  & 8.9914 & 9.162 & 10.3464 & 11.3341 \\
\hline
MPx  & 8.7337 & 8.797 & 8.4458 & 8.9154 \\
\hline
\end{tabular}

\end{center}
\end{table}

The best result obtained by the CRO-SL in the case $N=2$ is the following:
\begin{eqnarray} \label{Res2F}
\boldsymbol{\Omega}_t=[22.6586,14.9481] \ \text{rad/s}, \\ \nonumber
\boldsymbol{\Xi}_t=[0.2939,0.1149], \\ \nonumber
\mathbf{M}_t=[0.0473,0.0500], \ \text{kg} \\ \nonumber
\mathbf{FB}=[2,2]. \nonumber
\end{eqnarray}

\begin{figure}[!ht]
\centering
\includegraphics[width=0.7\linewidth]{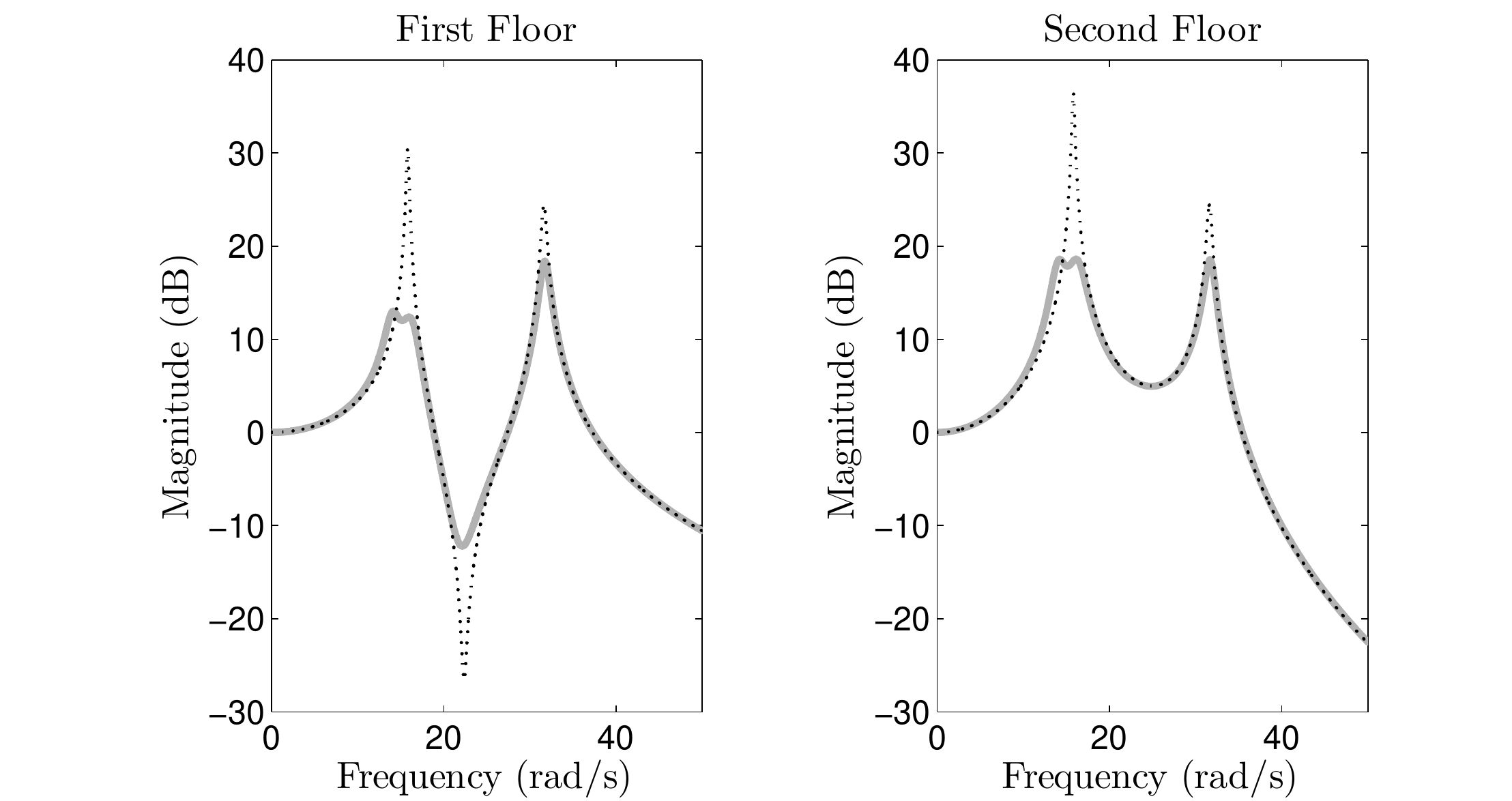}
\caption{FRF for the $N=2$ case, with the optimal position and design for the TMDs using the CRO-SL algorithm. (($\cdots$)-black) without TMDs and ((---)-gray) with TMDs.}
\label{Results2floors}
\end{figure}

\begin{figure}[!ht]
\begin{center}
\subfigure[]{\includegraphics[draft=false, angle=0,width=8cm]{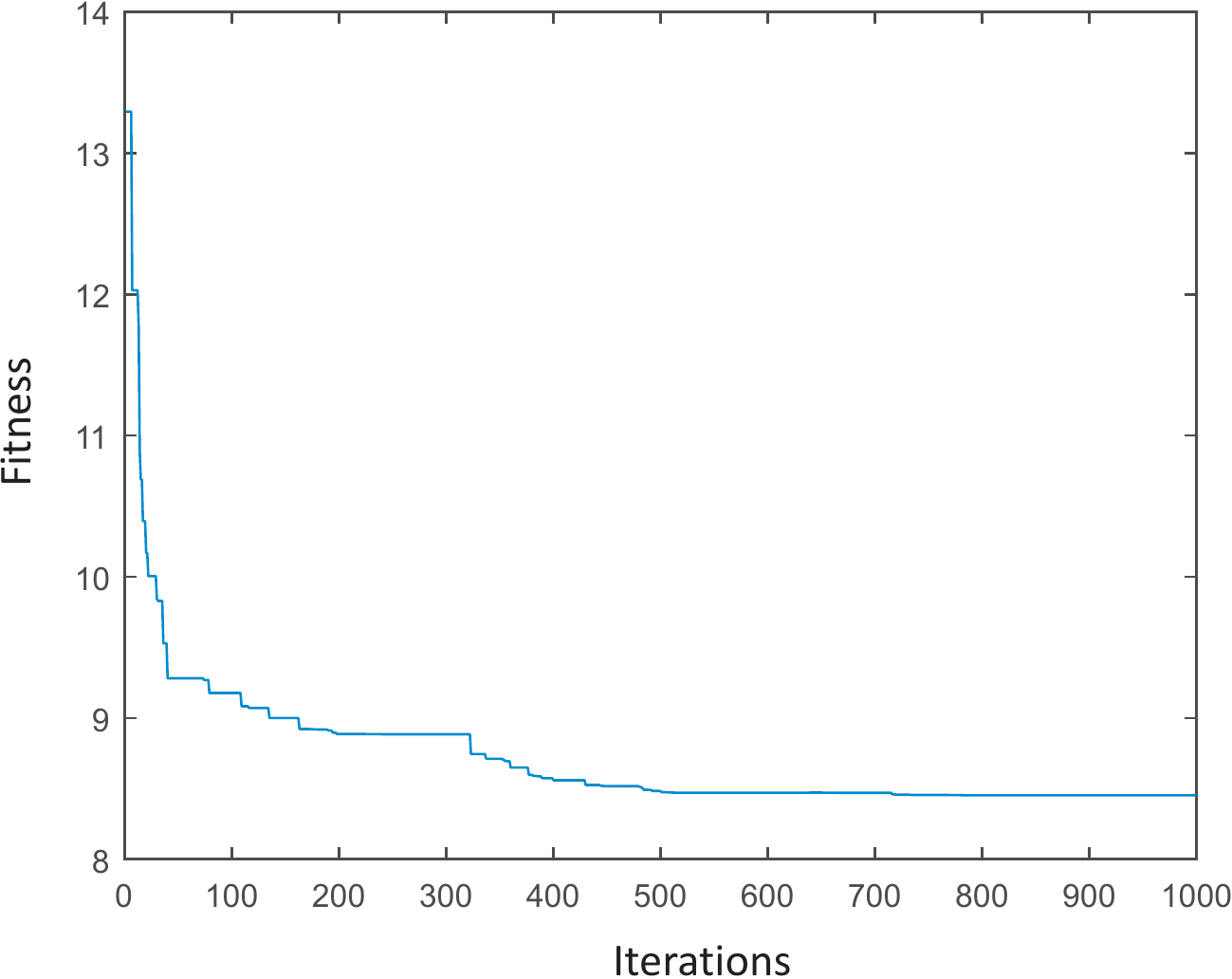}}
\subfigure[]{\includegraphics[draft=false, angle=0,width=8cm]{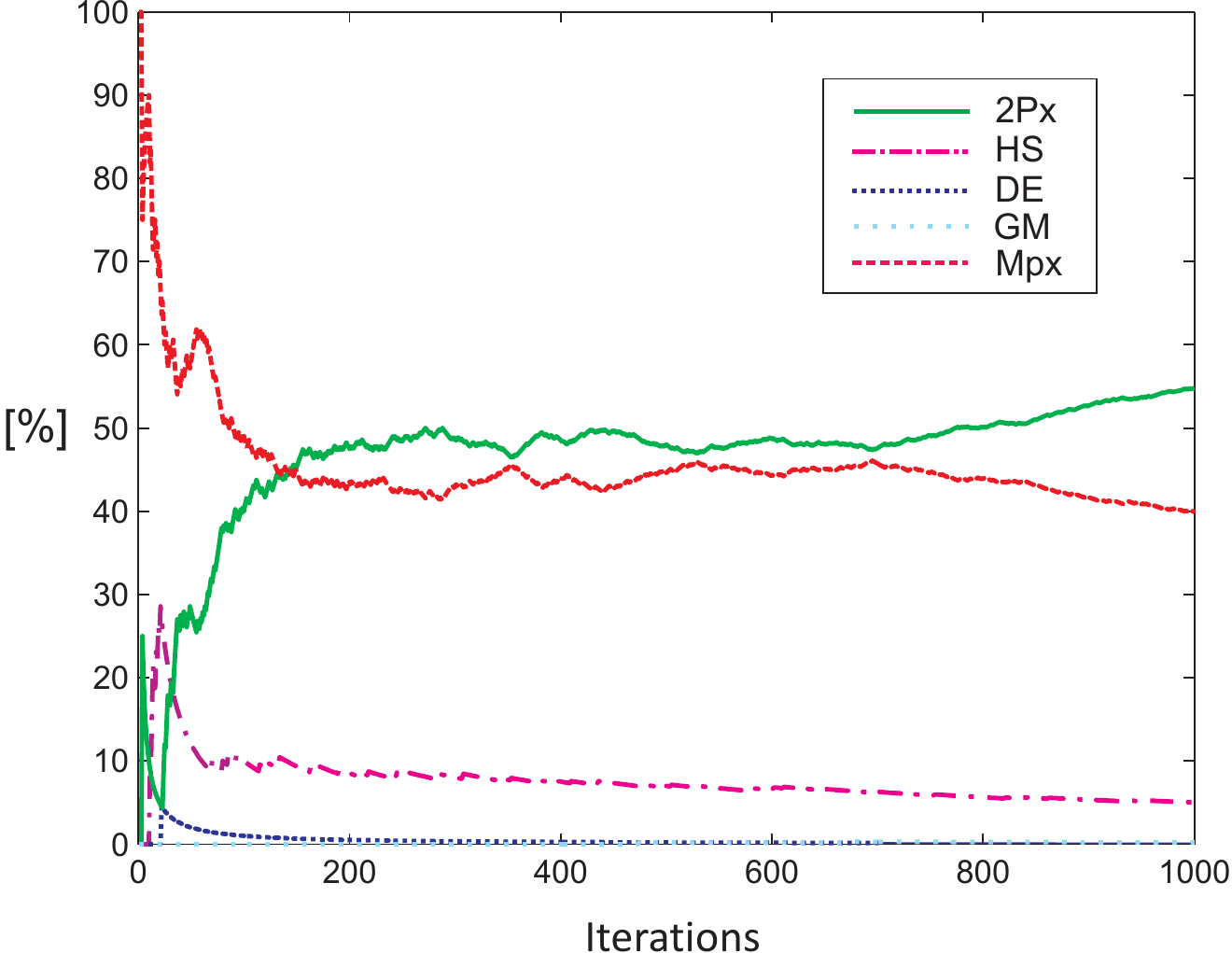}}
\end{center}
\caption{ \label{Evol_N2} Evolution of the best solution within the CRO-SL and ratio of times that each substrates produces the best larva in each iteration of the algorithm, for the $N=2$ floors TMDs location problem; (a) Best evolution; (b) Competitive ratio of the best solution found in each generation.}
\end{figure}

Regarding this best solution (see Figure \ref{Results2floors}), note the following: i) the two TMDs are located in the second floor, ii) the first TMD is syntonized to the first vibration mode and iii) the natural frequency of the second TMD is between the first and second vibration mode of the structure. It should be remarked that the maximum of the FRFs without TMDs is located in the second floor-first vibration mode (36.5dB). The maximum of the FRFs with TMDs is located in the second vibration mode, where the amplitudes are 18.4 and 18.6dB for the first and second floor, respectively. Therefore, the second TMD is syntonized to damp both vibration modes and to level the amplitude of the second vibration mode at both floors. Figure \ref{Evol_N2} (a) shows the evolution of the best solution found by the CRO-SL algorithm for the $N=2$ floors case. Note how the convergence of the proposed approach to the optimal solution is fast, in around 200 iterations. Note also that the CRO-SL is a co-evolution algorithm which evolves different exploration patterns in the substrate layers. The question is how to evaluate what is the substrate layer that contributes the most to the search in this problem of TMDs location and design. To clarify this point, Figure \ref{Evol_N2} (b) shows the ratio of times that every substrate generates the best larva per generation in the CRO-SL approach ($N=2$ floors case). It indicates that the MPx crossover and the 2-points crossover are the two exploration operators that contribute the most to the CRO-SL search. The HS substrate seems to contribute to the CRO-SL search as well, and the rest of operators (substrates) contribute very little, and only in the earliest stages of the algorithm. This behaviour is consistent to the performance of the different algorithms run on their own, as shown in Table \ref{tab:62Comparacion_Resultados}, where it was shown that the DE and GM searchers do not work well in this problem.

The example of TMD design and location with $N=4$ floors is more complex as optimization algorithm, since the search space is much larger than in the $N=2$ problem. In order to better motivate this application example, two different evaluations are carried out: first, a free-location of the 4 TMDs and their parameters, for $N=4$. Second, in order to compare this solution, we consider the case of an intuitive solution in which the four TMDs are located in the top floor, and only the rest of their parameters are sought with the proposed CRO-SL. The best solutions obtained by the CRO-SL in these cases are the following:

\begin{eqnarray} \label{Res4F}
\boldsymbol{\Omega}_t=[9.8264,10.5978,21.3608,31.8252]\ \text{rad/s}, \\ \nonumber
\boldsymbol{\Xi}_t=[0.0985,0.1070,0.2398,0.3000], \\ \nonumber
\mathbf{M}_t=[0.0500,0.0500,0.0500,0.0500], \ \text{kg} \\ \nonumber
\mathbf{FB}=[4,4,4,1]. \nonumber
\end{eqnarray}
in the first case (free TMD locations), and

\begin{eqnarray} \label{Res4F1}
\boldsymbol{\Omega}_t=[9.8887,10.3113,27.8916,46.8533] \ \text{rad/s}, \\ \nonumber
\boldsymbol{\Xi}_t=[0.3000,0.1107,0.1731,0.0055], \\ \nonumber
\mathbf{M}_t=[0.0500,0.0500,0.0500,0.0500], \ \text{kg} \\ \nonumber
\mathbf{FB}=[4,4,4,4], \nonumber
\end{eqnarray}
for the second evaluation problem in which TMD locations are pre-set.

\begin{figure}[!ht]
\centering
\includegraphics[width=1\linewidth]{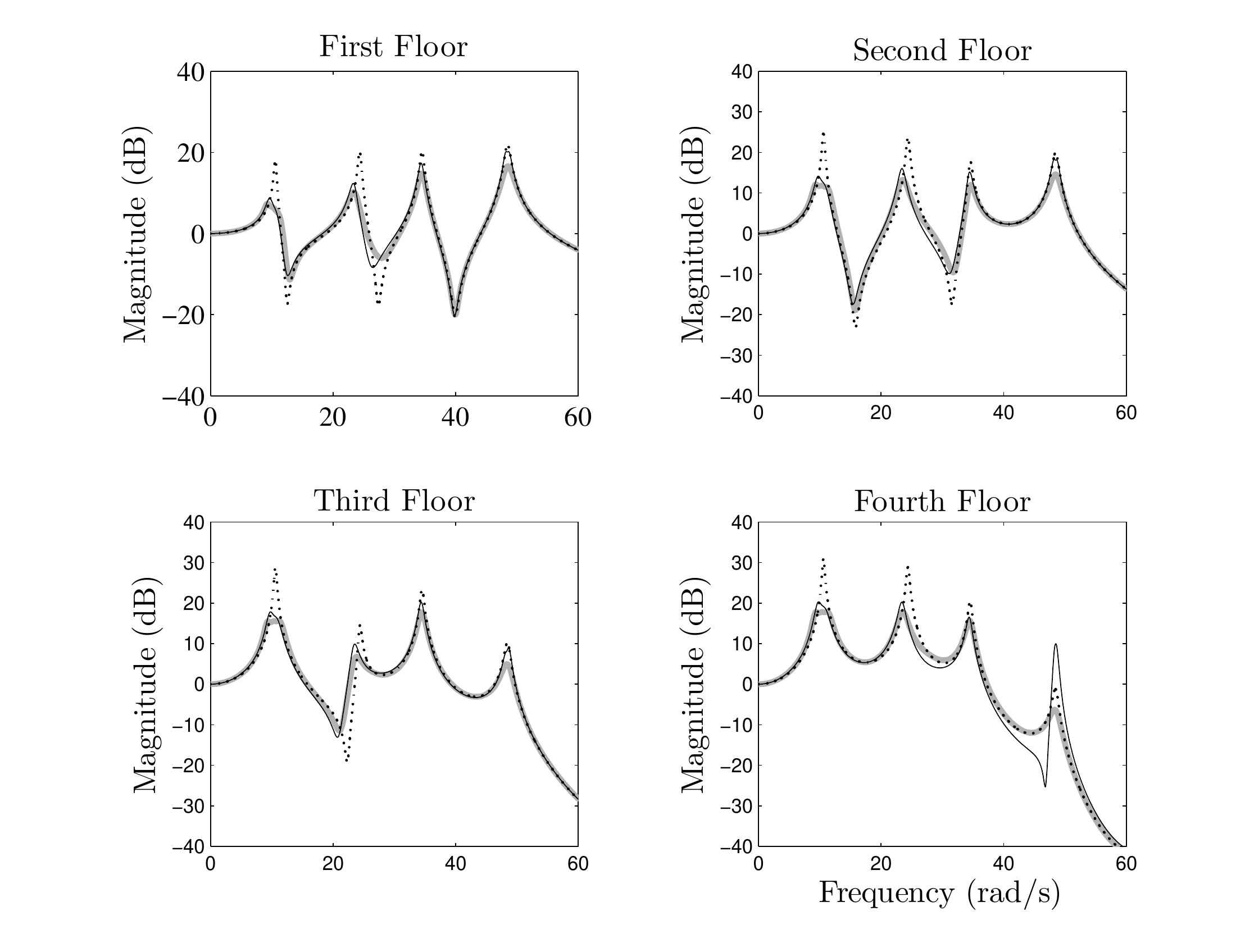}
\caption{FRF for the $N=4$ case, with the optimal position and design for the TMD using the CRO-SL algorithm. (($\cdots$)-black) without TMDs, ((---)-gray) with TMDs located at $\mathbf{FB}$=[4,4,4,1] and ((---)-black) with TMDs located at $\mathbf{FB}$=[4,4,4,4].}
\label{Results4floors}
\end{figure}

The results of the four storey building model ($N=4$) with the TMD optimal configuration of Equations (\ref{Res4F}) and (\ref{Res4F1}) are shown in Figure \ref{Results4floors}. The maximum of the FRFs is located in the fourth floor-first vibration mode without any TMD (30.9dB). The maximum of values of the FRFs, when the four TMDs are located as Equation (\ref{Res4F}), are in the third floor-third vibration mode, fourth floor-third vibration mode and fourth floor-fourth vibration mode (approximately equal to 17.8dB). The maximum values of the FRFs, when the four TMDs are located as Equation (\ref{Res4F1}), are in fourth floor-first vibration mode, fourth floor-third vibration, third floor-third vibration mode and first floor-fourth vibration mode (approximately equal to 20.2dB).

The following conclusions can be deduced from the comparison between both optimal solutions: i) both of them place two TMDs on fourth floor and their natural frequencies are close to the first vibration mode, ii) although both solutions place another TMD on fourth floor to reduce the second vibration mode, the natural frequency of the TMD corresponding to the first case (free TMD locations) is between first and vibration mode, which implies a better reduction in the first vibration mode with a less damping ratio, and iii) although it might seem unlikely, the last TMD is placed on first floor when there is freedom to locate TMDs, which improves the damping performance in the third and fourth vibration comparing with the design of all TMDs on the fourth floor. Therefore, although both evaluation problems show that the amplitude of the FRFs is reduced in all the vibration modes and in all the floors, the first case (free TMD locations in the CRO-SL) produces better results. Note that this illustrates an example where the optimum location is not the most obvious solution.

\begin{figure}[!ht]
\begin{center}
\subfigure[]{\includegraphics[draft=false, angle=0,width=7cm]{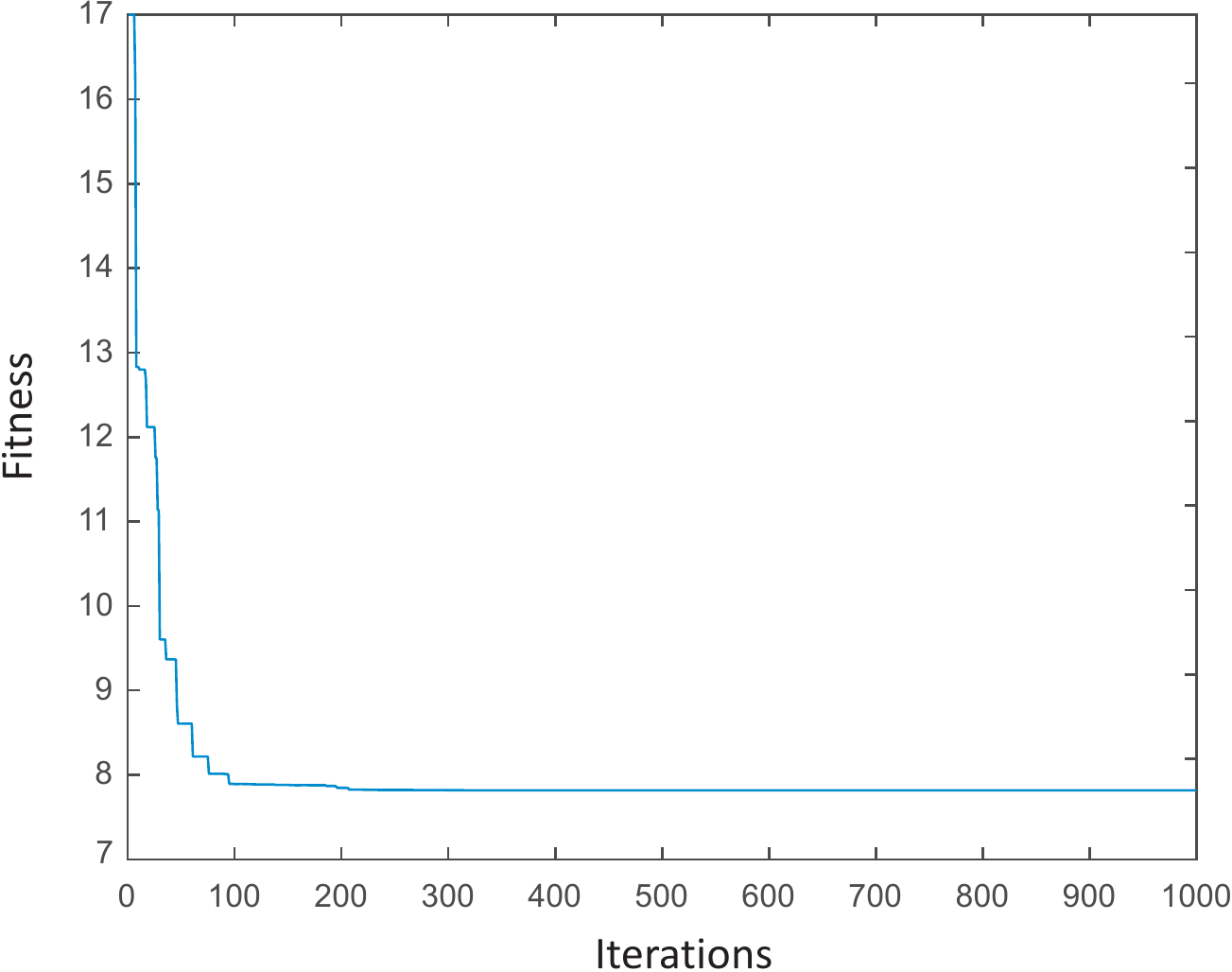}}
\subfigure[]{\includegraphics[draft=false, angle=0,width=7cm]{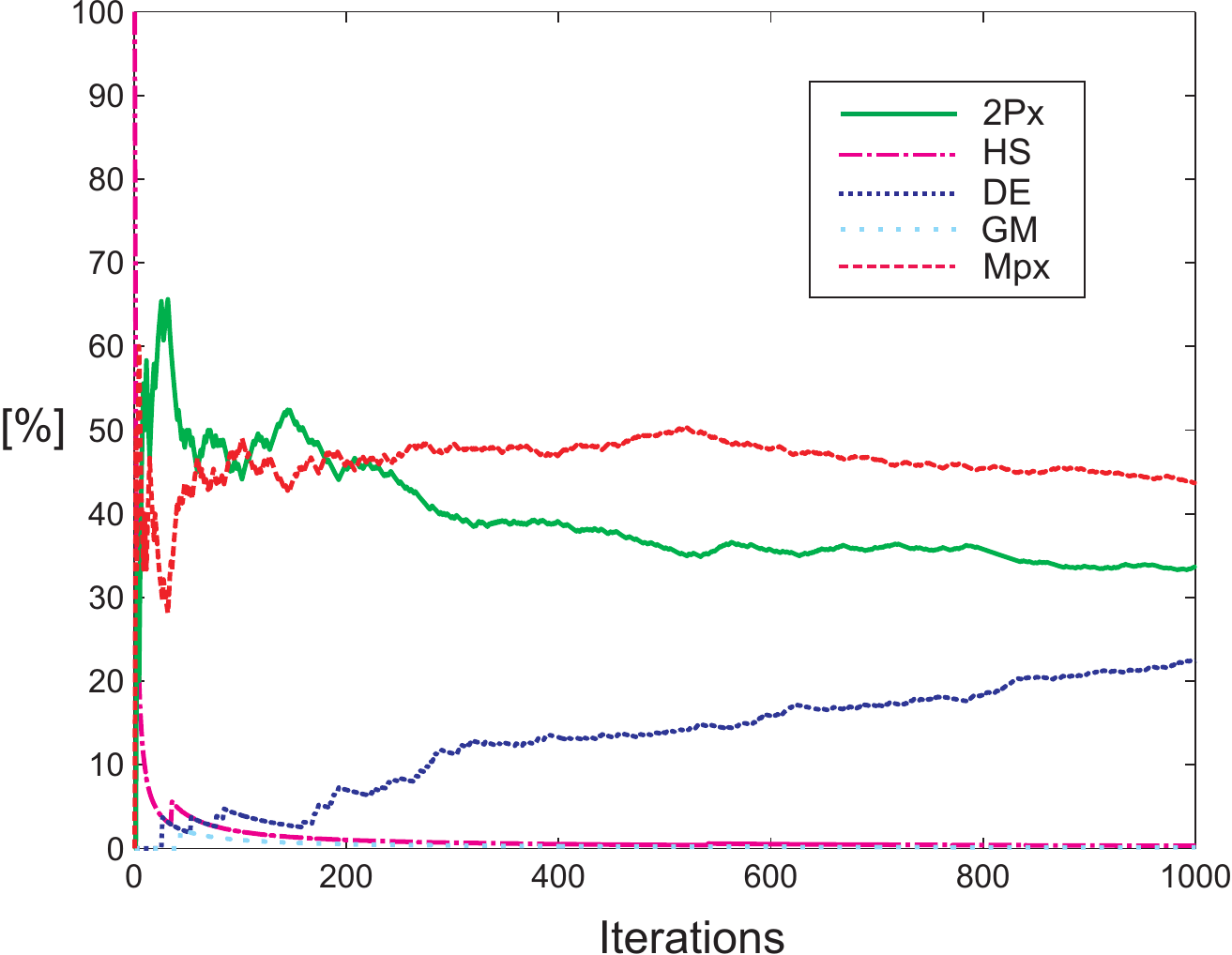}}
\end{center}
\caption{ \label{Evol_N4} Evolution of the best solution within the CRO-SL and ratio of times that each substrates produces the best larva in each iteration of the algorithm, for the $N=4$ floor TMD location problem; (a) Best evolution; (b) Competitive ratio of the best solution found in each generation.}
\end{figure}

Figure \ref{Evol_N4} (a) shows the evolution of the best solution found by the CRO-LS algorithm in this problem of TMD design and location for the case $N=4$ floors. Regarding the ratio of times that every substrate generates the best larva per generation in the CRO-SL approach, Figure \ref{Evol_N4} (b) shows it for this problem. It indicates that the MPx crossover and the 2-points crossover are again the two exploration operators that contribute the most to the CRO-SL search, but in this problem the DE substrate seems to contribute more than in the previous case, whereas the HS and Gaussian substrates barely contribute to obtain the best solutions in each iteration of the CRO-SL. This also coincides to the results reported in Table \ref{tab:62Comparacion_Resultados}.

The results reported in this work show that the CRO-SL is able to obtain excellent results for problems of TMD tuning and location, improving other meta-heuristics algorithms in this hard optimization problem in structures engineering.

\section{Experimental implementation}\label{Maqueta}

\begin{figure}[!ht]
\centering
\includegraphics[width=0.7\linewidth]{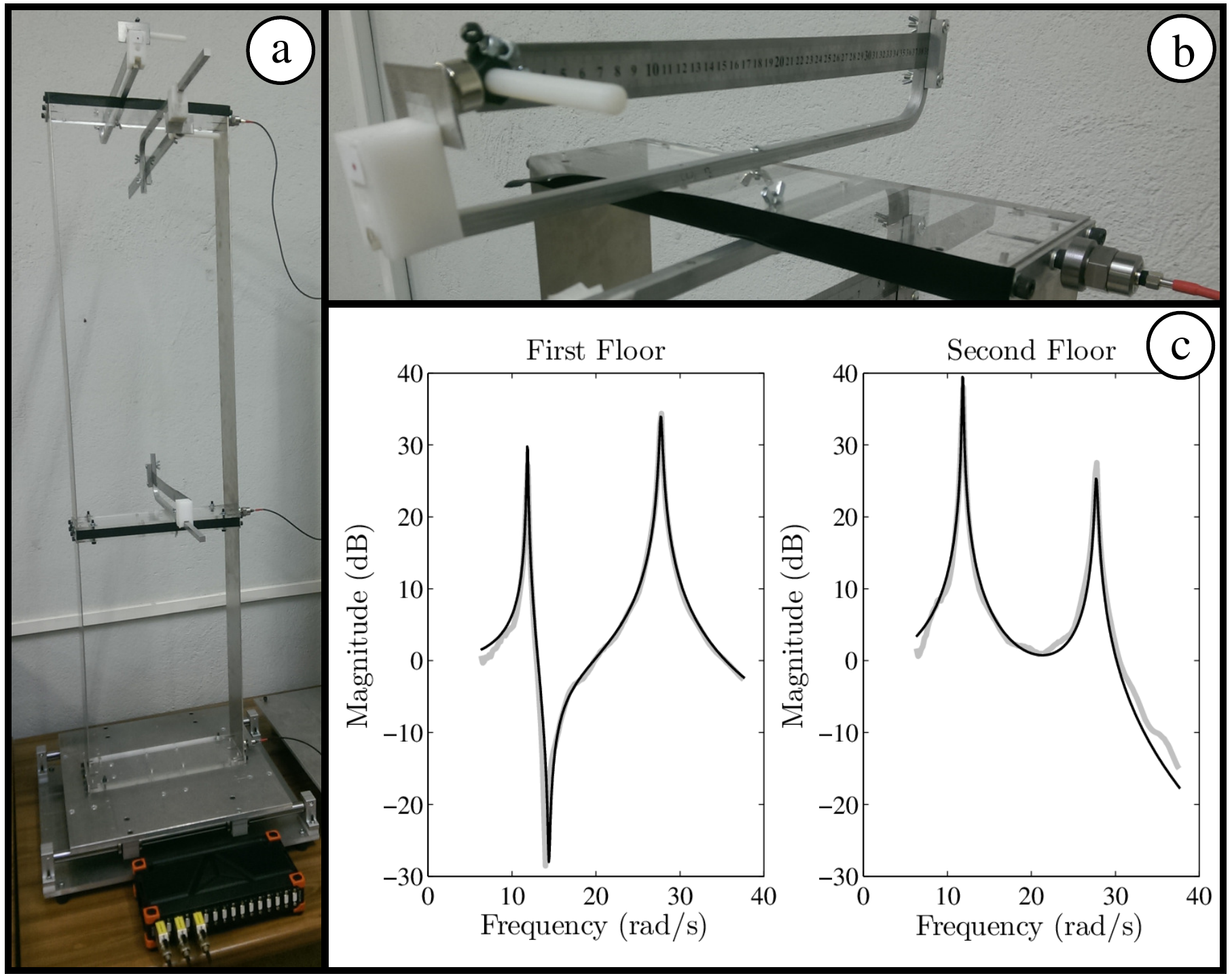}
\caption{Experimental set-up; a) $N=2$ floor building with TMD frames; b) Detail of the experimental TMD; c) Structure Identification, where ((---)-black) shows the FRF of the identified model, ((---)-gray) shows the experimental FRF.}
\label{Identification}
\end{figure}

The experimental set-up consists of a $N=2$ storey building. The parameters for the $N=2$ floor building are obtained  through experimental identification of the scale model shown in Figure \ref{Identification} (a), instaled on a sliding table. For the experimental identification methodology, three accelerometers (MMF-KS76C,  with 100 mV/g of sensitivity) attached to the ground, first floor and second floor, are connected to SIRIUS-HD-16xSTGS datalogger. The response under a soft impact applied to the sliding table is postprocessed using the Modal Testing tool available in DeweSoft X -DSA SP5. Sampling rate was set to 1000 Hz and the FRFs in Figure \ref{Identification} (c)-gray are obtained after averaging 5 impacts.

Values for the stiffness, mass and damping are $k_1=1111.8$ N/m, $k_2=389.1$  N/m, $m_1=2.14$ kg, $m_2=1.88$ kg and $\xi_s=0.006$. With these parameters, the natural frequencies and damping are $\omega_1=11.842$ rad/s, $\omega_2=27.733$ rad/s and $\xi_1=\xi_2=0.006$. Figure \ref{Identification} (c) shows a very good agreement between computational and experimental FRFs, revealing that the hypothesis of proportional damping matrix can work well in practical engineering application.

The laboratory TMDs comprise (see Figure \ref{Identification} (b))) a cantilever leaf spring with adjustable length mounted in the frame. At the end of the cantilever the magnet is in close proximity to an aluminum plate, also supported in the frame. The effective moving mass depends not only on the masses (and magnet) placed at the end of the cantilever but also on the length of the leaf, that has been adjusted to obtain the tuning frequency of the TMD. The damping coefficient is adjusted moving closer or away the aluminum plate with regards to the magnet. This set-up, whith 3 mounting frames, is ready to undergo any of the logical solution of the problem, which are both TMD in the upper floor or each one on a floor.

In order to better motivate this application example (like $N=4$ example), two different evaluations are carried out: first, a free-location of the two TMDs and their parameters, for $N=2$. Second, in order to compare this solution, we consider the case of an intuitive solution in which the two TMDs are located in the top floor, and only the rest of their parameters are sought with the proposed CRO-SL. The best solutions obtained by the CRO-SL in these cases are the following:

\begin{eqnarray} \label{Res2FExp1}
\boldsymbol{\Omega}_t=[23.3822,11.3105]\ \text{rad/s}, \\ \nonumber
\boldsymbol{\Xi}_t=[0.2000,0.1344], \\ \nonumber
\mathbf{M}_t=[0.100,0.100], \ \text{kg} \\ \nonumber
\mathbf{FB}=[1,2], \\ \nonumber
g(\bf{x})=7.5033,
\end{eqnarray}
in the first case (free TMD locations), and

\begin{eqnarray} \label{Res2FExp2}
\boldsymbol{\Omega}_t=[11.3408,26.6638] \ \text{rad/s}, \\ \nonumber
\boldsymbol{\Xi}_t=[0.1852,0.0460], \\ \nonumber
\mathbf{M}_t=[0.0100,0.0100], \ \text{kg} \\ \nonumber
\mathbf{FB}=[2,2], \\ \nonumber
g(\bf{x})=9.8443,
\end{eqnarray}
for the second evaluation problem in which TMD locations are pre-set.

\begin{figure}[!ht]
\centering
\includegraphics[width=1\linewidth]{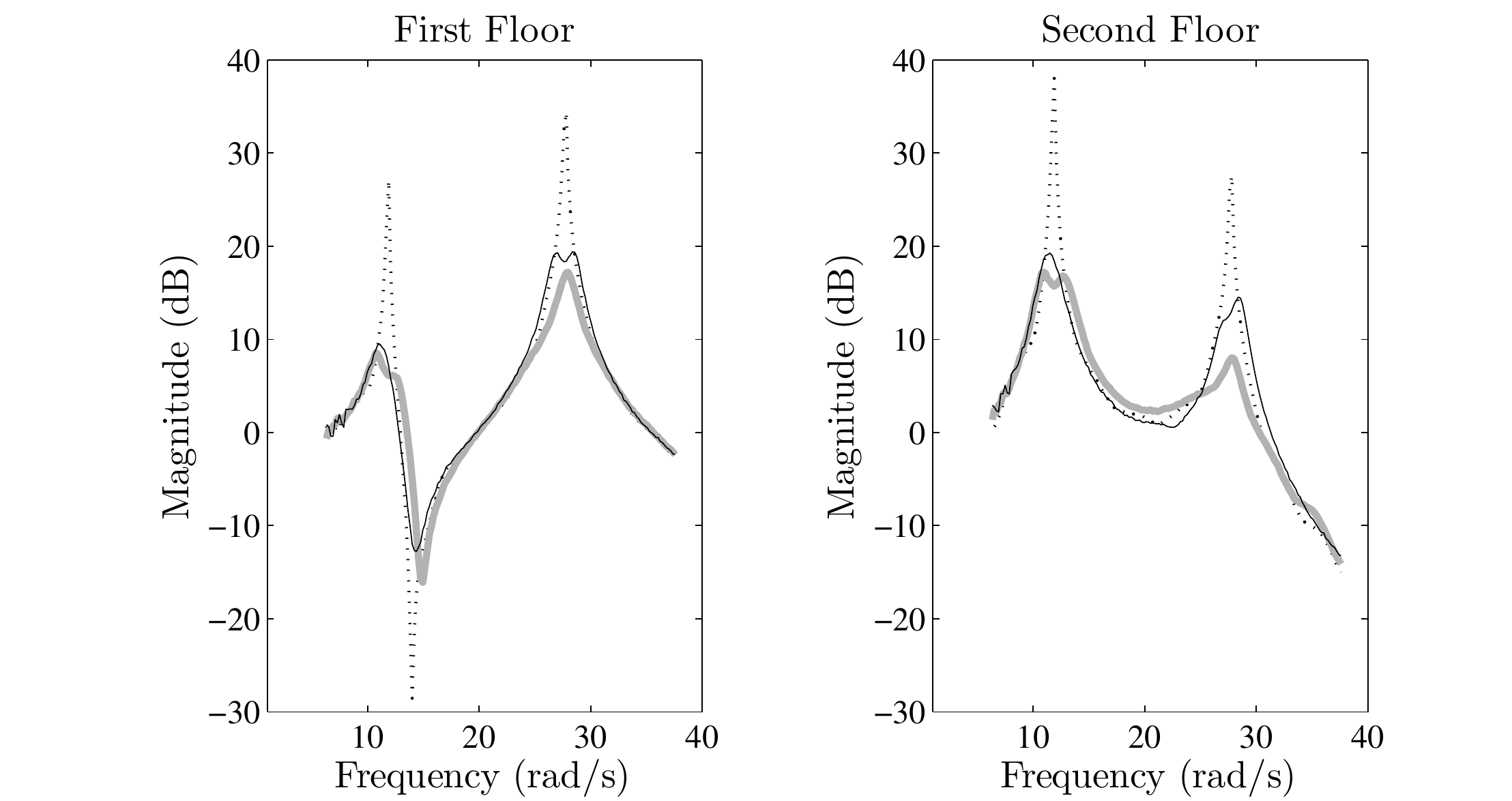}
\caption{Experimental FRF for the $N=2$ case, with the optimal position and design for the TMDs using the CRO-SL algorithm. (($\cdots $)-black) without TMDs,((---)-gray) with TMDs located at $\mathbf{FB}$=[2,1] and (((---)-black) with TMDs located at $\mathbf{FB}$=[2,2].}
\label{EXPTMD}
\end{figure}

The results of the experimental set-up with the TMD optimal configuration of Equations (\ref{Res2FExp1}) and (\ref{Res2FExp2}) are shown in Figure \ref{EXPTMD}. Note that the intuitive solution in which the two TMDs are located in the top floor is worst than the one in which a first TMD placed on the second floor and tuned to the first mode and the second one is placed on the first floor. It is also noteworthy to realize that the second TMD is not really tuned to the second mode but placed in a frequency between the one of the first and the second mode. Note the good agreement, regardless of the difficulty in adjusting the values of the moving mass and the damping coefficient obtained by the optimization algorithm (i.e., the experimental values of the functions, $g(\bf{x})$, are approximately the same).

The maximum of the FRFs is located in the second floor-first vibration mode without any TMD (38dB). Although the maximum values of the FRFs for both designs are in the first floor-second vibration mode and second floor-first vibration mode, these maximum values are 17.2 dB and 19.3 dB when the two TMDs are located as Equations (\ref{Res2FExp1}) and (\ref{Res2FExp2}), respectively.

\section{Conclusions}\label{62Conclusions}

 This chapter discusses a problem of design of TMDs for vibration control in structures. In order to apply the CRO-SL in this problem, a modification of the generalized framework presented in \cite{Mohtat2011} has been used to formulate a $N$ floor building where $M$ TMDs must be installed. The proposed modification allows the optimization algorithm deciding the position of each TMD, in such a way that a TMD can be placed at any floor to damp any vibration mode. We have shown that the CRO-SL is an excellent approach to solve optimization problems related to the design and optimal location of TMDs in structures, by solving two case studies of TMD location in two building models with two and four floors, respectively, within a low computation time. Finally, an experimental set-up has been made to test the algorithm by using a model of the two floors building. These experimental results show that: i) the generalized framework proposed herein can work well in practical engineering applications, ii) the mass, stiffness and damping values of each TMDs can be accurate tuned with the laboratory equipment used.

\chapter{Active Vibration Control design using inertial mass actuators}\label{cap:avcs}

\section{Introduction and state of the art}
Improvements in design and construction have led to light and slender floor structures, which have, in turn, increased susceptibility to vibrations. These structures satisfy ultimate limit state criteria but have the potential of attracting complaints coming from excessive human-induced vibrations. Active vibration control via inertial mass actuators has been shown to significantly reduce the level of response, allowing structures to satisfy vibration serviceability limits \cite{Hudson12}. This problematic may seem to be similar to the TMDs optimal design and location problem tackled in the previous chapter, however in this approach the vibration control is active and is focused to vibrations induced by humans, not to ground motions.

Single-input single-output (SISO) strategies based on collocated control (i.e., the pair sensor/actuator are placed physically at the same point) are widely used. However, a better performance can be achieved if a multi-input multi-output (MIMO) control strategy is used. This was firstly shown in \cite{Hanagan2000}, in which an optimal Direct output Velocity Feedback MIMO controller was presented. This DVF-MIMO control strategy finds the optimal gain matrix and the optimal location for a predefined number of actuators and sensors. The optimal sensor/actuator placement and the gain matrix are obtained by minimizing a Performance Index (PI) that considers the amplitude and duration of the vibration, and the maximum force imparted by each actuator. Simulation results were presented in \cite{Hanagan2000}, demonstrating the advantages of using MIMO control as opposed to SISO control.

In \cite{Pereira2014}, the PI proposed in \cite{Hanagan2000} was used to experimentally implement a MIMO AVC. Furthermore, the frequency bandwidth where humans perceive the vibration \cite{ISO2631} was also considered in \cite{Pereira2014} to focus the control effort on the most important vibration modes. However, unlike \cite{Hanagan2000}, the MIMO AVC proposed in \cite{Pereira2014} takes into account practical considerations, such as the spillover effects due to high-frequency components \cite{Griggs2007}, the actuator dynamics and its nonlinearities limitations due to stroke and force saturations. The spillover effects were reduced by considering low-pass filters and the stroke and force saturations were mitigated by including high-pass filters. This approach has been successfully implemented in practice on an indoor walkway sited at Forum building at the University of Exeter (Exeter, UK). The algorithm presented in \cite{Pereira2014}, which is based on a local gradient-based method, is useful when the number of test points is small. However, problems in structural optimization are often characterized by search spaces of extremely high dimensionality and nonlinear objective functions. In these optimization problems, classical approaches do not lead to good solutions, or in many occasions they are just not applicable, due to the unmanageable search space structure or its huge size, which implies an extremely high computation cost (i.e, the computation time to obtain a local solution may be months). Then only a few of test points and/or actuator/sensor pairs can be considered within these optimization processes. In this context, modern optimization meta-heuristics have been successfully applied to an important number of structural optimization problems. Modern optimization meta-heuristics have been lately the core of a huge research work, that have been mentioned in the previous chapter. Apart from the researches mentioned in Chapter \ref{cap:tmds}, there are other meta-heuristics previously applied to structure engineering problems as the Thermal Exchange optimization \cite{Kaveh17}.

In this research, the CRO-SL is applied to design MIMO-AVC for structures subjected to human induced vibration.  This optimization algorithm is particularly interested when vibrations on complex floor structures with several closely-frequency spaced vibration modes have to be cancelled. Thus, this algorithm promotes a powerful evolutionary-like search, ideal for solving high-burden optimization problems, which will be shown to be very effective in this particular problem of MIMO-AVC design. The proposed algorithm's performance has been evaluated and compared with several reference algorithms in a finite element (FE) model of a complex floor structure.

\section{Problem definition} \label{sec:63Model}

\begin{figure}[!ht]
	\begin{center}
		\includegraphics[width=0.95\columnwidth]{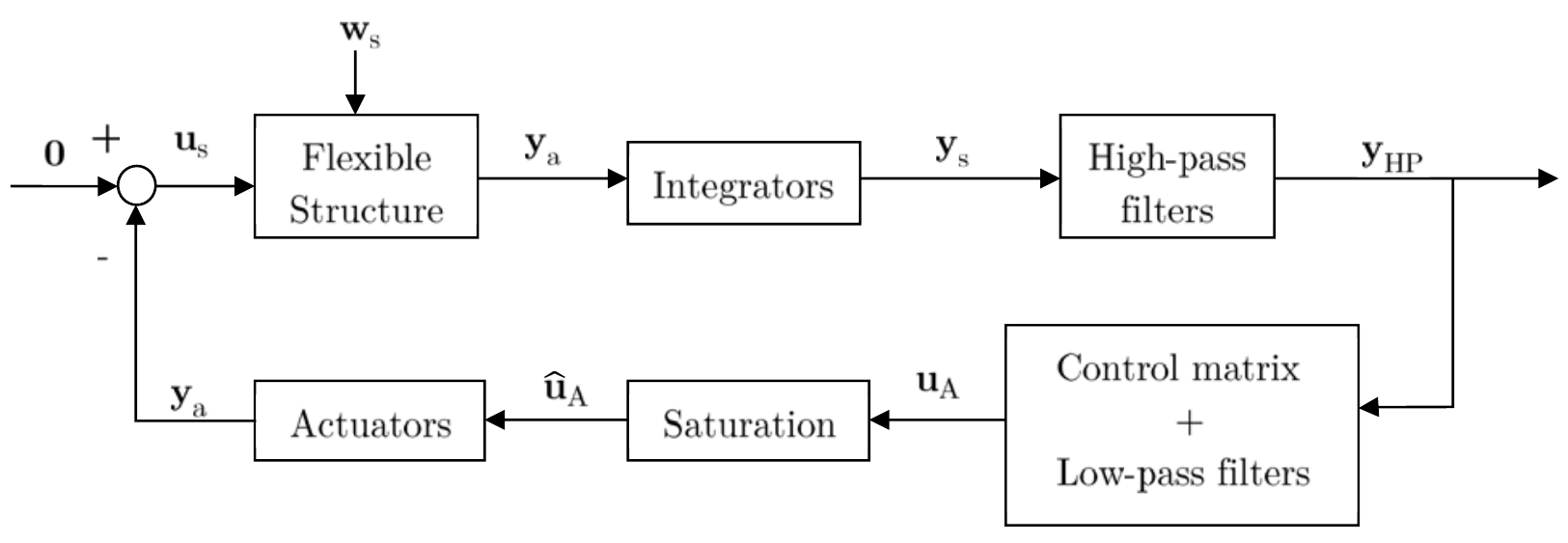}
		\caption{General control scheme.}
		\label{fig1}
	\end{center}
\end{figure}

The problem tackled in this work consists of finding the optimum locations and control gains of the AVC MIMO control strategy presented in \cite{Pereira2014}. This section explains the general scheme shown in Figure \ref{fig1} and how to formulate the cost function to use the CRO-SL as solver. In addition, the FE floor structure model, the AVC design methodology and the optimization problem are also described in this section.

\subsection{Floor structure}
\begin{figure}[!ht]
	\begin{center}
		\includegraphics[width=0.85\columnwidth]{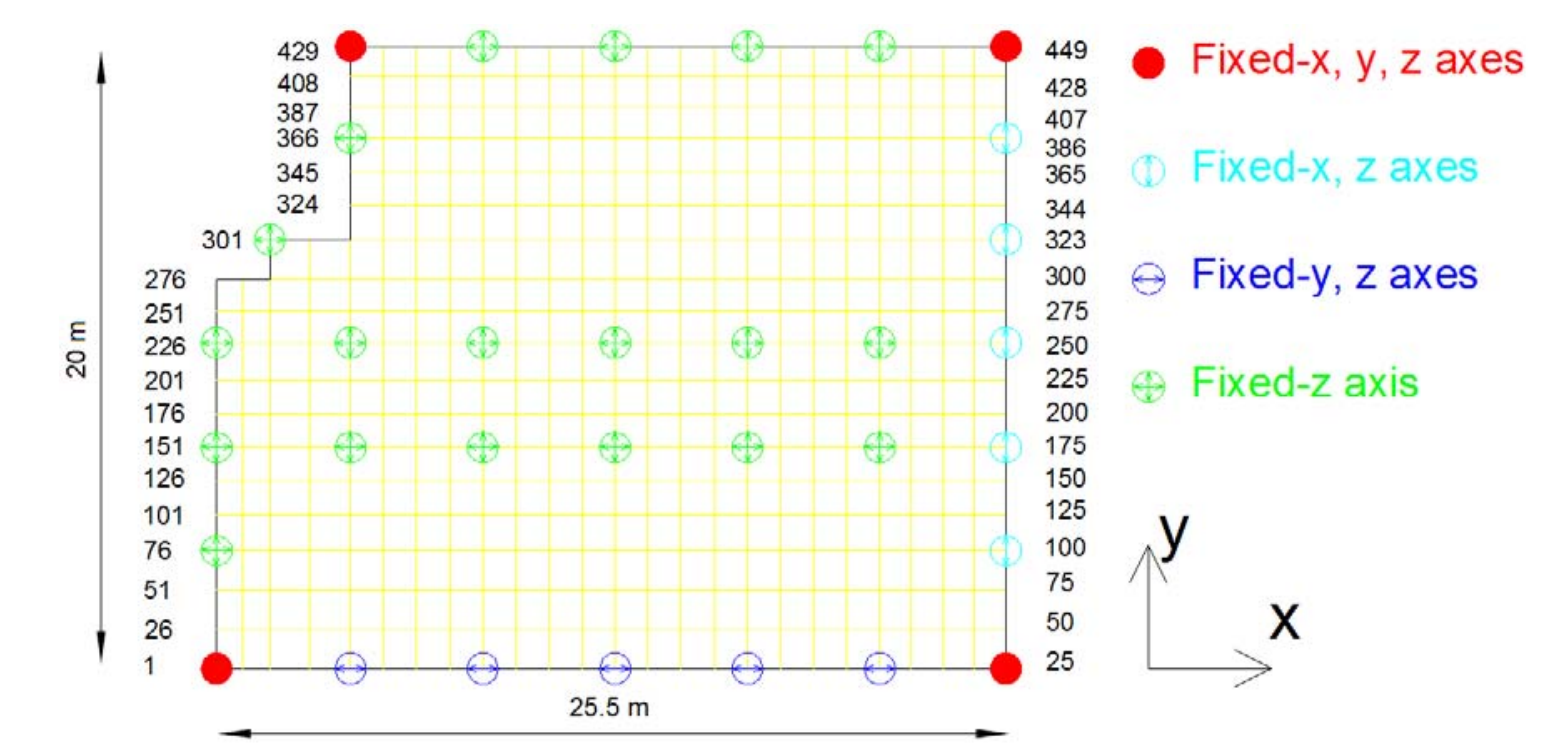}
		\caption{Floor structure considered.}
		\label{fig:floor}
	\end{center}
\end{figure}

\begin{figure}[!ht]
	\begin{center}
		\includegraphics[width=0.75\columnwidth]{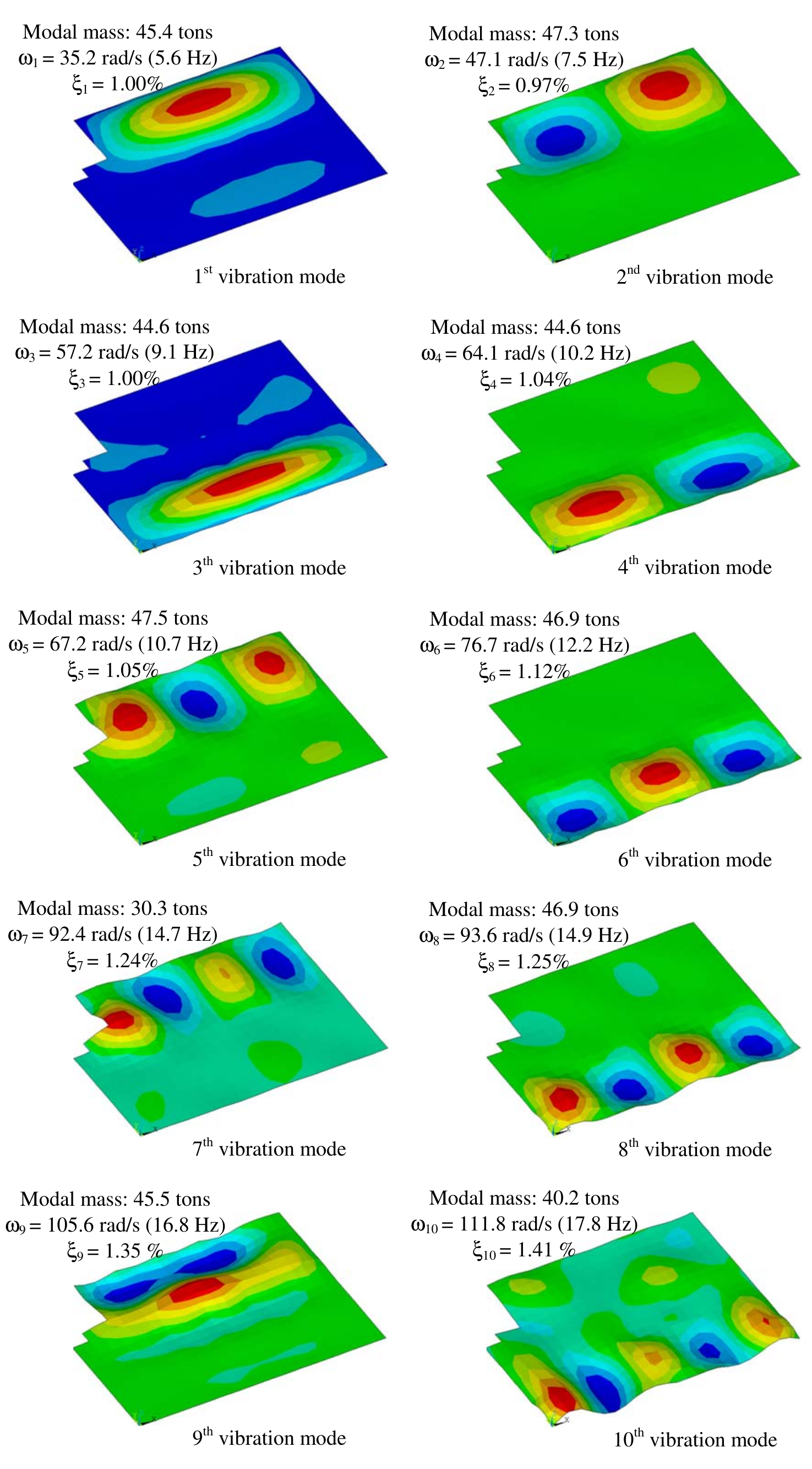}
		\caption{Floor mode shapes, natural frequencies, damping ratios and modal masses.}
		\label{fig:floorMS}
	\end{center}
\end{figure}

The structure considered in this research is a dining room floor of a primary-secondary school sited in Madrid (Spain). The general arrangement of beams and pillars is shown in Figure \ref{fig:floor}. The FE model was created in ANSYS \cite{Ansys10} using shell elements and 449 nodes. It is an irregular rectangular composite floor with the dimension of 25.5 m$\times$20 m$\times$0.3 m. As is shown in Figure \ref{fig:floor}, the floor is supported by 33 columns. Different connections between the floor and columns are marked with different colours: red ones represent those whose displacements in x, y and z directions are all restricted; cyan ones that are restricted in x and z displacements; blue ones that are restricted in y and z displacements; green ones are connections only restricted in z direction. None of the connections between columns and deck are restricted in rotations. The yellow lines show the meshing grids of the shell elements in ANSYS \cite{Ansys10}. The material properties considered are: modulus of elasticity $E=20 \times 10^{9}$ N/m$^{2}$, Poisson's ratio $\nu = 0.15$ and density $\rho = 3000$ kg/m$^{3}$. The density has been increased from 2500 kg/m$^{3}$ to 3000 kg/m$^{3}$ in order to include a portion of the imposed load (approximately 30$\%$) and the total dead load, following the recommendation of \cite{Smith07} for analysis of floor vibrations. The modal shapes, natural frequencies, damping ratios and modal masses of the first ten vibration modes can be seen in Figure \ref{fig:floorMS}.

For the sake of simplicity, the flexible structure and the integrators are grouped, so that the output of the resulting system is $\textbf{y}_s$, which is the velocity at \textit{q} locations. Thus, the standard state-space representation of the model for this flexible structure (with $n$ vibration modes, $p$ actuators, $q$ sensors and $r$ perturbations) is represented as follows (see Figure \ref{fig1}):

\begin{equation}
\label{eq1}
	\begin{split}
		\mathbf{\dot{x}}_s=	&	\mathbf{A}_s\mathbf{x}_s+\mathbf{B}_{s_1}\mathbf{u}_s+\mathbf{B}_{s_2}\mathbf{w}_s,\\
		\mathbf{y}_s=		&	\mathbf{C}_s\mathbf{x}_s,
	\end{split}
\end{equation}

If Equation (\ref{eq1}) is defined in modal coordinates, the state-space matrices are as follows \cite{Gawronski2004}:

\begin{equation}
\label{eq2}
	\begin{split}
		\mathbf{A}_s=	&	\begin{bmatrix}	\mathbf{0}	&	\mathbf{I}\\	-\mathbf{\Omega}^2	&	-2\mathbf{Z}\mathbf{\Omega}	\end{bmatrix},\mathbf{B}_{s_1}=	\begin{bmatrix}	\mathbf{0}\\	\mathbf{\Phi}_{u}	\end{bmatrix},\\
		\mathbf{B}_{s_2}=	&	\begin{bmatrix}	\mathbf{0}\\	\mathbf{\Phi}_{w}	\end{bmatrix},\mathbf{C}=	\begin{bmatrix}	\mathbf{\Phi}_{y}	&	\mathbf{0}	\end{bmatrix},
	\end{split}
\end{equation}
\noindent in which $\mathbf{\Omega}$ and $\mathbf{Z}$ are, respectively, the diagonal matrices formed by the natural frequencies ($[\omega_1,\cdots,\omega_n]$) and the damping ratios ($[\xi_{1},\cdots,\xi_{n}]$); and the matrices $\mathbf{\Phi}_u$, $\mathbf{\Phi}_y$ and $\mathbf{\Phi}_w$ are matrices with dimensions $n\times p$, $q\times n$ and $n\times r$. It is important to highlight that each \textit{k}$^{th}$ column of $\mathbf{\Phi}_u$ and $\mathbf{\Phi}_w$ and each row of $\mathbf{\Phi}_y$ is formed by the \textit{k}$^{th}$ vibration mode values at the positions of the actuators ($\mathbf{\Phi}_u$), perturbations ($\mathbf{\Phi}_w$) and sensors ($\mathbf{\Phi}_y$), respectively. The state vector is defined as: $\mathbf{x}_{s}=[x_{s_1},\cdots,x_{s_n},\dot{x}_{s_1},\cdots,\dot{x}_{s_n}]$ , where $[x_{s_1},\cdots,x_{s_n}]$ are the modal coordinates of the structure, and $[\dot{x}_{s_1},\cdots,\dot{x}_{s_n}]$ are their derivatives.

\subsection{AVC methodology}

The AVC is implemented by using (see Figure \ref{fig1}): i) inertial actuators to generate the forces through the acceleration of inertial masses to the structure on which they are placed, ii) second-order Butterworth high-pass filters to reduce the gain of the loop at low-frequencies, which can saturate the actuator, iii) second-order Butterworth high-pass filters to reduce the gain of the loop at high-frequencies, reducing thus the risk of spillover problems \cite{Griggs2007}, and iv) force saturation blocks to simulate the maximum voltage inputs of the inertial actuators. Note that if the cut-off frequencies of the filters are separated from the vibration modes, the AVC can be approximated by a DVF.

The AVC methodology consists of the two following main steps: i) define the cut-off frequencies of the filters according to the vibration modes of the structure and to the bandwidth of the actuators and ii) decide the locations of $p$ inertial actuators and $q$ sensors (i.e., the values of $\mathbf{\Phi}_{u}$ and $\mathbf{\Phi}_{y}$ in Equation (\ref{eq2})) and tune the matrix gain in order to minimize a cost function.

\subsubsection{Actuator dynamics}

The actuator consists of an inertial (or moving) mass $m_A$ attached to a current-carrying coil moving in a magnetic field created by an array of permanent magnets. The inertial mass is connected to the frame by a suspension system. The mechanical part is modelled by a spring stiffness $k_A$ and a viscous damping $c_A$. The electrical part is modelled by the resistance \textit{R}, the inductance of the coil \textit{L} and the voice coil constant $C_E$, which relates coil velocity and the back electromotive force (Figure \ref{Actuator}(a)) \cite{A.Preumont2011}. Combining the mechanical and the electrical part, the linear behaviour of the actuator can be closely described as a third-order dynamic model. Thus, the state space model of the \textit{p} actuators is as follows:

\begin{figure}[!ht]
\begin{center}
\includegraphics[width=0.9\linewidth]{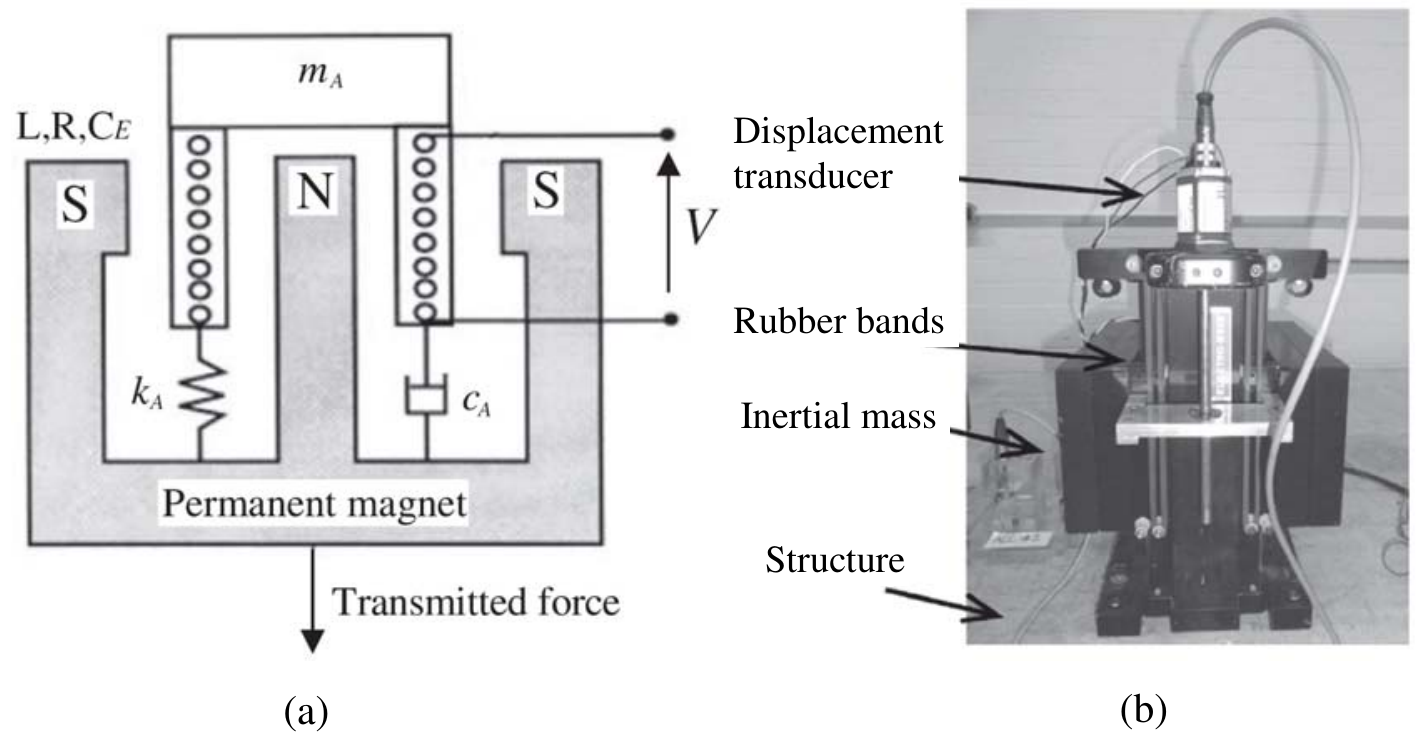}
\end{center}
\caption{Inertial-mass actuator; (a) Sketch of typical electrodynamic inertial actuator and (b) APS Dynamic Model 400 Shaker.}
\label{Actuator}
\end{figure}

\begin{equation}
\label{eq3}
	\begin{split}
		\mathbf{\dot{x}}_A=	&	\mathbf{A}_{A_T}\mathbf{x}_A+\mathbf{B}_{A_T}\hat{\mathbf{u}}_A,\\
		\mathbf{y}_A=		&	\mathbf{C}_{A_T}\mathbf{x}_A,
	\end{split}
\end{equation}

\noindent in which matrices $\mathbf{A}_{A_T}=diag(\mathbf{A}_{A},\cdots,\mathbf{A}_{A})$, $\mathbf{B}_{A_T}=diag(\mathbf{B}_{A},\cdots,\mathbf{B}_{A})$ and $\mathbf{C}_{A_T}=diag(\mathbf{C}_{A},\cdots,\mathbf{C}_{A})$ are block diagonal, in which matrices $\mathbf{A}_{A}$, $\mathbf{B}_{A}$ and $\mathbf{C}_{A}$ are defined as follows \cite{Diaz2010}:

\begin{equation}
\label{eq4}
	\mathbf{A}_A=\begin{bmatrix}	0	&	0	&	-\varepsilon\omega_{A}^{2} \\	1	&	0	&	-(\omega_{A}^{2}+2\xi_{A}\omega_{A}\varepsilon)	\\	0	&	1	&	-(\varepsilon+2\xi_{A}\omega_{A})	\end{bmatrix},\mathbf{B}_{A}=	\begin{bmatrix}	0\\	0\\	g_{A}	\end{bmatrix},\mathbf{C}_{A}=	\begin{bmatrix}	0	&	0	&	1	\end{bmatrix},
\end{equation}

\noindent in which the actuator is defined by $g_{A}>0$, its damping ratio $\xi_{A}$ and natural frequency $\omega_{A}$. The value of $\varepsilon$ models the low-pass properties of the actuator. The actuator considered in this work is an APS Dynamics model 400 electrodynamic shaker, which is shown in Figure \ref{Actuator}(b). The identified parameters of \ref{eq4} are \cite{Diaz2010}: $\omega_{A}=13.2$ rad/s (2.1Hz), $g_{A}=12,000$ and $\varepsilon=47.1$.

\subsubsection{Filters design}

The cut-off frequencies of the high-pass filters (denoted by $\omega_{HP}$) is the result of the tradeoff between the resonance frequency of actuator, since small values of $\omega_{HP}$ increase the risk of stroke saturation, and the first vibration mode of the structure, since higher values of $\omega_{HP}$ reduce the damping imparted by a DVF controller \cite{Pereira2014}. The state-space model of each high-pass filter is as follows:

\begin{equation}
\label{eq5}
	\begin{split}
		\mathbf{\dot{x}}_{HP}=	&	\mathbf{A}_{HP_T}\mathbf{x}_{HP}+\mathbf{B}_{HP_T}\mathbf{y}_{s},\\
		\mathbf{y}_{HP}=		&	\mathbf{C}_{HP_T}\mathbf{x}_{HP}+\mathbf{D}_{HP_T}\mathbf{y}_{s},
	\end{split}
\end{equation}
in which matrices $\mathbf{A}_{HP_T}=diag(\mathbf{A}_{HP},\cdots,\mathbf{A}_{HP})$, $\mathbf{B}_{HP_T}=diag(\mathbf{B}_{HP},\cdots,\mathbf{B}_{HP})$, $\mathbf{C}_{HP_T}=diag(\mathbf{C}_{HP},\cdots,$ $\mathbf{C}_{HP})$ and $\mathbf{D}_{HP_T}=diag(1,\cdots,1)$ are block diagonal, in which $\mathbf{A}_{HP}$, $\mathbf{B}_{HP}$ and $\mathbf{C}_{HP}$ are defined as follows \cite{MiddlehurstJack/Harrison1993}:

\begin{equation}
\label{eq6}
	\mathbf{A}_{HP}=\begin{bmatrix}	0	&	1 \\	-\omega^{2}_{HP}	&	-\sqrt{2}\omega_{HP}	\end{bmatrix},\mathbf{B}_{HP}=	\begin{bmatrix}	0\\	1	\end{bmatrix},\mathbf{C}_{HP}=	\begin{bmatrix}	-\omega^{2}_{HP}	&	-\sqrt{2}\omega_{HP}	\end{bmatrix}.
\end{equation}

The low-pass filters to avoid spillover problems \cite{Griggs2007} are defined as follows:
\begin{equation}
\label{eq7}
	\begin{split}
		\mathbf{\dot{x}}_{LP}=	&	\mathbf{A}_{LP_T}\mathbf{x}_{LP}+\mathbf{B}_{LP_T}\mathbf{y}_{HP},\\
		\mathbf{y}_{LP}=		&	\mathbf{C}_{LP_T}\mathbf{x}_{LP},
	\end{split}
\end{equation}
in which matrices $\mathbf{A}_{LP_T}=diag(\mathbf{A}_{LP},\cdots,\mathbf{A}_{LP})$, $\mathbf{B}_{LP_T}=diag(\mathbf{B}_{LP},\cdots,\mathbf{B}_{LP})$ and $\mathbf{C}_{LP_T}=diag(\mathbf{C}_{LP},\cdots,$
\noindent $\mathbf{C}_{LP})$ are block diagonal, in which $\mathbf{A}_{LP}$, $\mathbf{B}_{LP}$ and $\mathbf{C}_{LP}$ are defined as follows \cite{MiddlehurstJack/Harrison1993}:

\begin{equation}
\label{eq8}
	\mathbf{A}_{LP}=\begin{bmatrix}	0	&	1 \\	-\omega^{2}_{LP}	&	-\sqrt{2}\omega_{LP}	\end{bmatrix},\mathbf{B}_{LP}=	\begin{bmatrix}	0\\	1	\end{bmatrix},\mathbf{C}_{LP}=	\begin{bmatrix}	\omega^{2}_{LP}	&	0	\end{bmatrix}.
\end{equation}

The value of $\omega_{LP}$, which is the cut-off frequency, must be sufficiently high when compared with the maximum vibration mode frequency that may be controlled.

\subsubsection{Sensor/actuator locations and gain matrix tuning}

In addition to configuring the values of $\mathbf{\Phi}_u$ and $\mathbf{\Phi}_y$, which depend on the position of the $p$ actuators and $q$ sensors, the control gain matrix must be tuned. This matrix, which is denoted by \textbf{K}, is defined in a general form as follows:
\begin{eqnarray}\label{K_Matrix}
\textbf{K}=\left[\begin{array}{cccc}
K_{11} & K_{12} & \cdots & K_{1q} \\
K_{21} & K_{22} & \cdots & K_{2q} \\
\vdots & \vdots & \ddots & \vdots       \\
K_{p1} & K_{p2} & \cdots & K_{pq} \\
\end{array}\right],
\end{eqnarray}
in which $K_{ij}$ is the control gain applied at control input $i$ due to control output $j$.

Therefore, the state equation of the closed-loop system is obtained from Figure \ref{fig1} and Equations (\ref{eq1})-(\ref{K_Matrix}), and results in
\begin{eqnarray} \label{A_FEED_S}
\mathbf{\dot{x}}_{CL}=	&	\mathbf{A}_{CL}\mathbf{x}_{CL}+\mathbf{B}_{_{CL}}\mathbf{w}_s,\\ \nonumber
		\mathbf{y}_s=		&	\mathbf{C}_{CL}\mathbf{x}_{CL} + \mathbf{D}_{CL}\mathbf{w}_s,
\end{eqnarray}
where the state vector is $\mathbf{x}_{CL}=\left[\textbf{x}_{s},\textbf{x}_{I},\textbf{x}_{LP},\textbf{x}_{A}\right]$ and the state matrices are:

\begin{eqnarray} \label{A_FEED}
\mathbf{A}_{CL}=\left[\begin{array}{cccc}
\textbf{A}_s																	&\textbf{0}												&\textbf{0}																&-\textbf{B}_{S_1}\textbf{C}_{A_T} \\
\textbf{B}_{I_T}\textbf{C}_s									&\textbf{A}_{I_T}									&\textbf{0}																&\textbf{0} \\
\textbf{B}_{LP_T}\textbf{D}_{I_T}\textbf{C}_s &\textbf{B}_{LP_T}\textbf{C}_{I_T}&\textbf{A}_{LP_T}												&\textbf{0} \\
\textbf{0}																		&\textbf{0}												&\textbf{B}_{A_T}\textbf{K}\textbf{C}_{s}	&\textbf{A}_{A_T}\\ \end{array}\right], \\ \nonumber
\mathbf{B}_{CL}=\left[\begin{array}{c}
\textbf{B}_{s_1} \\
\textbf{0} \\
\textbf{0} \\
\textbf{0} \\
\end{array}\right], \mathbf{C}_{CL}=\left[\textbf{C}_{s}\textbf{A}_s,\textbf{0},\textbf{0},-\textbf{B}_{s_1}\textbf{C}_{A_T}\right] \ \text{and} \ \mathbf{D}_{CL} = \textbf{B}_{s_1}.
\end{eqnarray}

\begin{figure}[!ht]
\begin{center}
\includegraphics[width=7.5cm]{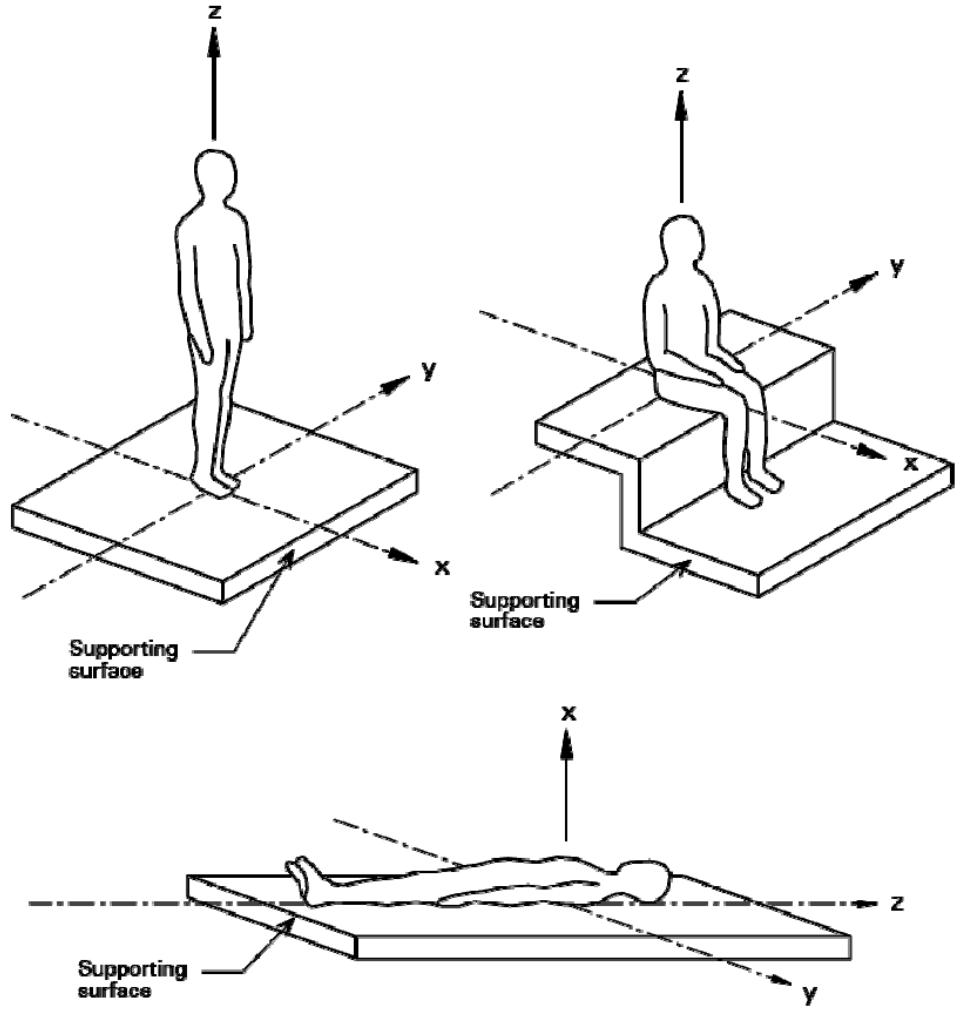}    
\caption{Directions for vibration according to ISO 2631 \cite{ISO2631}.}  
\label{Dire_ISO}                                 
\end{center}                                 
\end{figure}

\begin{figure}[!ht]
\begin{center}
\includegraphics[width=7.5cm]{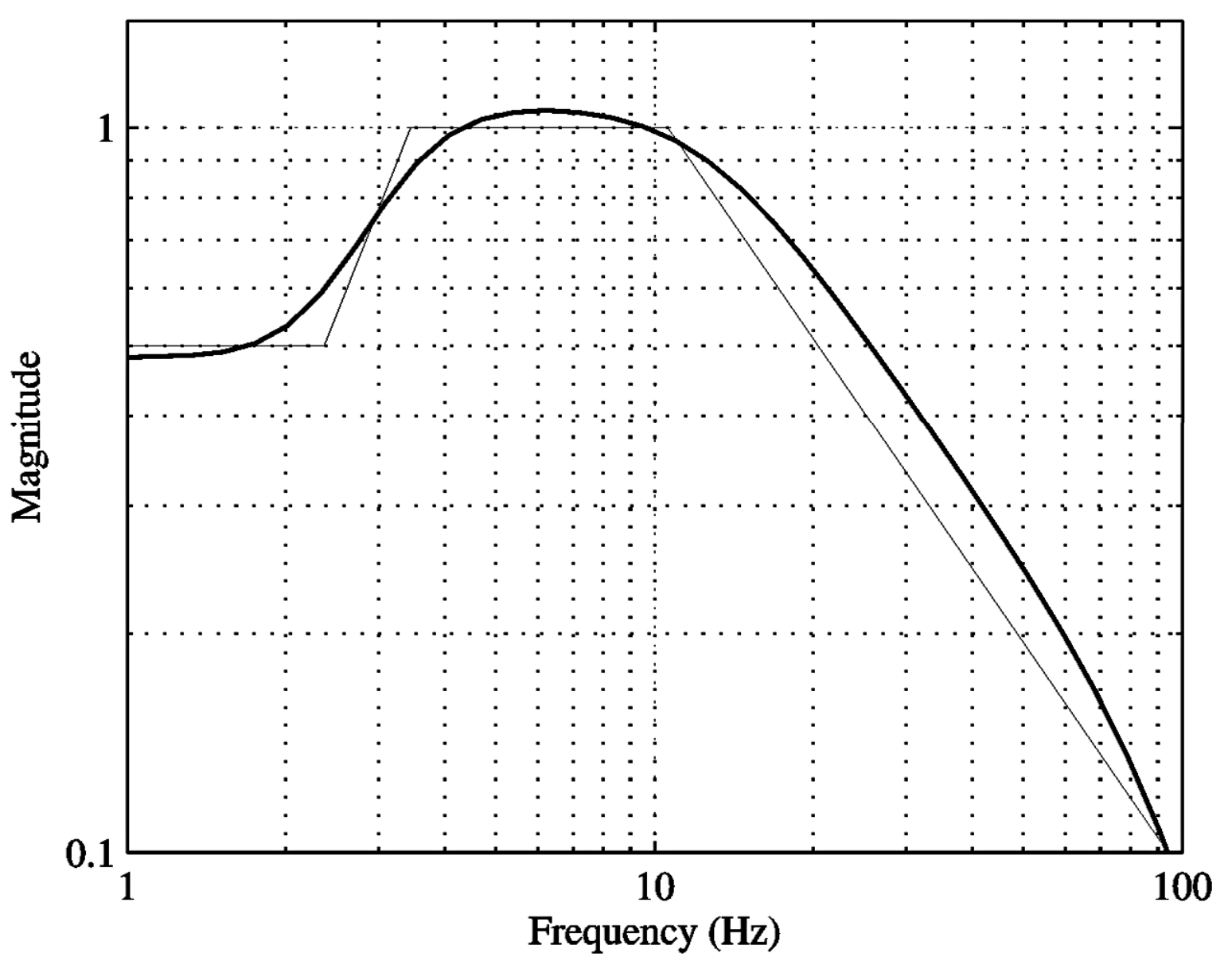}    
\caption{Frequency weighting function $W_k$ (thicker curve) and its asymptotic definition (thinner curve) \cite{ISO2631}.}  
\label{Freq_W}                                 
\end{center}                                 
\end{figure}

The sensor/actuator locations and gain matrix tuning consider the human vibration perception. This perception depends on the direction of incidence to the human body, the frequency content of the vibration (for given amplitude) and the duration of sustained vibration, among other factors. The frequency sensitivity variation for a body position can be taken into account by attenuating or enhancing the system response for frequencies where perception is less or more sensitive, respectively. The degree to which the response is attenuated or enhanced is referred to as frequency weighting. Thus, frequency weighting functions are applied in order to account for the different acceptability of vibrations for different directions and body positions. ISO 2631 \cite{ISO2631} provides details for frequency and direction weighting functions that can be applied, which are all based on the basicentric coordinate system shown in Figure \ref{Dire_ISO}. These have been included in current floor design guidelines such as the SCI guidance \cite{Smith07}. According to ISO 2631, for z-axis vibration and standing and seating, the frequency weighting function ($W_k$) is a filter with the frequency response shown in Figure \ref{Freq_W}.

Human comfort under vibration is also related to the duration of sustained vibration \cite{Lenzen1966}. Thus, persistent vibrations should be penalised in the control design, giving more importance to transient vibration of long-duration than those of short-duration. This is taken into account by multiplying the system response by an exponential time weighting (i.e., $e^{\alpha t}$), where $\alpha>0$ adds a constraint in the relative stability of the controlled system. Note that persistent states are penalised more heavily as $\alpha$ is increased. The value of $\alpha$ assumed in this work is the relative stability for the closed-loop system when $\textbf{K}=\boldsymbol{0}$, which is defined by $\min_{i}\left(\xi_i\omega_i\right)=\xi_1\omega_1=0.3547$. Note that even the worst-designed controller would be able to increase the relative stability, in other words, the real part of any eigenvalue of $\mathbf{A}_{CL}$ must be less than -0.3547.

Thus, the human vibration perception is considered in the controller design by weighting the state vector of the structure $\textbf{x}_s$ (see Equation (\ref{A_FEED})) as follows:
\begin{equation}
{x}_{s_{W_{l}}}=\left(e^{\alpha t}{x}_{s_l}(t)\right)*g_{FW}(t), \ l\in[1,\cdots,2n],
\label{W_xs}
\end{equation}
where (*) denotes the convolution process and $g_{FW}(t)$ is the impulse response function of a system with the frequency response function shown in Figure \ref{Freq_W}. Note that the weighted vector ${\textbf{x}}_{s_{W}}$ is only used to calculate the PI used to derive the optimal sensor/actuator locations and the gain matrix. In other words, the weighting functions are not included into the closed-loop system of Figure \ref{fig1} and into the Equations  (\ref{A_FEED_S}) and (\ref{A_FEED}).

\subsection{Optimization problem}\label{Optimization_Problem}

The optimization problem consists of minimizing the following functional:
\begin{equation} \label{min1}
g(\mathbf{z})=\max \left\{  \frac{1}{2}\int_{0}^{t_{f}}\mathbf{x}^{T}_{sw}(\mathbf{z})\mathbf{Q}\mathbf{x}_{sw}(\mathbf{z})\mathrm{d}t  \right\},
\end{equation}
by finding the optimal parameters of $\mathbf{z}=\left[\textbf{K},\boldsymbol{\Lambda}\right]$, where the matrix $\textbf{Q}$ is a $2n\times2n$ positive definite matrix, which is taken as \cite{Hanagan2000}
\begin{equation}
\textbf{Q}=\left[\begin{array}{cccccc}
\omega_1^2\phi_{1,\max}^2 & \cdots & 0												 & 0							 & \cdots & 0               \\
\vdots									  & \ddots & \vdots										 & \vdots				   & \ddots & \vdots          \\
0												  & \cdots & \omega_n^2\phi_{n,\max}^2 & 0							 & \cdots & 0               \\
0												  & \cdots & 0												 & \phi_{1,\max}^2 & \cdots & 0               \\
\vdots									  & \ddots & \vdots										 & \vdots				   & \ddots & \vdots          \\
0												  & \cdots & 0												 & 0							 & \cdots & \phi_{n,\max}^2 \\
\end{array}\right],
\label{Matrix_Q}
\end{equation}
in which $\phi_{k,\max}$ is the maximum value of the $k^{th}$ eigenvector $\boldsymbol{\phi}_{k}$. Note that the displacement states are weighted by the natural frequencies, thus making the displacement states comparable to velocity states. The variable $\boldsymbol{\Lambda}$ contains the locations of a set of $p$ actuators and $q$ sensors (i.e., $\boldsymbol{\Lambda}$ is a vector of $p=q$ components, whose values indicate the node number where each pair sensor/actuator is placed). Finally, the value of $t_f$ is the simulation time to obtain the PI, which must be large enough to achieve the steady state of $g(\mathbf{z})$ (i.e., the weighted state vector ${\textbf{x}}_{s_{W}}\cong0$).

Note that the problem of finding $\mathbf{z}=\left[\textbf{K},\boldsymbol{\Lambda}\right]$, which minimizes $g(\mathbf{z})$ is a hard optimization problem, in which traditional optimization methods (gradient-based approaches or similar) cannot be applied due to the problem's characteristics (for example, $g(\mathbf{z})$ cannot be obtained analytically).

\section{CRO substrates definition and main varieties}
This CRO-SL has the same characteristics shown in the Chapter \ref{cap:scheduling}. Thus, the substrates carried out in the experiments by the algorithm are the followings:
\begin{enumerate}
\item HS: Mutation from the Harmony Search algorithm with two $\delta$ values for both type of optimization variable (control gain matrix and positioning of active controllers) : $\delta=[10 20]$.
\item DE: Mutation from Differential Evolution algorithm with $F$ value linearly decreasing during the run,
      from $2$ to $0.5$.
\item 2Px: Classical 2-points crossover.
\item GM: Gaussian Mutation, with a $\delta$ value linearly decreasing during the run,
      from $0.2 \cdot (A-B)$ to $0.02 \cdot (A-B)$, where $[B,A]$ is the domain search.
\item MPx: Multi-points crossover ($M=3$).
\end{enumerate}

\section{Computational evaluation and comparisons}\label{Experimentos}

The examples carried out to evaluate the proposed CRO-SL in this context consist of two approaches of AVC, a SISO and a MIMO cases, where MIMO is configured with p=q=2 (each pair is placed at same location).

The CRO-SL parameters used in the experiments are shown in Table \ref{tab:parameters}.

\begin{table}[!ht]
\centering
\caption{Parameters values used in the CRO-SL.}
\label{tab:parameters}
\begin{tabular}{llr}
\hline
Parameter & Description & value\\
\hline
Reef & Reef size & 120\\
$F_b$ & Frequency of broadcast spawning & 97\%\\
Substrates & HS, DE, 2Px, MPx, GM & 5\\
$\mathcal{N}_{att}$ & Number of tries for larvae settlement & 3\\
$F_d$ & Fraction of corals for depredation & 15\%\\
$P_d$ & Probability of depredation & 10\%\\
$n_T$ & Maximum number of iterations & 150\\
\hline
\end{tabular}
\end{table}

Note that the linear state-space system defined in Equations (\ref{A_FEED_S}) and (\ref{A_FEED}) does not include the nonlinear actuator limitations in the stroke and voltage. It is the main reason why the computational strategy defined in \cite{Hanagan2000} cannot be applied when practical considerations are included. Therefore, since the value of $g(\mathbf{z})$ (see Equation (\ref{min1})) cannot be obtained analytically, the general control scheme of Figure \ref{fig1} is simulated in SIMULINK$^\copyright$. In order to compute this PI, a system disturbance must be defined. Note that the design of optimal controllers for unknown disturbances is not trivial, since prescribed disturbances are needed within the design process. The solution adopted in this work, similar to that used in \cite{Hanagan2000}, is to approximate the influence of zero initial conditions and a spatially distributed, but temporally impulsive, disturbance force by an appropriate initial condition and zero disturbance force. This is achieved by assigning a non-zero initial condition to the velocity states of the structure. Thus, the system disturbance is then defined as $\mathbf{x}_{s}(0)=[0,\cdots,0,\dot{x}_{s_1}(0),\cdots,\dot{x}_{s_n}(0)]$, where each value of $\dot{x}_{s_k}(0)$ is obtained as follows:

\begin{equation}
\label{eq15}
	\dot{x}_{s_k}(0)=F_{0}\phi_{k,max},
\end{equation}
\noindent where $F_{0}$ represents the impulse load value applied to a particular vibration mode. Note that the impulsive force is applied to the point of maximum amplitude of each vibration mode, creating thus an extreme scenario for the initial disturbance. It is expected that the control system will perform successfully under other loading conditions. The value of $F_{0}$ used in this work is 100 N, which excites the structure to achieve high acceleration level at some structures nodes that can be reduced by one or two commercial shakers (showed in Figure \ref{Actuator}).

The design parameters, which are not optimized by CRO-SL, are: i) $\omega_{HP}=12.6$ rad/s (2 Hz), ii) $\omega_{LP}=125.6$ rad/s (20 Hz), iii) stroke and voltage saturation of the actuator APS Dynamics Model 400 Shaker, which are 7.9 mm and 2 V, respectively, and iv) $\alpha=0.3547$ (see Equation (\ref{W_xs})).

\subsection{Results}

Table \ref{Comparacion_Resultados} shows the results obtained by the proposed CRO-SL, compared to different alternative algorithms. Specifically, all the algorithms which form the substrate layers in the CRO-SL approach have been tried on their own: the CRO with a single substrate has been run, with the same number of function evaluations than in the case of five substrates. This will show how the competitive co-evolution process promoted by the CRO-SL is able to obtain accurate solutions for the AVC design and location problem. In addition, a comparison with a high-performance recently proposed meta-heuristics for structures optimization, the Enhanced Colliding Bodies Optimization (ECBO) \cite{Kaveh14c} is included. This approach is an improved version of the CBO \cite{Kaveh14}, which includes memory and a specific mechanism to scape from local optima. The computer code for the ECBO has been released by the authors \cite{Kaveh14d}, so we have used that implementation with small adaptations to the problem at hand.

In Table \ref{Comparacion_Resultados} it can be seen how the CRO-SL (five substrates) obtain the best performance, both in the SISO and MIMO cases, with two actuators/sensors ($p=q=2$). In the SISO case, the differences among different methods are small, since it is the simplest case. In fact, the CRO-SL and CRO with DE substrate and the ECBO algorithm obtain a similar value of the PI. In this case, the HS substrate is the next algorithm in terms of performance, whereas the Gaussian mutation and the two crossover operators (2-points and multi-point) are the poorest in terms of the PI. In the case of the MIMO, the differences are much more significant. The CRO-SL with five substrates in co-evolution clearly obtains the best performance. The ECBO algorithm also performs well in this problem, as it obtains the second best result overall. In this case, the DE, is the third best approach among the tested algorithms. The HS is also the next better substrate in this version of the problems, and again the Gaussian and crossover operators do not obtain competitive results on their own in this case.

\begin{table}[!ht]
\begin{center}
\caption{\label{Comparacion_Resultados} Comparison of the results obtained in the two case-studies taken into account (SISO and MIMO) with different algorithms, in terms of the fitness function considered (Equation (\ref{min1})).}
\begin{tabular}[t1]{ccccc}
\hline
& \multicolumn{2}{c}{SISO}&\multicolumn{2}{c}{MIMO} \\
\hline
& Min & Mean & Min & Mean \\
\hline
\hline
CRO-SL & ${\bf 1.8923\cdot 10^{-5}}$ & ${\bf 1.8923\cdot 10^{-5}}$ & ${\bf 1.282\cdot 10^{-5}}$ & ${\bf 1.2835\cdot 10^{-5}}$ \\
\hline
HS & $2.0326\cdot 10^{-5}$ & $2.039\cdot 10^{-5}$ & $1.681\cdot 10^{-5}$ & $1.832\cdot 10^{-5}$\\
\hline
DE & ${\bf 1.8923\cdot 10^{-5}}$ & ${\bf 1.8923\cdot 10^{-5}}$ & $1.4199\cdot 10^{-5}$ & $1.5431\cdot 10^{-5}$\\
\hline
2Px & $3.2173\cdot 10^{-5}$ & $3.2205\cdot 10^{-5}$ & $2.3401\cdot 10^{-5}$ & $2.3491\cdot 10^{-5}$\\
\hline
GM  & $2.5513\cdot 10^{-5}$ & $2.5698\cdot 10^{-5}$ & $2.2238\cdot 10^{-5}$ & $2.2549\cdot 10^{-5}$ \\
\hline
MPx  & $3.2155\cdot 10^{-5}$ & $3.2304\cdot 10^{-5}$ & $2.3375\cdot 10^{-5}$ & $2.3577\cdot 10^{-5}$ \\
\hline
ECBO & ${\bf 1.8923\cdot 10^{-5}}$ & ${\bf 1.8923\cdot 10^{-5}}$ & $1.382\cdot 10^{-5}$ & $1.3832\cdot 10^{-5}$\\
\hline
\end{tabular}

\end{center}
\end{table}

\begin{figure}[!ht]
	\begin{center}
		\includegraphics[width=0.50\columnwidth]{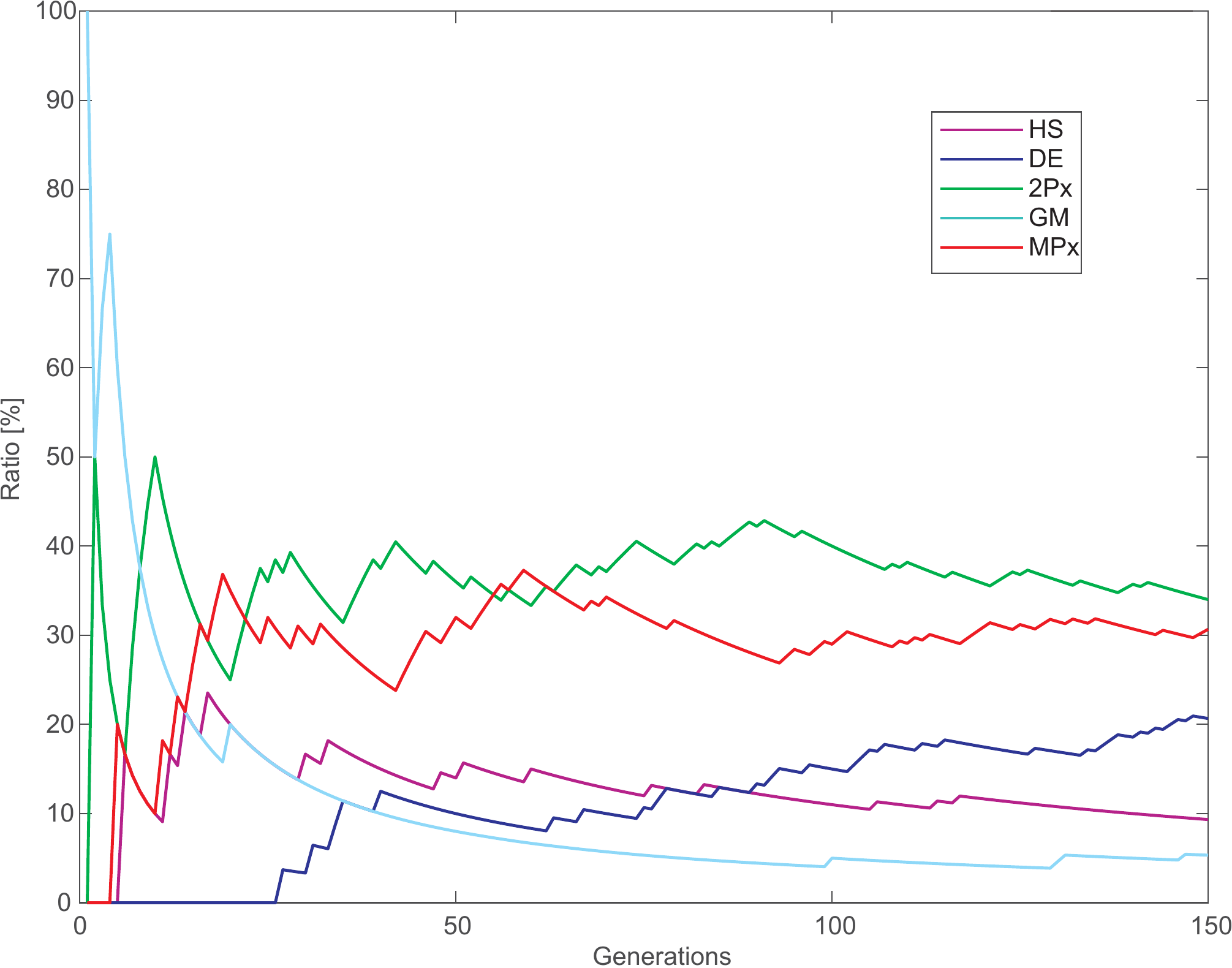}
		\caption{Ratio of times that each substrate produces the best larva in each iteration of the CRO-SL algorithm, for the AVC-MIMO design.}
		\label{ratio}
	\end{center}
\end{figure}

Figure \ref{ratio} shows the ratio of times that every substrate generates the best larva per generation in the CRO-SL approach. It is possible to see how the 2-points crossover and the MPx crossover are the two exploration operators that contribute the most to the CRO-SL search. The contribution of the DE substrate is also significant, and it grows during the search. Note that the HS and Gaussian substrates are the ones which contribute the less to the search, especially, the Gaussian substrate does not seem to be effective in this problem, and barely contributes to obtain the best solutions in each iteration of the CRO-SL.

Following, the best solutions obtained by the CRO-SL algorithm in the SISO and MIMO cases is analyzed in detail. The best result obtained by the CRO-SL for the SISO case (also reached by ECBO and DE) is the following:

\begin{eqnarray} \label{SISO_opt}
\textbf{K}=4758.6, \\ \nonumber
\boldsymbol{\Lambda}=329, \\ \nonumber
g(\mathbf{z})=1.8923\times10^{-5}.
\end{eqnarray}

\begin{figure}[!ht]
	\begin{center}
		\includegraphics[width=0.70\columnwidth]{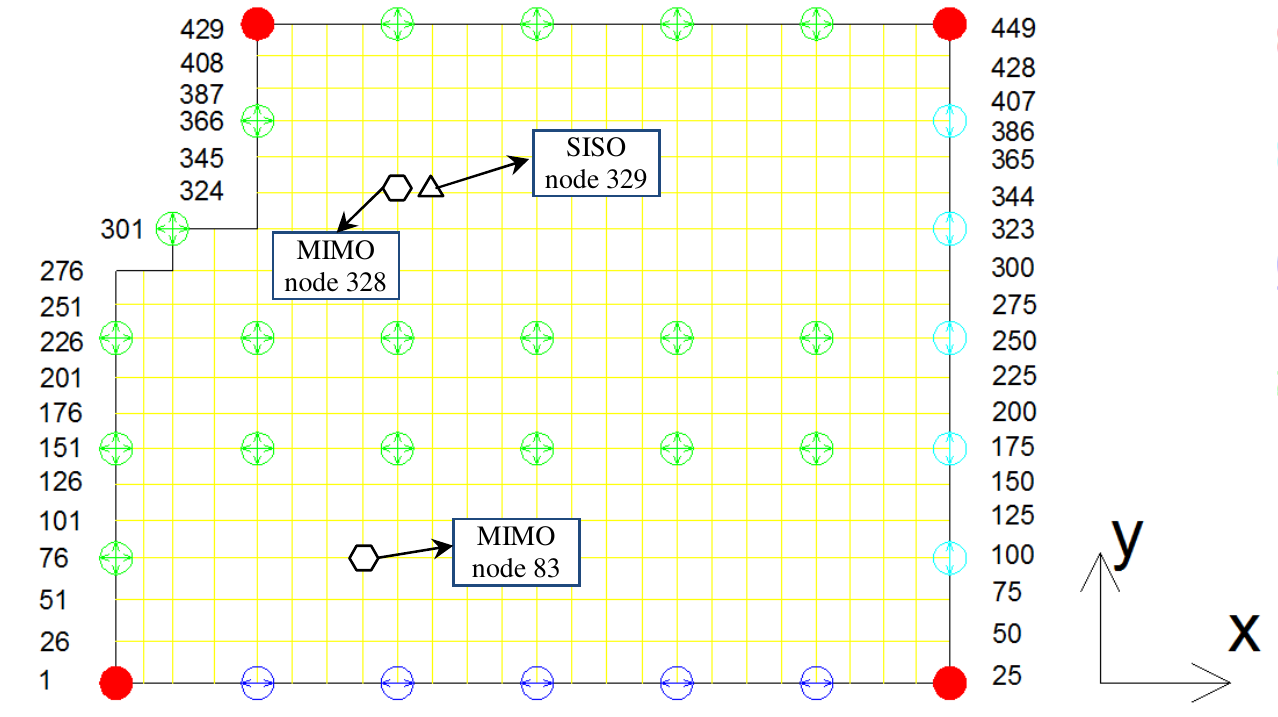}
		\caption{Optimum location of actuator/sensor pair for SISO (triangle) and MIMO (hexagons) designs. Excitation (circles) and response (squares) nodes.}
		\label{figoptloc}
	\end{center}
\end{figure}

This design is useful to reduce the vibration of the 1$^{st}$, 2$^{nd}$, 5$^{th}$, 7$^{th}$ and 9$^{th}$ modes. Note that the frequency weighting function $W_k$ of Figure \ref{Freq_W} shows that the modes with natural frequency higher than 10 Hz are less important from the human perception point of view. Then, this SISO control is mainly designed for 1$^{st}$, 2$^{nd}$ and 5$^{th}$ mode. The location of the actuator/sensor pair is shown in Figure \ref{figoptloc} (triangle). This point is a location where these three vibration modes can be measured and controlled.

The best result obtained by the CRO-SL for the MIMO case (best solution over all comparison algorithms) is the following:

\begin{eqnarray} \label{MIMO_opt}
\textbf{K}=\left[\begin{array}{cc}
1272.5 & 1367.8 \\
\cong0      & 3753.2 \\ \end{array}\right], \\ \nonumber
\boldsymbol{\Lambda}=[83 \ 328], \\ \nonumber
g(\mathbf{z})=1.282\times10^{-5}.
\end{eqnarray}

The two actuator/sensor pairs are shown in Figure \ref{figoptloc} (hexagons). The first actuator/sensor location is practically the same as SISO design (node 328). The second actuator/sensor pair is useful to reduce the vibration of the 3$^{rd}$, 4$^{th}$, 6$^{th}$, 8$^{th}$ and 10$^{th}$ mode. However, if the human perception is considered, the actuator placed at node 83 is mainly focused on 3$^{rd}$ and 4$^{th}$ mode. Therefore, this MIMO design can reduce the vibration of the first five vibration modes, all which can influence the human comfort. It should be also remarked that the optimum matrix gain \textbf{K} configures the first actuator as a two inputs - one output system, whereas the second actuator is configured as a decentralized collocated control system. Therefore, although the actuator/sensor locations may be obvious, since MIMO AVC reduces the vibration level of the first five vibration modes, the configuration of \textbf{K} together with the problem of actuator/sensor locations cannot be solved with traditional control design tuning methods. Finally, the value of PI is reduced 32.25 $\%$ with respect to the SISO case.

The performance of both controllers (SISO and MIMO) is also tested by exciting the structure with jumping and heel-drop inputs. These two perturbations are representative examples of periodic (jumping is a rhythmic human activity, such as dancing or aerobics) and impulsive (heel-droop allows measuring the damping imparted by controllers, evaluating the transient response evaluation of floor structures and checking stability properties) loads, which can be generated by humans.

The performance of both controllers tested using a heel-drop excitation allows monitoring the peak acceleration and settling time. The force of the heel-drop is modelled by a ramp with initial value of 2670 N and 50 ms of duration. The settling time is defined as the time taken for the response to fall and remain within some specified percentage of the maximum peak value of the acceleration response. For the sake of simplicity, here it is calculated as the time taken for the acceleration response to fall and remain within the range of 0.005m/s2. The same nodes are chosen (excitation nodes are 82, 88, 327, 355 and 360, monitoring nodes are 333 and 90).

\begin{figure}[!ht]
\centering
\includegraphics[width=0.6\linewidth]{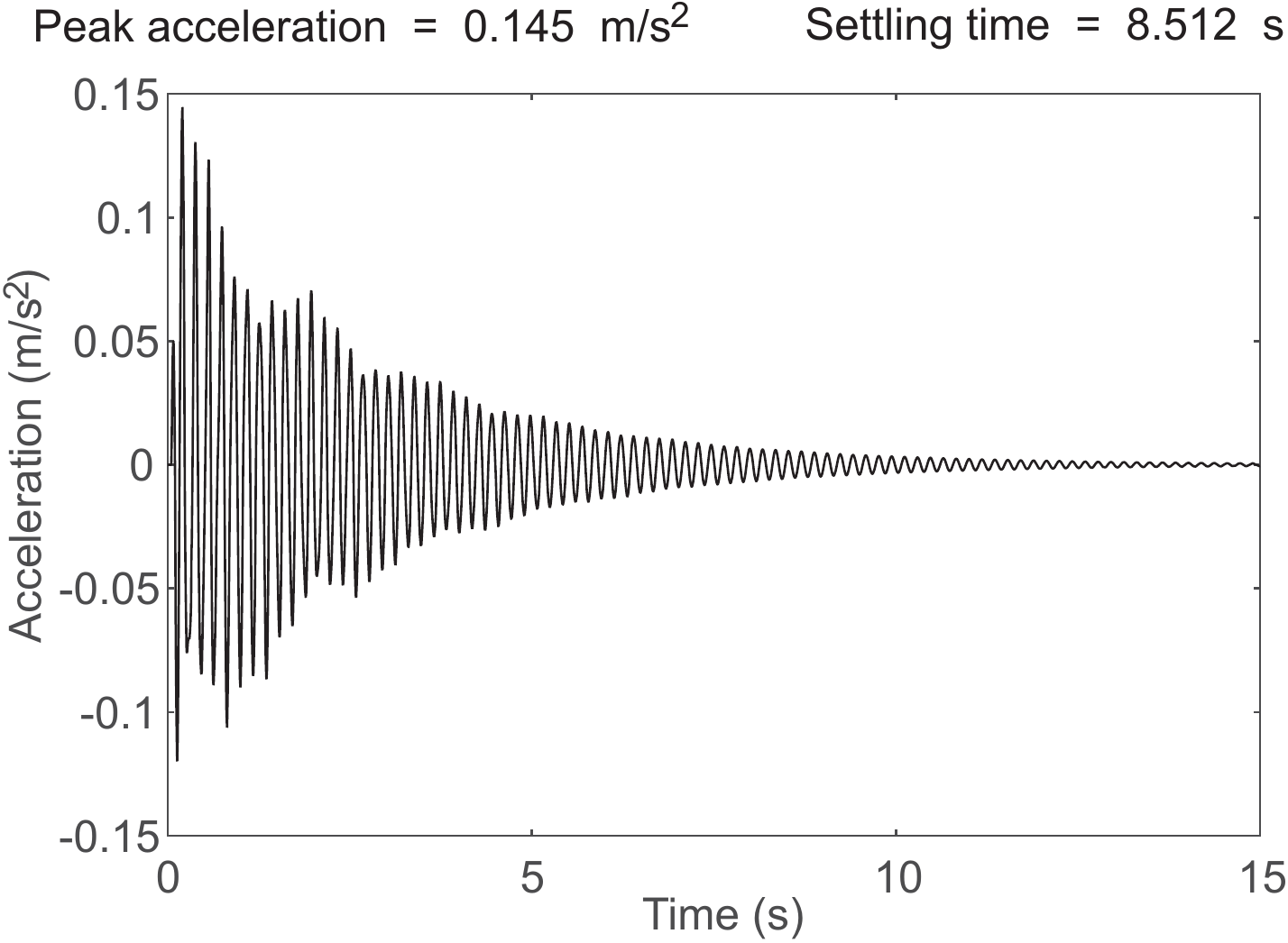}
\caption{Simulated time response without AVC of node 333 for a normalized heel-drop excitation at nodes 82, 88, 327, 355 and 360.}
\label{2NoAVC}
\end{figure}

\begin{figure}[!ht]
\centering
\includegraphics[width=0.6\linewidth]{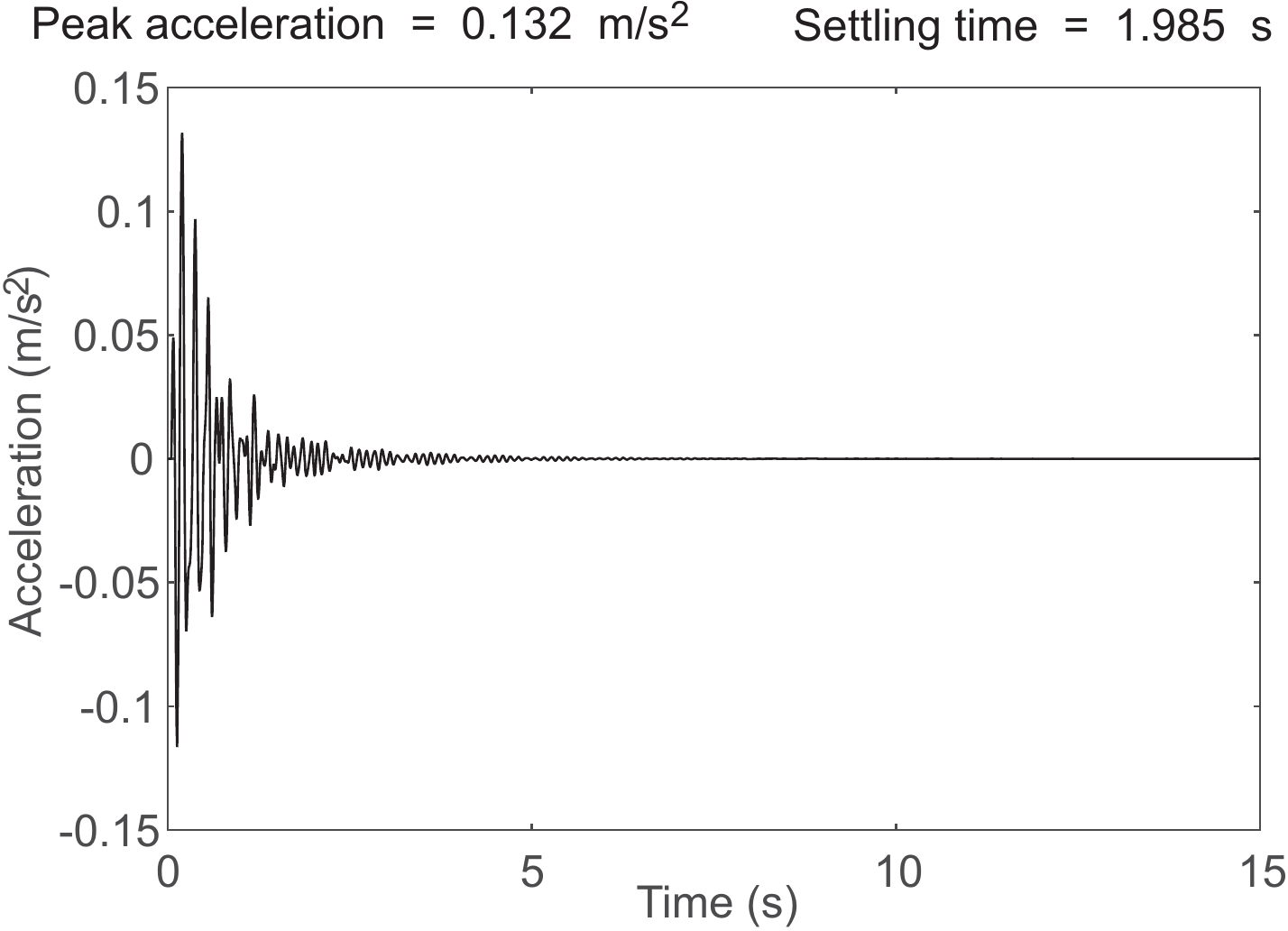}
\caption{Simulated time response with SISO AVC of node 333 for a normalized heel-drop excitation at nodes 82, 88, 327, 355 and 360.}
\label{2SISOAVC}
\end{figure}

\begin{figure}[!ht]
\centering
\includegraphics[width=0.6\linewidth]{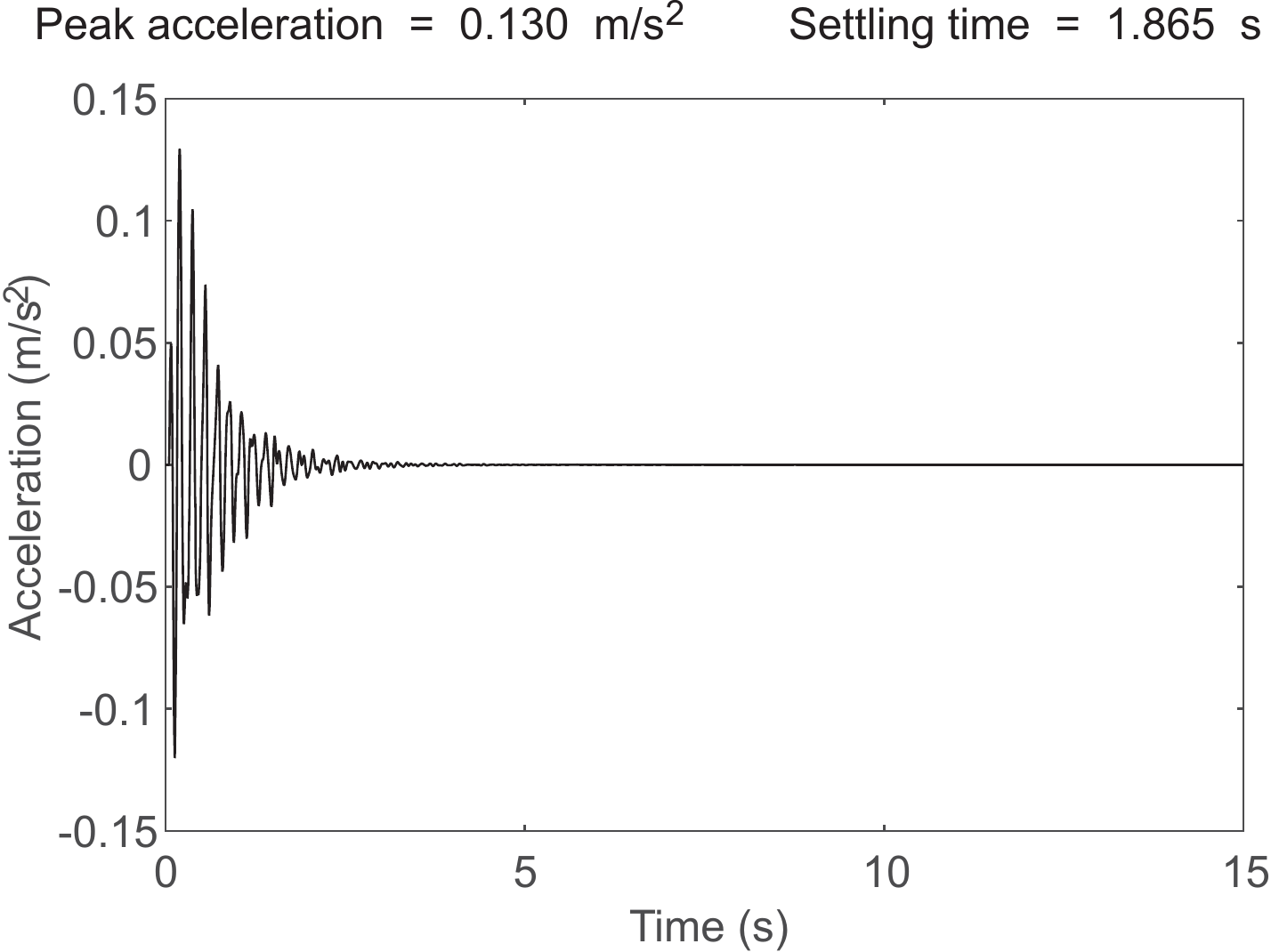}
\caption{Simulated time response with MIMO AVC of node 333 for a normalized heel-drop excitation at nodes 82, 88, 327, 355 and 360.}
\label{2MIMOAVC}
\end{figure}

Figures \ref{2NoAVC}, \ref{2SISOAVC} and \ref{2MIMOAVC} show the simulated time response of node 333 for no controlled case, the SISO AVC of Equation (\ref{SISO_opt}) and the MIMO AVC of Equation (\ref{MIMO_opt}). Note that the vibration reduction level is practically the same for SISO and MIMO AVC at node 333 since the peak acceleration is 0.132 m/s$^2$ for SISO and 0.130 m/s$^2$ for MIMO, while the settling time is 1.985 s for SISO and 1.865 s for MIMO.

\begin{figure}[!ht]
\centering
\includegraphics[width=0.6\linewidth]{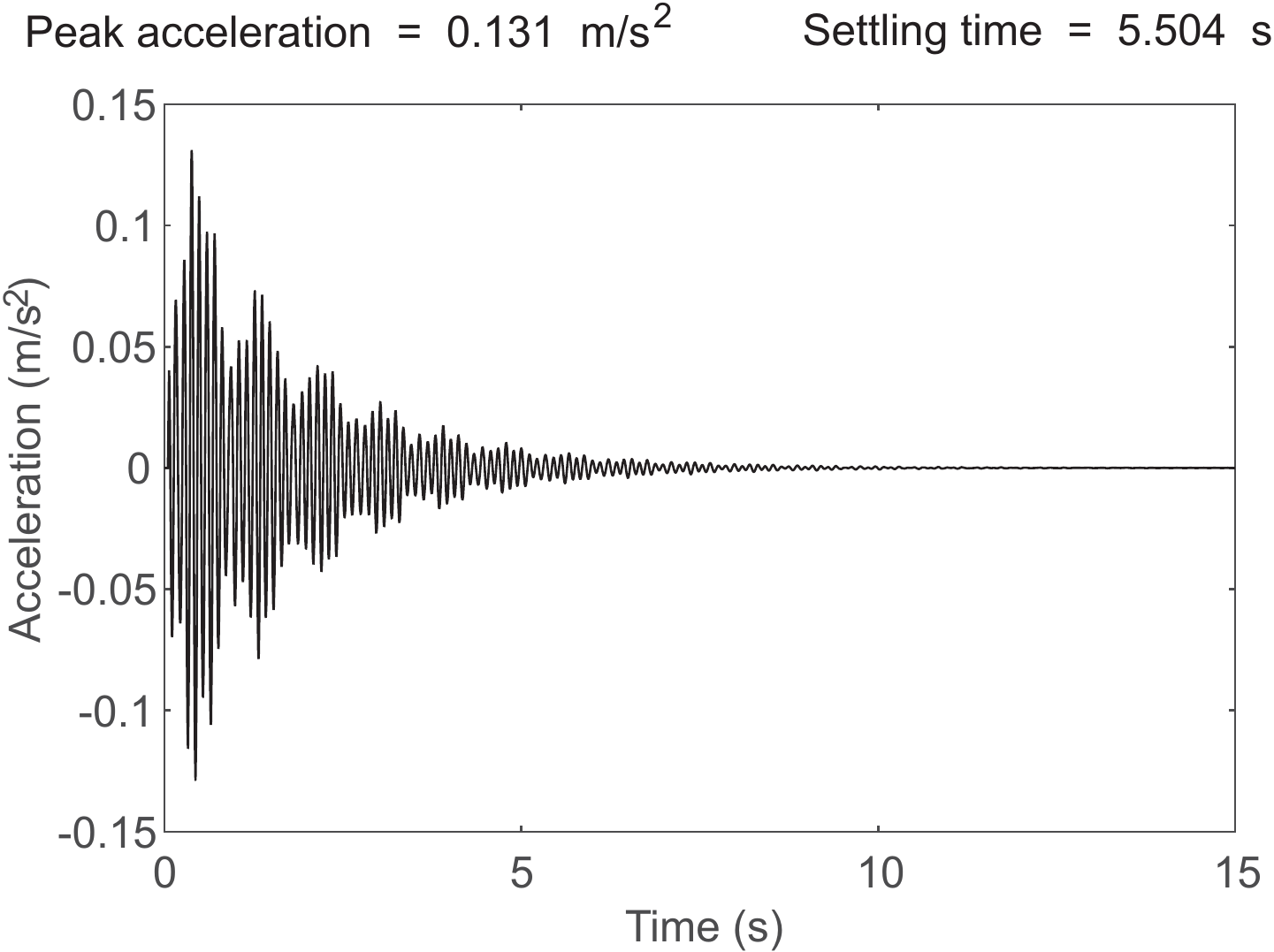}
\caption{Simulated time response without AVC of node 90 for a normalized heel-drop excitation at nodes 82, 88, 327, 355 and 360.}
\label{1NoAVC}
\end{figure}

\begin{figure}[!ht]
\centering
\includegraphics[width=0.6\linewidth]{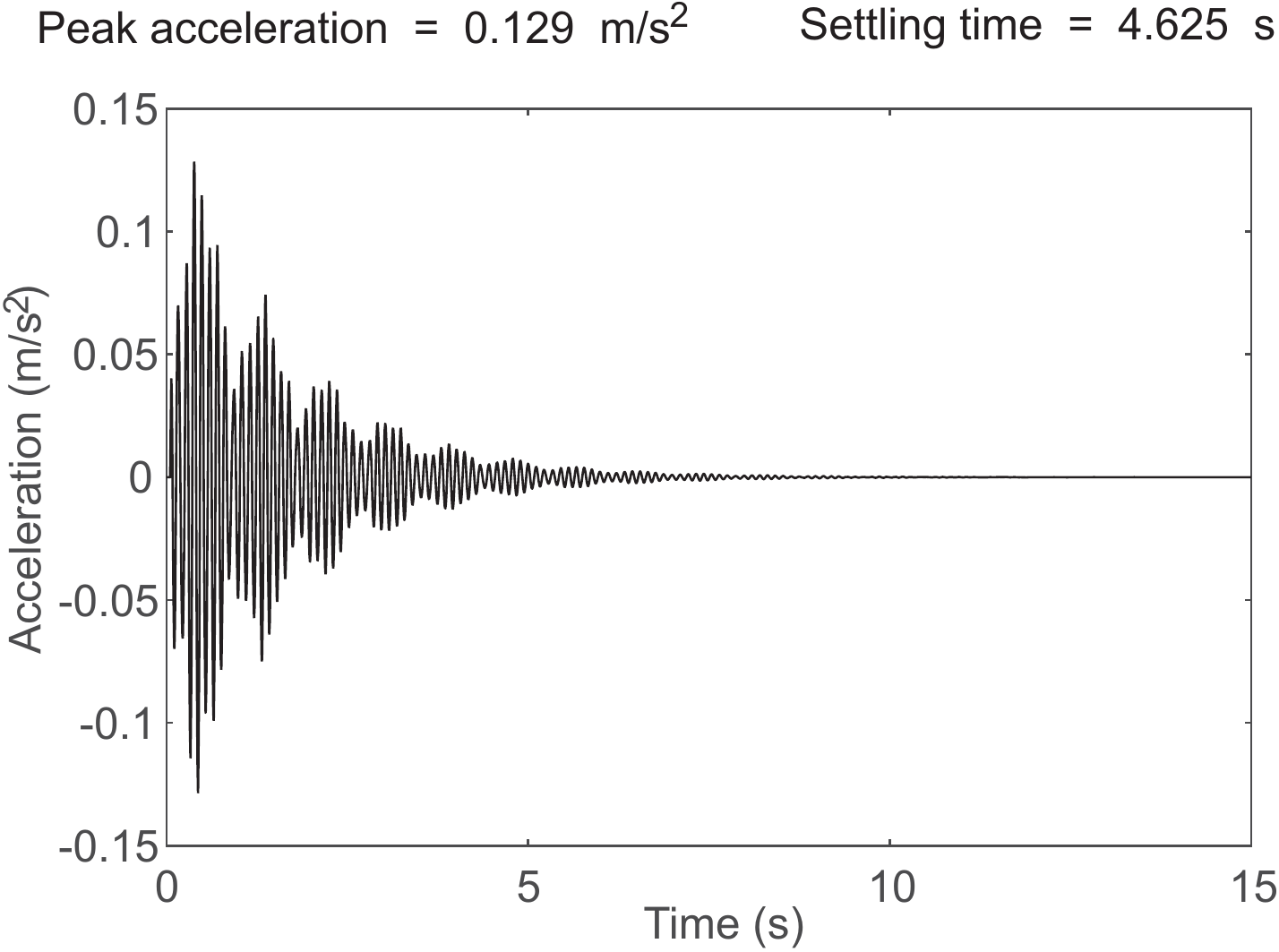}
\caption{Simulated time response with SISO AVC of node 90 for a normalized heel-drop excitation at nodes 82, 88, 327, 355 and 360.}
\label{1SISOAVC}
\end{figure}

\begin{figure}[!ht]
\centering
\includegraphics[width=0.6\linewidth]{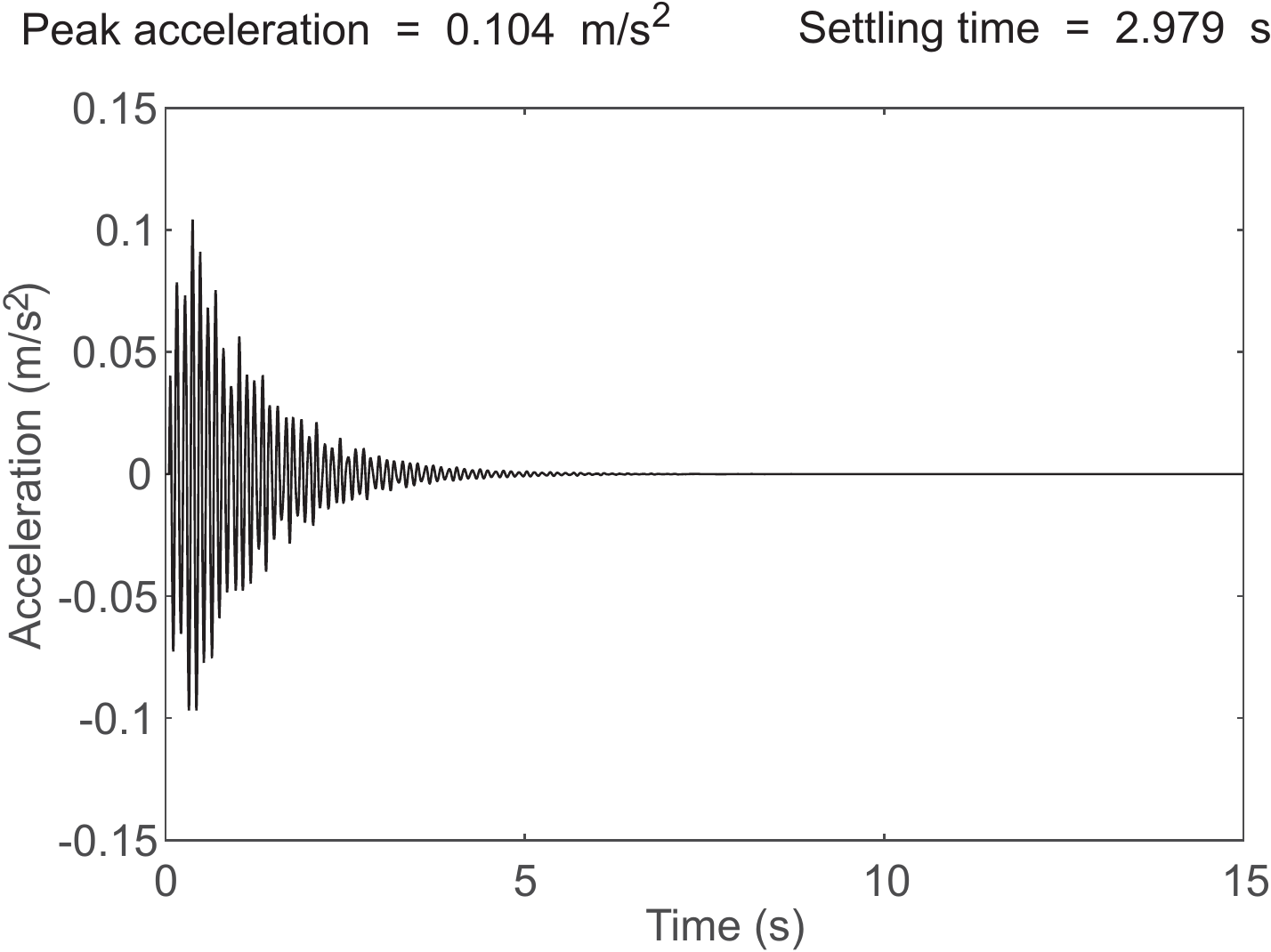}
\caption{Simulated time response with MIMO AVC of node 90 for a normalized heel-drop excitation at nodes 82, 88, 327, 355 and 360.}
\label{1MIMOAVC}
\end{figure}

Figures \ref{1NoAVC}, \ref{1SISOAVC} and \ref{1MIMOAVC} show the simulated time response of node 90 for: no controlled case, the SISO AVC of Equation (\ref{SISO_opt}) and the MIMO AVC of Equation (\ref{MIMO_opt}). Note that, unlike node 333, the vibration reduction level is significant if SISO and MIMO AVC are compared since the peak acceleration is 0.129 m/s$^2$ for SISO and 0.104 m/s$^2$ for MIMO, while the settling time is 4.625 s for SISO and 2.979 s for MIMO. Therefore, the MIMO AVC also improve substantially, like for jumping excitation example, the floor area defined by the node numbers between 1 and 175 (see Figure \ref{figoptloc}).

For the jumping excitation, a pedestrian of weight 800 N was considered and the dynamic load factors (amplitude of each harmonic) of the excitation proposed in \cite{ISO10137} were assumed (the periodic excitation is represented as a Fourier serie). Five nodes (82, 88, 327, 355 and 360) are chosen to excite with (see Figure \ref{figoptloc}) with these frequencies: i) 2.8 Hz - Node 355 - 1$^{st}$ vibration mode (5.6 Hz), ii) 2.5 Hz - Node 360 - 2$^{nd}$ vibration mode (7.5 Hz), iii) 3.0 Hz - Node 88 - 3$^{rd}$ vibration mode (9.1 Hz), iv) 3.4 Hz - Node 82 - 4$^{th}$ vibration mode (10.2 Hz) and v) 3.6 Hz - Node 327 - 5$^{th}$ vibration mode (10.7 Hz).

\begin{figure}[!ht]
\centering
\includegraphics[width=0.7\linewidth]{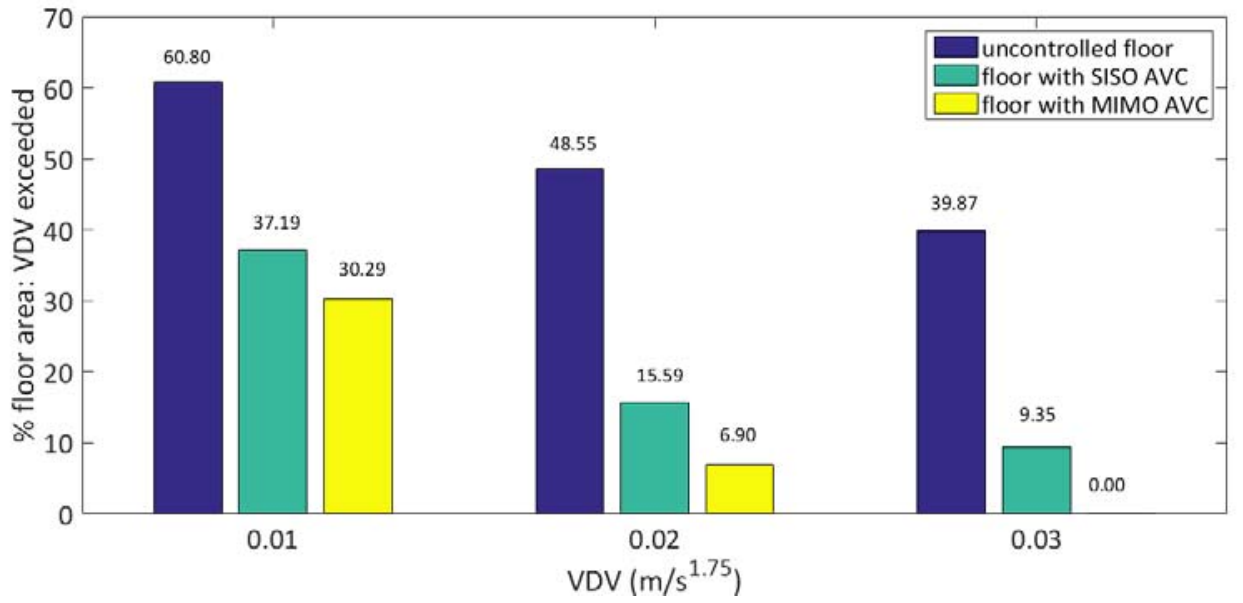}
\caption{Percentage of floor area where VDV is exceeded.}
\label{JumpAll}
\end{figure}

In order to illustrate the influence of the controllers, the vibration dose value (VDV), also known as fourth power vibration dose method, is used to provide information about the serviceability of the floor for rhythmic activities \cite{Smith07}. The VDV of all nodes on the floor during 60 seconds of jumping excitation is calculated for all nodes. Figure \ref{JumpAll} shows the percentage of floor area where VDV is exceeded. It can be seen that MIMO achieves a significant improvement respect to SISO when the VDV is higher or equal to 0.02 m/s$^{1.75}$. This improvement is due to the fact that SISO cannot cancel the vibration of 3$^{rd}$ and 4$^{th}$ vibration modes. In addition, MIMO AVC can eliminate the vibration higher than 0.03 m/s$^{1.75}$. Therefore, it can be concluded that a MIMO AVC may be necessary when a SISO AVC cannot guarantee that the vibration level does not exceed a maximum.

\section{Conclusions}\label{Conclusions}

In this research, the CRO-SL has been applied to a problem of active vibration control design for human-induced vibration mitigation in civil structures. In this case we have included state-of-the-art operators in the CRO-SL, such as two-points crossover mutations, Harmony Search, Differential Evolution or Gaussian mutations. In addition, we have included a comparison with a recently proposed meta-heuristic for structures design, the Enhanced Colliding Bodies Optimization algorithm. In terms of structural design, the main contribution of this research is to have the possibility of designing a multi-input multi-output AVC for complex floor structures with several closely frequency space vibration modes, where the number of test points and sensor/actuator pairs is not a problem to obtain a global optimum solution in a affordable computation time. In addition, this research also shows that a MIMO-AVC improves substantially the vibration reduction compared with a single-input and single output AVC for the proposed application example, which is a real complex floor structure. Future works will be focused on including new high-performance substrates in the CRO-SL, such as the ECBO compared, and also on the application of this work to experimentally reduce human-induced vibration in real complex floors by using inertial mass actuators.

\chapter{Optimal Design of a Planar Textile Antenna for RFID Systems}\label{cap:antena}

\section{Introduction and state of the art} Since its introduction in 1948 by H. Stockman \cite{Stockman48}, Radio Frequency IDentification have become one of the most numerous manufactured devices worldwide. They are used in access and money cards, product labels and many other applications \cite{Jacob17}. The fact that this technology has provided a reliable and inexpensive means of locating people has opened several important application lines. For example in health-care, in specific cases of dementia care, where it is important that patients can freely move, but at the same time it is mandatory a strict control of their location; or in critical professions such as security corps or firemen, who face risk situations where they need to be located every time.

Basically a RFID system is based on a radio-scanner unit, denoted {\em reader} and a set of remote transponders, usually called {\em tags}, which include an antenna and a microstrip transmitter with internal memory. Recent research works have allowed developing the design of antennas using textile materials in the substrate, leading to devices called {\em wearable antennas}, which are becoming very popular in many applications, including the design of RFID systems \cite{Hertleer09,Serra11}. The main advantages of the antennas manufactured with textile materials are the following: a reduced size, good bandwidth thanks to low relative permittivity ($\epsilon_r$=1.3), high flexibility, lightweight and an excellent electromagnetic response \cite{Khaleel14,Koski14}. In addition, they can be manufactured using Smart Fabric \& Interactive Textile Systems \cite{Lymberis08}, in which unobtrusive integration of electronic components increases functionality of the garment \cite{Dierck14,Lemey16}. All of these benefits have supported the implementation of this kind of antennas in wearables, in order to use them in RFID services \cite{Lee17,Locher06,Bayram10,Tak15,Liu15}.

According to RFID Handbook \cite{HRFID}, the band of 2.45GHz is mainly employed as RFID frequency, mainly because, at that frequency, it is possible to to use transponders and readers in the majority of countries in the world, without any modification. This important fact has contributed to the international spreading of RFID systems and technologies. Moreover, the European regulation \cite{ERC} allows RFID systems on this frequency with a higher transmitting power, which in turn increases the communication distance range using passive-type RFID systems \cite{Jin04,Hung06,Mahmoud10}.

One of the most used antennas in wireless communications is the inverted-F antenna (IFA) \cite{Volakis09, Elsheakh14}. An inverted-F antenna is a planar metallic line printed on a PCB (Printed Circuit Board), without ground plane. The main advantage of this kind of antennas is their easy implementation and matching \cite{Volakis09}. Moreover, the fact of etching the ground plane and having the radiator in the same side allows reducing the antenna SAR, which is very important in RFID services with textile materials. On the contrary, in terms of size, the IFA occupies quite large space since its length is close to quarter wavelength. Its miniaturization leads to a degradation of antenna bandwidth and efficiency \cite{Harrison63,Hansen75}. However, the dimensions can be reduced adequately by folding \cite{Elsheakh12}, capacitive loading \cite{Quevedo11} or as proposed in \cite{Marrocco03}, using {\em meander lines}. Meander  line technology  allows designing antennas with a small size and provides good wideband performance \cite{Khaleghi05}. In fact, in \cite{Seshagiri} a detailed analysis about the design process of this kind of RFID tag has been done. Additionally, the use of meander line antennas was introduced to reduce the resonant length without great deterioration of its performance \cite{Wong03,Warnagiris98}. Usually, the width of the meander-line and the spacing between meanders are set to a constant (unique) value. This simplifies the initial calculation and the further optimization process using classical methods. Note that when the meander lines width and spacing are not unique, the number of design variables becomes very high, and then classical optimization methods are no longer suitable. In that case, the employment of meta-heuristic techniques has shown to be very useful for antenna design \cite{Rocio10,Rocio12,Rocio16}.

In this work we propose a meander-line IFA with variable width and spacing meanders, optimized with the CRO-SL algorithm. Another important part of our proposal is that the proposed antenna has been constructed considering a wearable substrate, which makes more difficult the designing process. In addition to the wearable substrate, the proposed antenna allows a good bandwidth and radiation pattern in RFID working frequencies, which leads to a high performance in these systems. The CRO-SL \cite{CRO_Tutorial} is able to obtain excellent results in the optimization process of the proposed antenna, tuning it for its use in RFID systems. In the experimental section of the chapter, we also detail the accuracy on the antenna design (with the CRO-SL and simulation software), and its construction, and how it performs in terms of different measurement parameters.

\section{Proposed antenna design}\label{sec:Antenna}

The use of an IFA allows keeping the antenna and the ground at the same plane. Moreover, as previously mentioned, the meander shape has been used to reduce the final antenna dimensions \cite{Khaleghi05}. Thus, an IFA with a meander shape seems to have a proper requirement to be used as antenna for RFID systems. In order to have a starting point in the design of the IFA meander-line antenna, we consider an existing Printed Circuit Board (PCB) antenna prototype described in \cite{Note_DN023}. Note that this starting point device is not adjusted to 2.45 GHz, and in addition, the substrate considered to implement it is not textile material, so important changes on its design must be carried out in order to adapt it to RFID systems. Unfortunately, there are not specific equations for the design of IFA meander-lines antennas,  which makes very difficult to tune its parameters to optimal required values. This latter point leads to the use of meta-heuristics approaches, as will be further described in the next sections.

\begin{figure}[!ht]
\centerline{\includegraphics[width=0.5\columnwidth]{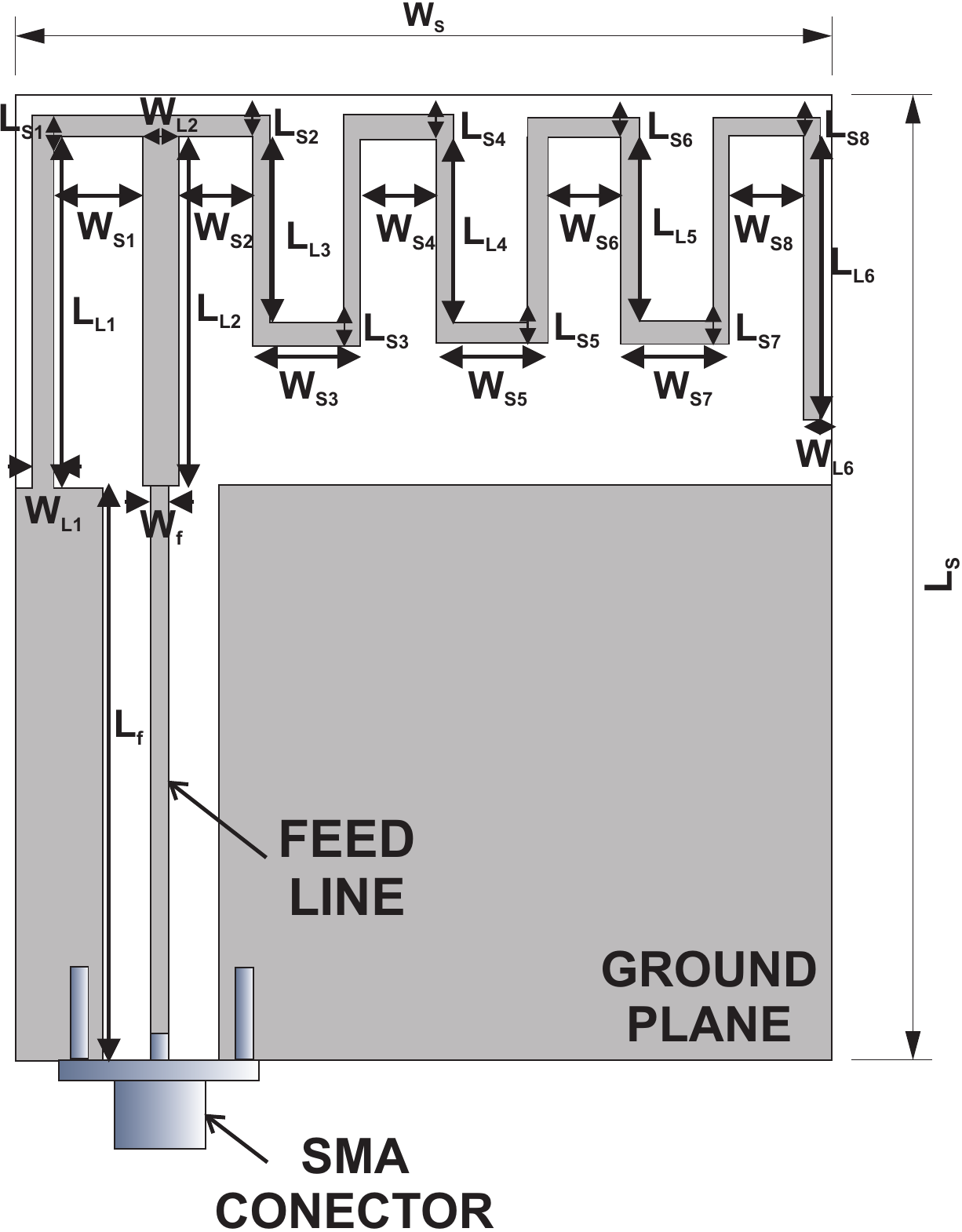}}
\caption{Geometry of the IFA meander-line antenna considered, and design variables.}
\label{fig:Layout}
\end{figure}

In order to start the design phase, the first step consists of a simulation of the IFA meander-line antenna described in \cite{Note_DN023}, but using a textile substrate instead of the original PCB. For this, textile materials such as FELT ($\epsilon_r$=1.3) as dielectric and flexible copper as conductor, which thickness are 1.1 and 0.035 mm respectively, have been considered. These material allows an easy antenna implementation for RFID systems, and therefore its embedding in clothes such as life jackets, t-shirts, etc. \cite{Lemey16,Lee17,Khaleel14,Koski14}. The layout and design variables of the original antenna are shown in the figure \ref{fig:Layout}, and the specific values of the initial antenna design (previous to optimization) are indicated in Table \ref{table:original}.

\begin{table}
\centering
\caption{Variable values for the original IFA meander-line antenna in \cite{Note_DN023}.}
\label{table:original}
\begin{tabular}{|p{40pt}|p{40pt}|p{40pt}|p{40pt}|}
\hline
Variable& Value (mm)& Variable & Value (mm)\\
\hline
$L_s$& 52& $W_s$& 44.5\\
\hline
$L_f$& 31& $W_f$& 1\\
\hline
$L_{L1}$& 19& $W_{L1}$& 1 \\
\hline
$L_{L2}$&19& $W_{L2}$&2 \\
\hline
$L_{L3}$&10.25& $W_{L3}$&1\\
\hline
$L_{L4}$&10.25& $W_{L4}$&1\\
\hline
$L_{L5}$&10.25& $W_{L5}$&1 \\
\hline
$L_{L6}$&15& $W_{L6}$&1 \\
\hline
$L_{S1}$&1& $W_{S1}$&5\\
\hline
$L_{S2}$&1& $W_{S2}$&4\\
\hline
$L_{S3}$&1& $W_{S3}$&6\\
\hline
$L_{S4}$&1& $W_{S4}$&4\\
\hline
$L_{S5}$&1& $W_{S5}$&6 \\
\hline
$L_{S6}$&1& $W_{S6}$&4 \\
\hline
$L_{S7}$&1& $W_{S7}$&6 \\
\hline
$L_{S8}$&1& $W_{S8}$&4 \\
\hline
\end{tabular}
\end{table}

\begin{figure}[!ht]
\centerline{\includegraphics[width=0.6\columnwidth]{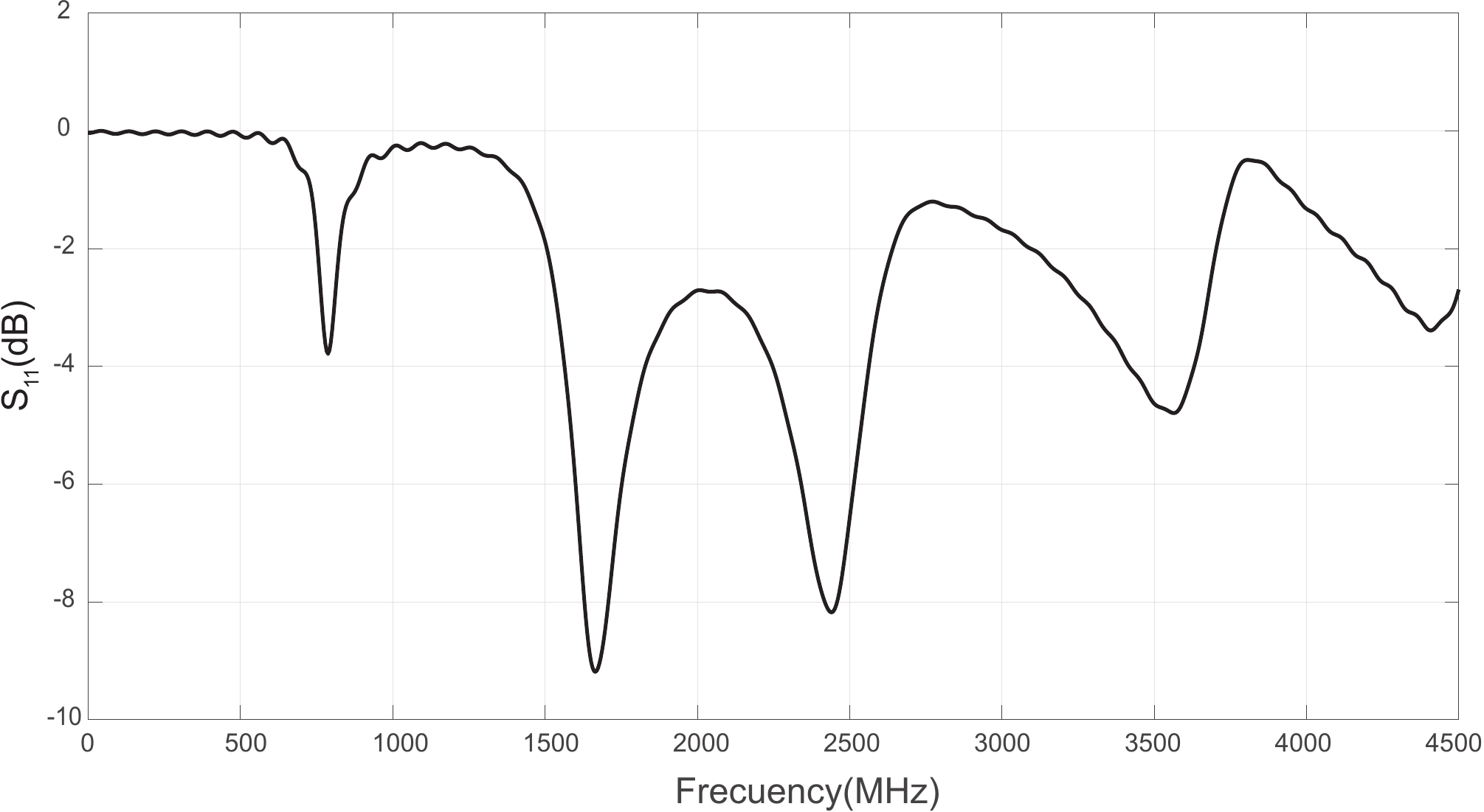}}
\caption{Reflection coefficient of the original antenna design over textile substrate (previous to optimization).}
\label{fig:S11_or}
\end{figure}

In this design phase, all the antennas simulations have been carried out with CST Microwave Studio \cite{CST}. Figure \ref{fig:S11_or} shows the reflection coefficient ($S_{11}$) obtained during the simulation process of the original IFA meander-line antenna (previous to optimization). According to these results, it is necessary to modify the dimensions of the antenna to achieve the resonant frequency at 2.45 GHz (for working on the RFID band). Moreover, it is also important to take into account the bandwidth requirement for the RFID service. According to \cite{HRFID, ERC}, the reflection coefficient of a RFID antenna must be lower than -10dB from 2.4 to 2.483 GHz. Thus, it is necessary to optimize the considered IFA meander-line antenna to adjust its resonant frequency at 2.45 GHz and to get a bandwidth higher than 83MHz. In order to fulfil these parameters of design for the antenna, the objective function of the optimization process must be carefully selected, as shown in the next section.

\subsection{Objective function}

The objective function is a key point of any optimization problem, since the algorithms uses it consistently in order to evaluate the quality of the solutions found. In meta-heuristics and constructive algorithm, the objective function leads the search with continuous evaluations of different tentative solutions to the problem. In this case, the considered function ($f({\bf x})$) to guide the antenna optimization must take into account different design requirements of the device, such as its resonant frequency and bandwidth, as was described in the previous section. Specifically, in order to calculate $f({\bf x})$, we first take into account a discretization of the $S_{11}$ antenna parameter, which will be calculated by the CST software. In this case, a discretization in steps of 2 MHz is considered. A measurement window between 2400MHz and 2500MHz is also selected to calculate $f({\bf x})$. The mathematical formulation of the objective function is the following:

\begin{equation}\label{fitness}
f({\bf x})=0.8 \cdot \mathcal{N}^{-10db}+0.1 \cdot \mathcal{M}+0.1 \cdot \mathcal{M}^*
\end{equation}
where $\mathcal{N}^{-10db}$ stands for the number of $S_{11}$ points in the observation window under $-10dB$, $\mathcal{M}=|mean(S_{11})|$ and  $\mathcal{M}^*=|min(S_{11})|$.

Note that this function takes into account the antenna bandwidth (within the first part of the equation), and also the resonant frequency and the actual value of the reflection coefficient by means of the second and third parts of the equation.

\section{CRO substrates definition and main varieties}\label{Substrates_Ant}
General purpose substrates, such as Differential Evolution or Harmony Search-based, and other very specific substrates with crossovers adapted to the problem at hand are described in this section. A total of 5 substrate layers were finally considered in the algorithm. They are the following:
\begin{enumerate}
\item HS: Mutation from the Harmony Search algorithm with $\delta$ value set to $1.5$.
\item DE: Mutation from Differential Evolution algorithm with $F$ value linearly decreasing during the run,
      from $2$ to $0.5$.
\item 2Px: Classical 2-points crossover.
\item GM: Gaussian Mutation, with a $\delta$ value linearly decreasing during the run,
      from $0.2 \cdot (A-B)$ to $0.02 \cdot (A-B)$, where $[B,A]$ is the domain search.
\item SAbM: Mutation based on fractal geometric patterns.
\end{enumerate}

\section{Experiments and results}\label{sec:Experiments}

This section presents the experimental results obtained in the design of the proposed antenna with the CRO-SL approach. The CRO-SL algorithm's main parameters used in all the simulations are shown in Table \ref{tab:Opt_Param}. Different experiments including different number of substrates in the CRO-SL algorithm were then performed in order to show the capabilities of the algorithm. The best results have been obtained by using the CRO-SL algorithm with 4 (all the described in Section \ref{Substrates_Ant} but the HS) and 5 substrates (CRO-4SL and CRO-5SL, respectively). The final values for the modified IFA meander-line after the optimization problem in both cases are shown in Table \ref{table:optimization}.

\begin{table}[!ht]
\caption{CRO-SL optimization parameters.}
\label{tab:Opt_Param}
\centering
\begin{tabular}{ll}
\hline
Phase & Parameter\\
\hline
\hline
Initialization & Reef size = $20\times 10$\\
	& $\rho_0 = 0.9$\\
	\hline
External sexual reproduction & $F_b =0.80$ \\
		& $\mathcal{T}=5$ substrates: HS, DE, 2Px, GM, SA\\
\hline
Internal sexual reproduction & $1-Fb=0.20$ \\	
\hline
Larvae setting & $\kappa =3$ \\
\hline
Asexual reproduction & $F_a = 0.05$ \\
\hline
Depredation &	$F_d = 0.15$ \\
	& $P_d = 0.05$\\
\hline
Stop criterion &	$k_{max}=50$ iterations. \\
\hline
\end{tabular}
\end{table}

\begin{table}[ht!]
\caption{Antenna dimensions after optimization process with the CRO-4SL and CRO-5SL.}
\centering
\label{table:optimization}
\begin{tabular}{|c|c|c|c|}
\hline

Variable (mm)  & \textsc{Original}  & \textsc{CRO-4SL} & \textsc{CRO-5SL} \\
\hline
$L_s$& 52& 52& 52 \\
\hline
$W_s$&44.5&44.5&44.5 \\
\hline
$L_f$& 31& 31.01& 29.76 \\
\hline
$W_f$& 1& 1& 1 \\
\hline
$L_{L1}$& 19& 18.93& 19.67 \\
\hline
$W_{L1}$& 1& 1& 1.10 \\
\hline
$L_{L2}$& 19& 19.74& 20.66 \\
\hline
$W_{L2}$& 2& 2.97& 3.47 \\
\hline
$L_{L3}$& 10.25& 17.5& 13.06 \\
\hline
$W_{L3}$& 1& 1& 5 \\
\hline
$L_{L4}$& 10.25& 17.5& 13.06 \\
\hline
$W_{L4}$& 1& 1& 4 \\
\hline
$L_{L5}$& 10.25& 17.5& 13.06 \\
\hline
$W_{L5}$& 1& 1& 4 \\
\hline
$L_{L6}$& 15& 0& 5.96 \\
\hline
$W_{L6}$& 1& 1& 1.73 \\
\hline
$L_{S1}$& 1& 0.81& 0.87 \\
\hline
$W_{S1}$& 5& 5& 5 \\
\hline
$L_{S2}$& 1& 1.19& 1.13 \\
\hline
$W_{S2}$& 4& 5& 5 \\
\hline
$L_{S3}$& 1& 17.5& 0 \\
\hline
$W_{S3}$& 6& 6& 0 \\
\hline
$L_{S4}$& 1& 1.19& 1 \\
\hline
$W_{S4}$& 4& 6& 6 \\
\hline
$L_{S5}$& 1& 17.5& 1.13 \\
\hline
$W_{S5}$& 6& 6& 6 \\
\hline
$L_{S6}$& 1& 1.19& 1.13 \\
\hline
$W_{S6}$& 4& 6& 6 \\
\hline
$L_{S7}$& 1& 17.5& 0 \\
\hline
$W_{S7}$& 6& 6&0 \\
\hline
$L_{S8}$& 1& 0.99&1.13 \\
\hline
$W_{S8}$& 4& 5&0 \\
\hline
\end{tabular}
\end{table}

\begin{figure}[!ht]
\centering
\includegraphics[width=0.7\columnwidth]{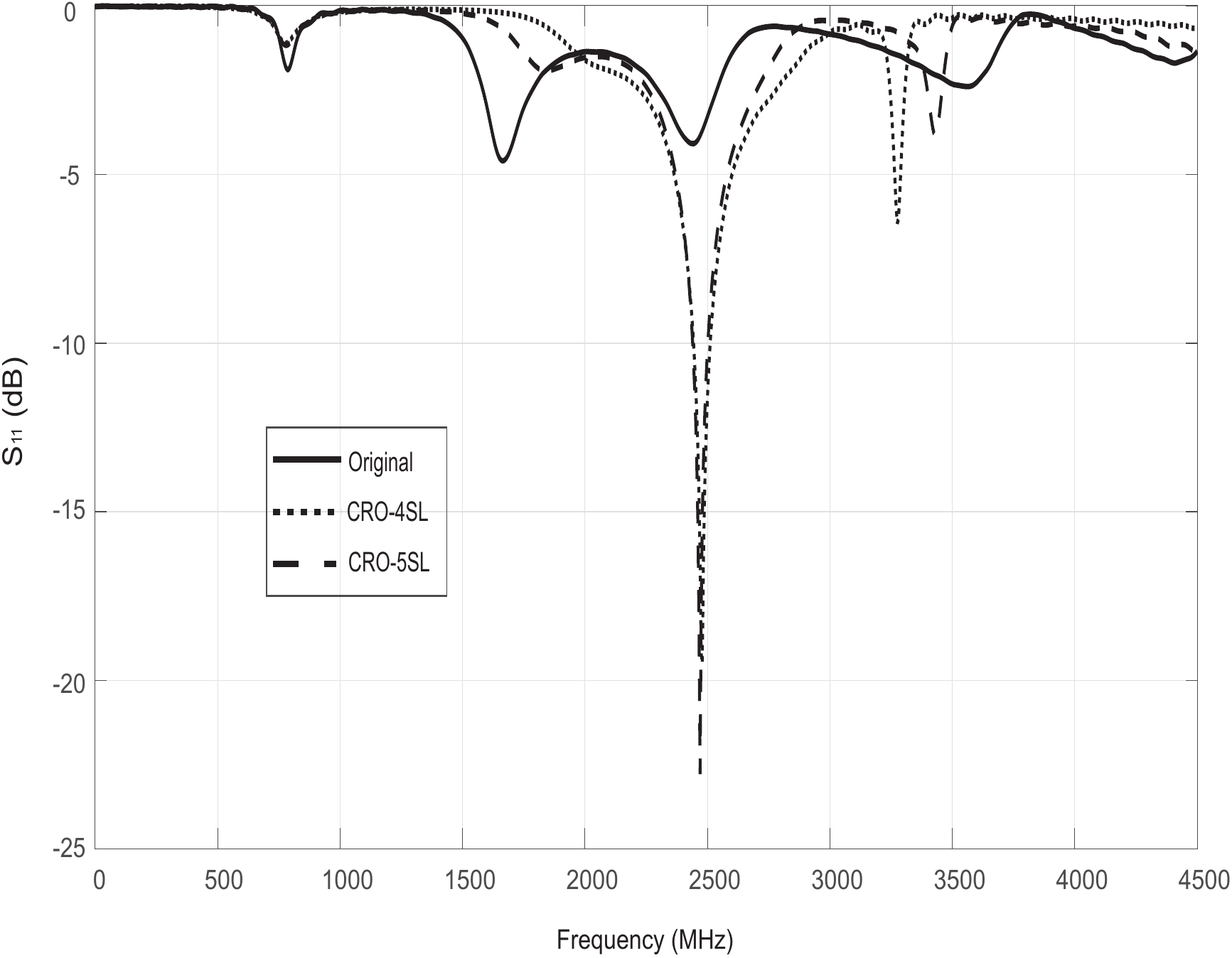}
\caption{Reflection coefficients after the optimization process with the CRO-4SL and CRO-5SL.}
\label{fig:S11_opt}
\end{figure}

The new responses for the $S_{11}$ after the antenna optimization process with the CRO-4SL and CRO-5SL are represented in Figure \ref{fig:S11_opt}. Note that the CRO-SL (both versions with 4 and 5 substrates) provide solutions with the resonant frequency of the antenna adjusted at 2.45 GHz, as defined in the required design. Moreover, the antenna bandwidth satisfies the RFID service requirements in both cases.

\begin{figure}[!ht]
\centering
\includegraphics[width=.6\columnwidth]{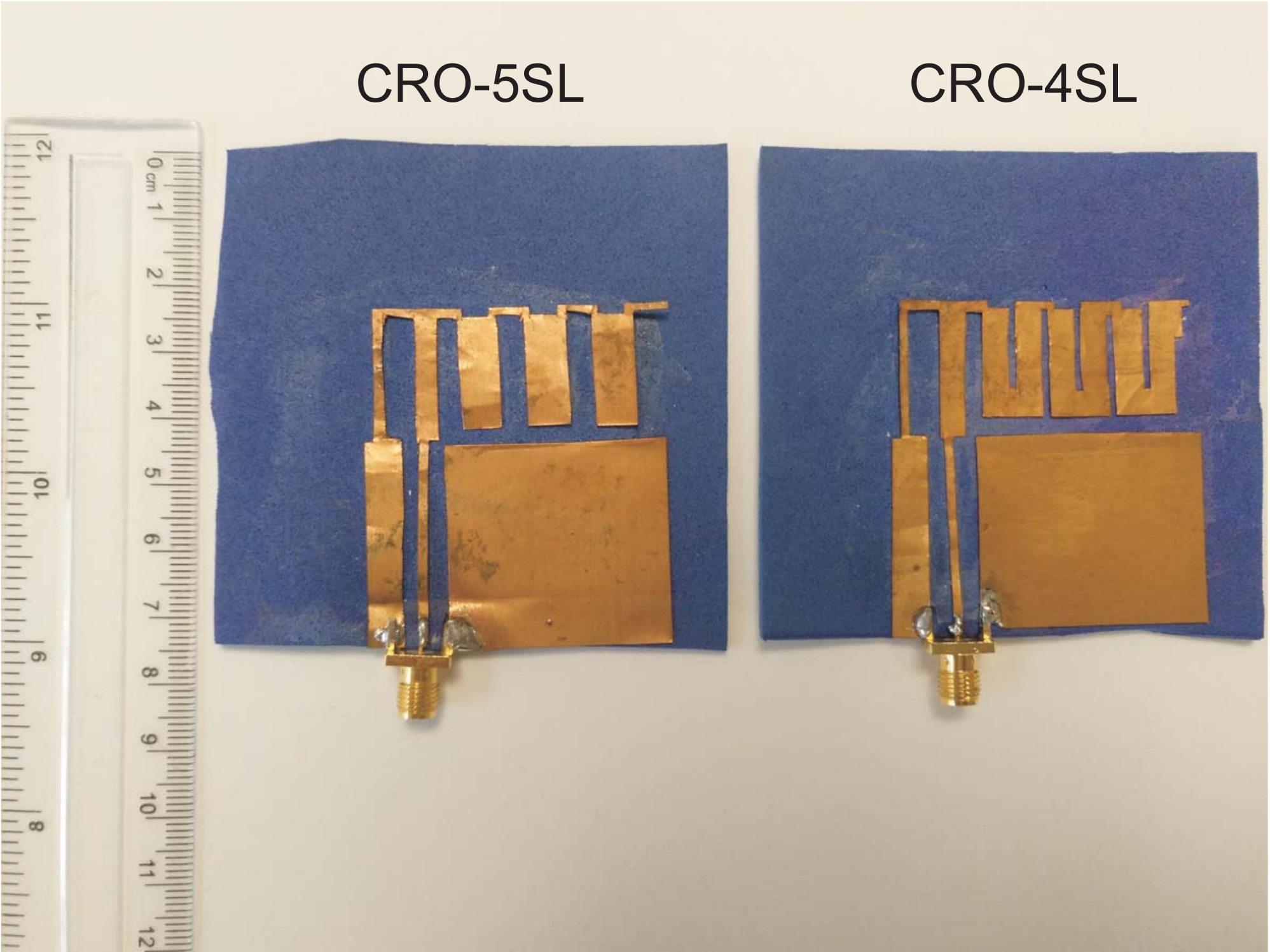}
\caption{Final prototypes of the modified IFA meander line antenna over textile substrate constructed. On the left, the one obtained with the CRO-5SL. On the right the one obtained with the CRO-4SL.}
\label{fig:Foto}
\end{figure}

In order to validate the simulation results, both prototypes obtained with the CRO-4SL and CRO-5SL algorithms have been manufactured. They are shown in Figure \ref{fig:Foto}.

\begin{figure}[!ht]
\begin{center}
\subfigure[]{\includegraphics[width=0.4\textwidth]{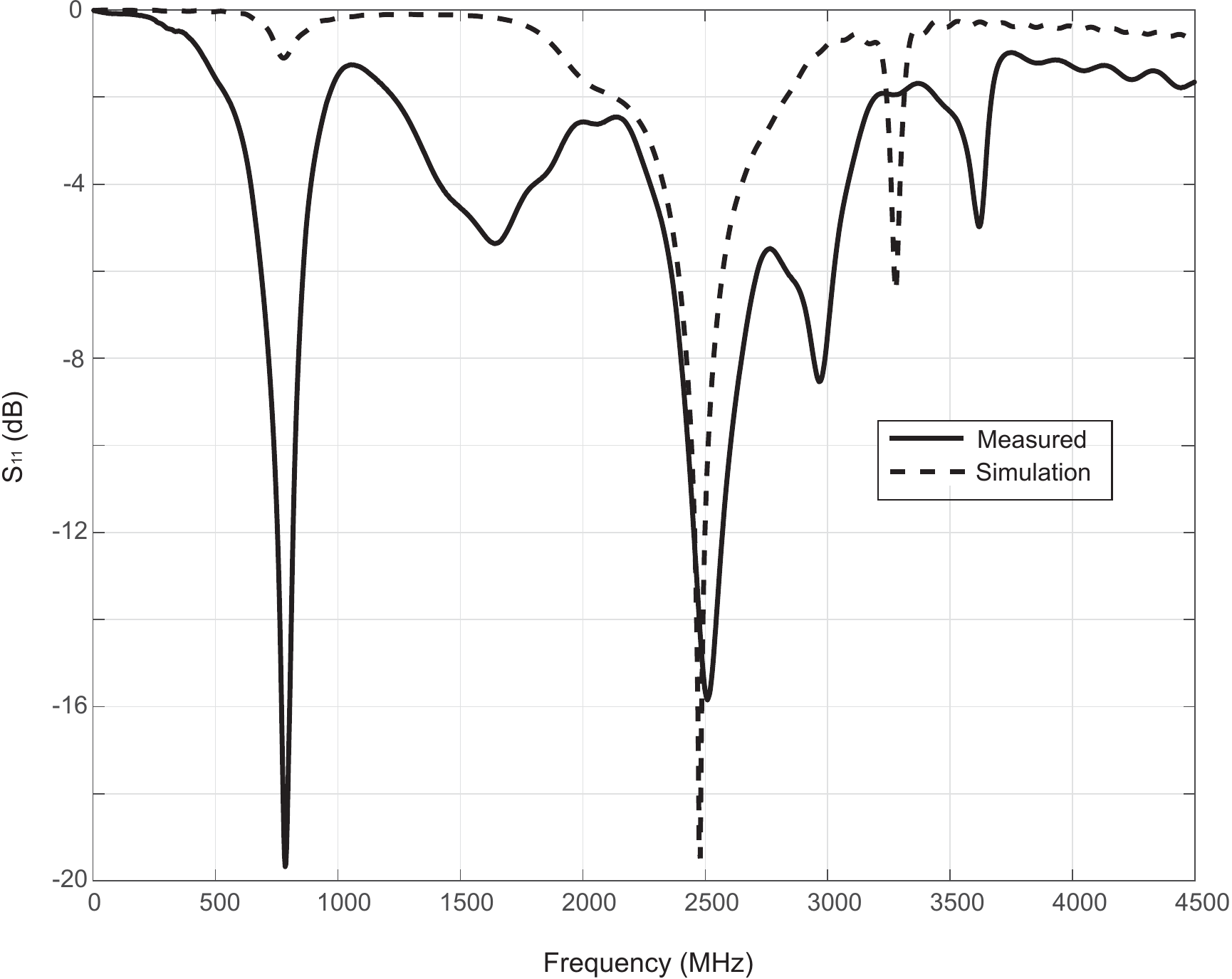}}
\subfigure[]{\includegraphics[width=0.4\textwidth]{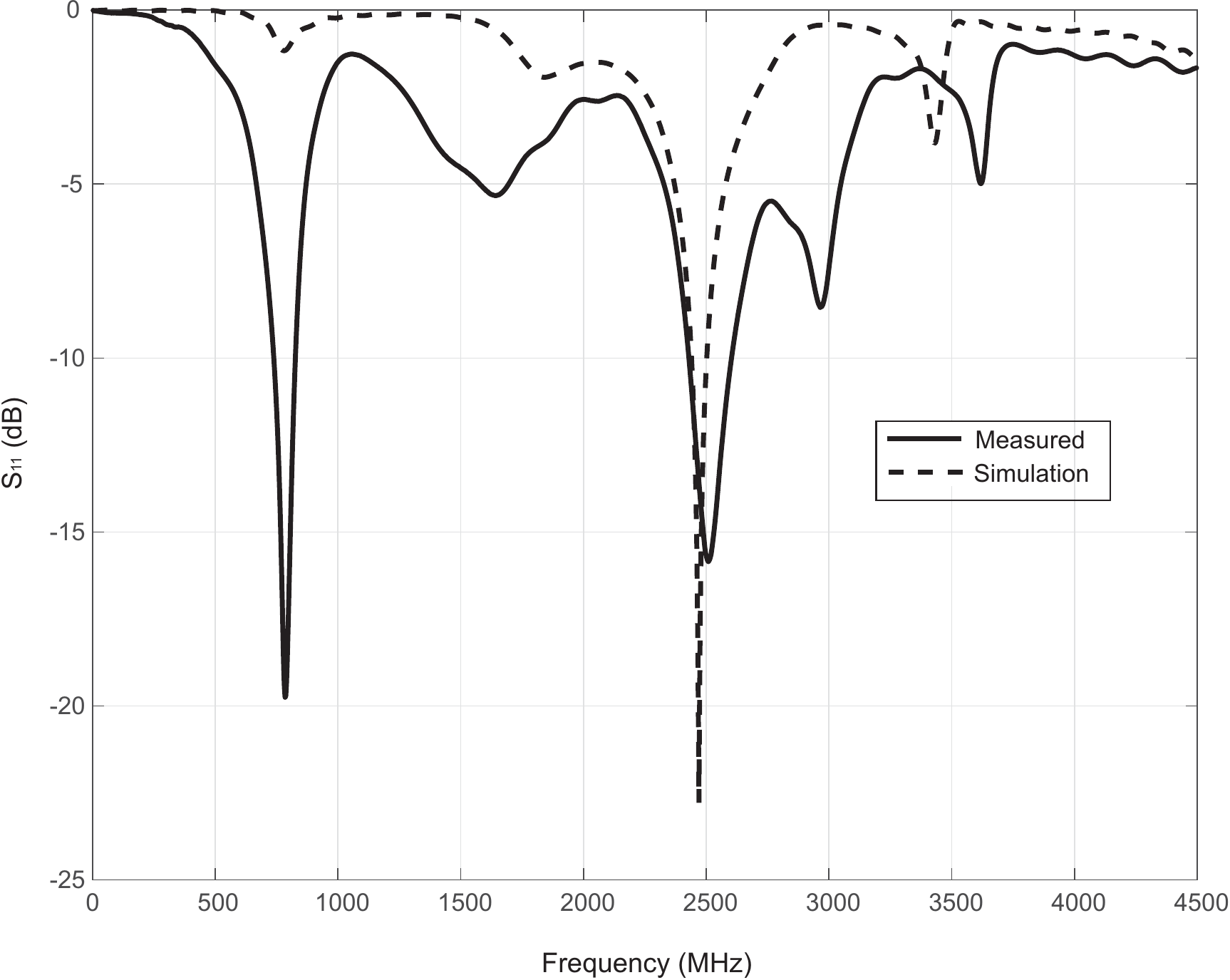}}
\end{center}
\caption{\label{fig:S11_med} Reflection coefficient simulated and measured for prototypes obtained with the CRO-4SL and CRO-5SL; (a) CRO-4SL prototype; (b) CRO-5SL prototype.}
\end{figure}

Measurements of $S_{11}$ for both prototypes have been carried out, and they have been compared with the simulation results, see figures \ref{fig:S11_med} (a) and (b). As figures show, the behaviour of the antenna at the resonant frequency is very good and in both cases the bandwidth requirements are satisfied. However, note that the prototype antenna obtained with the CRO-5SL is better than its counterpart obtained with the CRO-4SL, since the former exhibits a wider bandwidth, and the values of the $S_{11}$ are lower than for the antenna optimized with the CRO-4SL algorithm.

\begin{figure}[!ht]
\begin{center}
\subfigure[]{\includegraphics[width=0.4\textwidth]{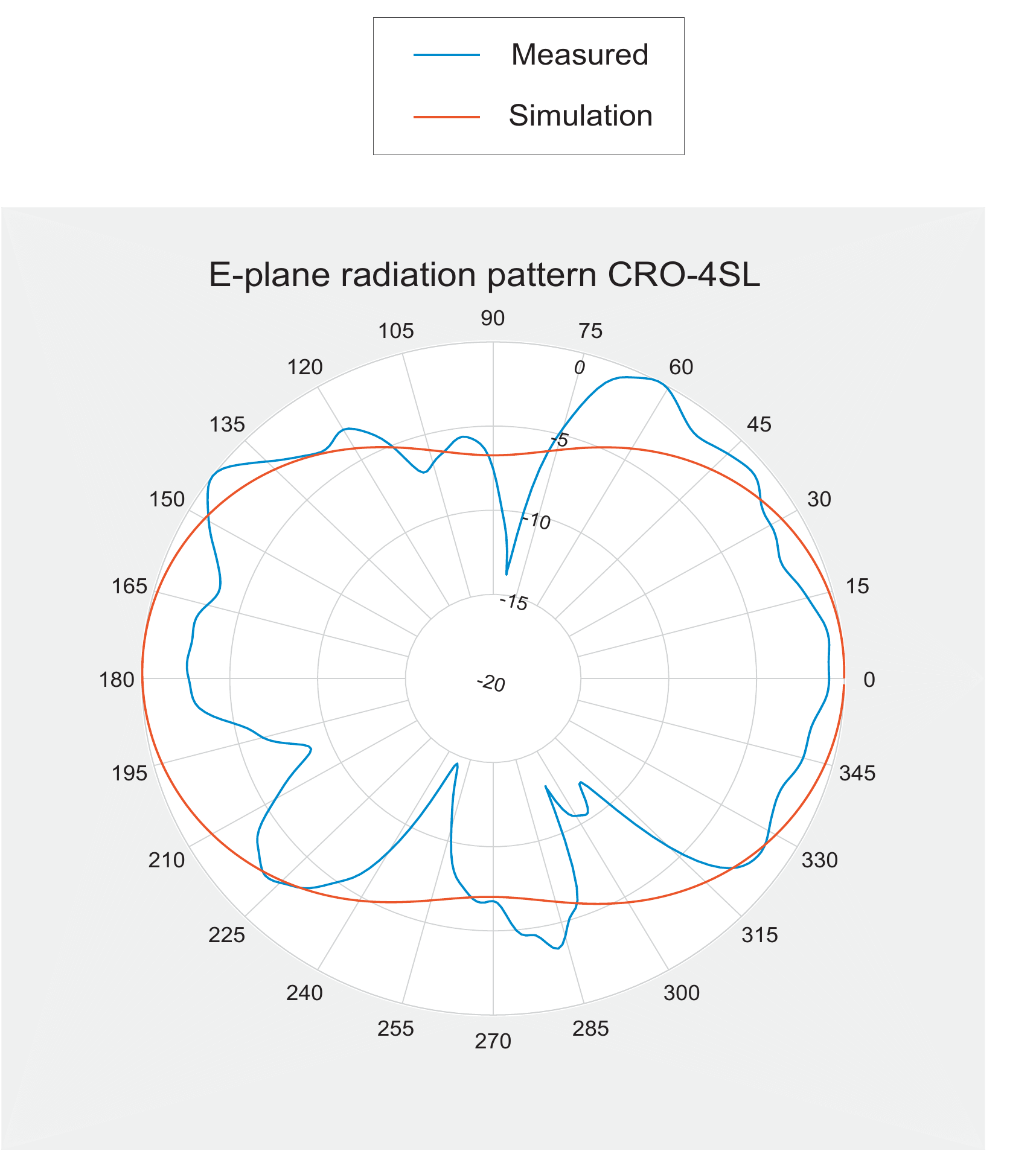}}
\subfigure[]{\includegraphics[width=0.4\textwidth]{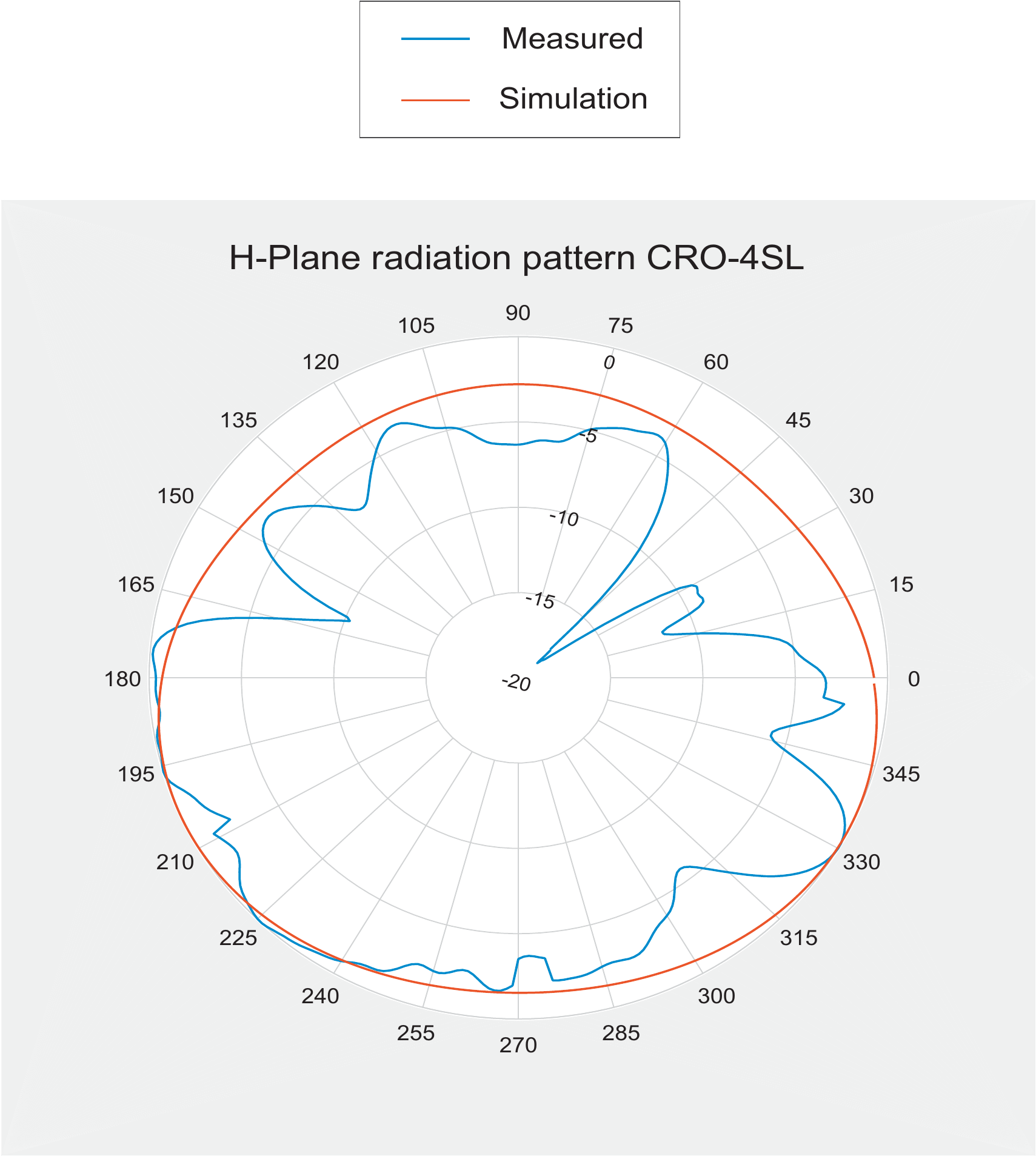}}
\end{center}
\caption{\label{fig:EHplane4} E plane and H plane H radiation patterns for the antenna optimized with the CRO-4SL algorithm; (a) E plane; (b) H plane.}
\end{figure}

\begin{figure}[!ht]
\begin{center}
\subfigure[]{\includegraphics[width=0.4\textwidth]{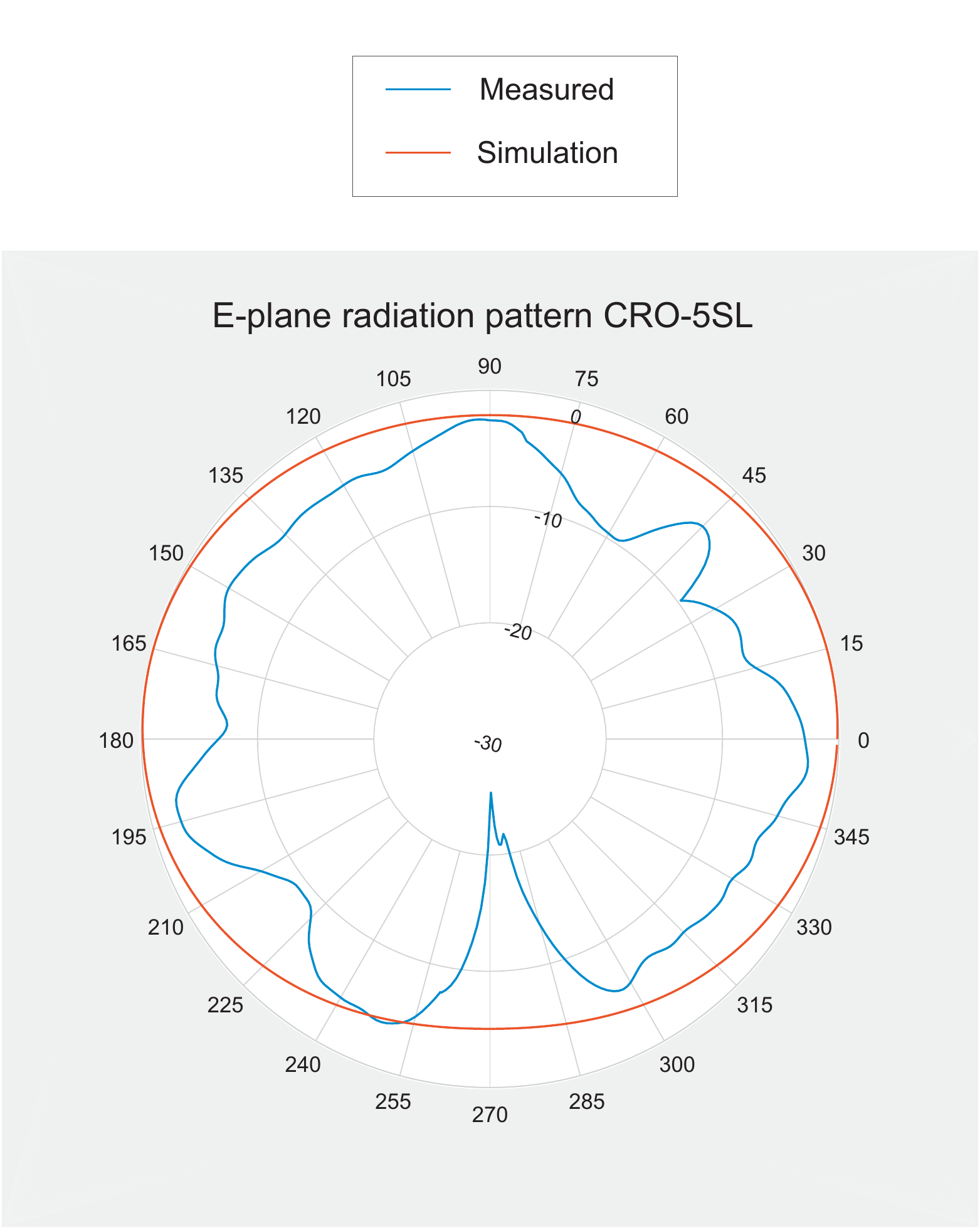}}
\subfigure[]{\includegraphics[width=0.4\textwidth]{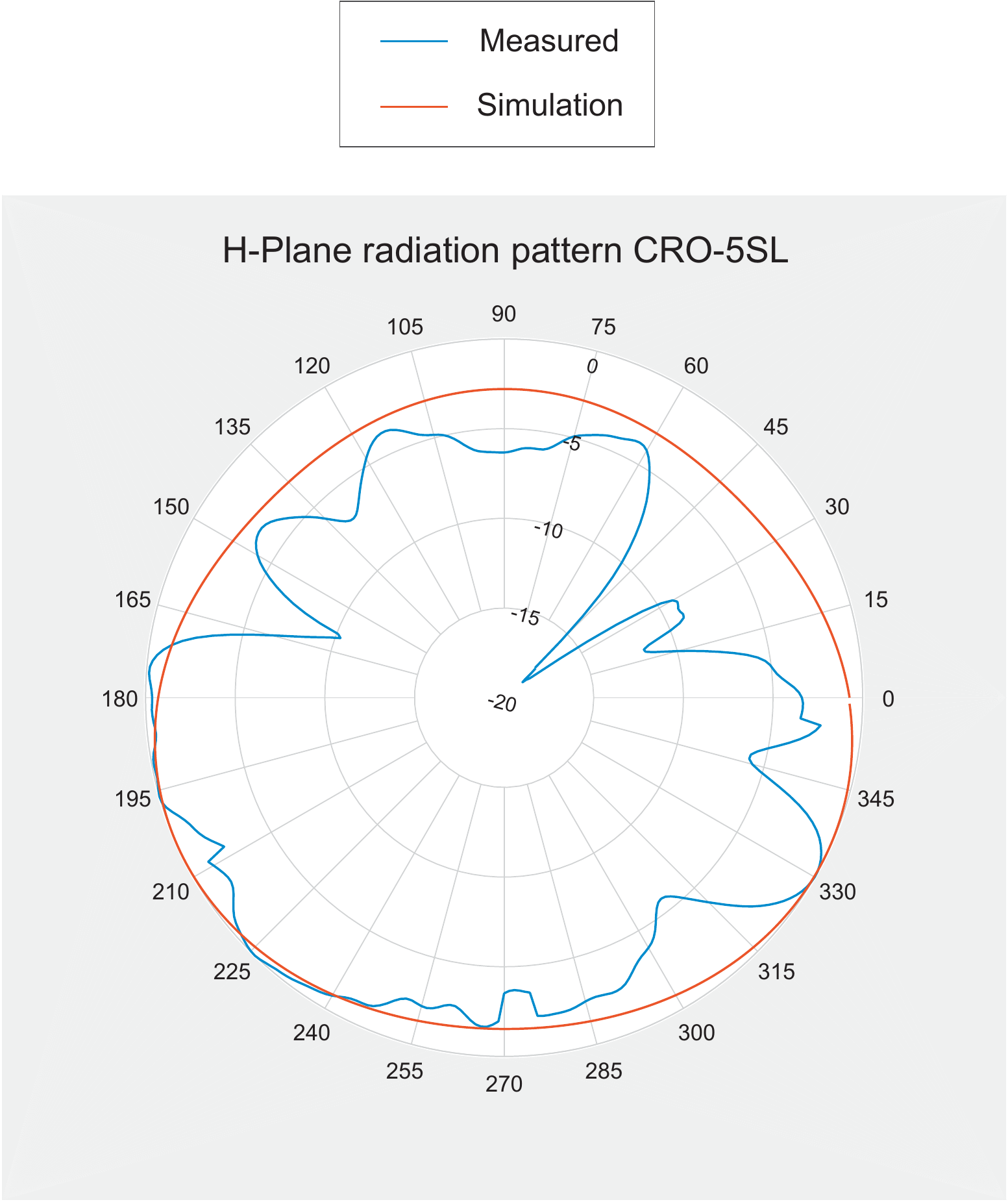}}
\end{center}
\caption{\label{fig:EHplane5} E plane and H plane radiation patterns for the antenna optimized with the CRO-5SL algorithm; (a) E plane; (b) H plane.}
\end{figure}

In addition to the reflection coefficient results, the radiation patterns of the constructed antennas have been analyzed and compared to those obtained by simulation. Figures from \ref{fig:EHplane4} and \ref{fig:EHplane5} show these results.


Regarding the simulation of the antennas' gain values, in both cases they are very similar, around 3.143 dBi. Note that when both antennas were measured, these values resulted lower than in the simulations, around 2.9 dBi. However, note that according to RFID specifications, these gains are still enough to use these devices as RFID antennas in the 2.45 GHz band. According to all these results, reflection coefficient, radiation pattern and gain values, the prototype designed and optimized using the CRO-5SL algorithm presents the best performance at 2.45 GHz for RFID applications. Moreover, the textile material which have been used in the antenna manufacturing makes easy its embedding into clothes, with multiple real applications in location science.

\subsection{A note on the CRO-SL algorithm performance}

The performance of the proposed CRO-SL in this optimization problem of textile meander-line IFA design can be further analyzed by comparing its results with that of alternative state-of-the-art meta-heuristics. Specifically, Table \ref{CROresults} shows the best results in terms of the objective function, given by Equation \ref{fitness}, obtained by the CRO-5SL (best CRO-SL version tested), with that of a HS approach, a DE algorithm, a GA with 2Px crossover and an ES with GM search procedure.

\begin{table}[!ht]
\caption{\small{Comparative results of the CRO-5SL and alternative meta-heuristic performance (in terms of the objective function given by Equation \ref{fitness}).}}
\label{CROresults}
\begin{center}
\begin{tabular}{|c|c|}
\hline
Algorithm & Best Fitness \\
\hline
\hline
CRO-5SL& 39.9009 \\
\hline
HS & 9.4808 \\
\hline
GA (2Px crossover) & 9.4376 \\
\hline
ES (GM Mutation) & 11.0827 \\
\hline
DE & 9.8170 \\
\hline
\end{tabular}
\end{center}
\end{table}

\begin{figure}[!ht]
	\centering
	\includegraphics[width=.5\columnwidth]{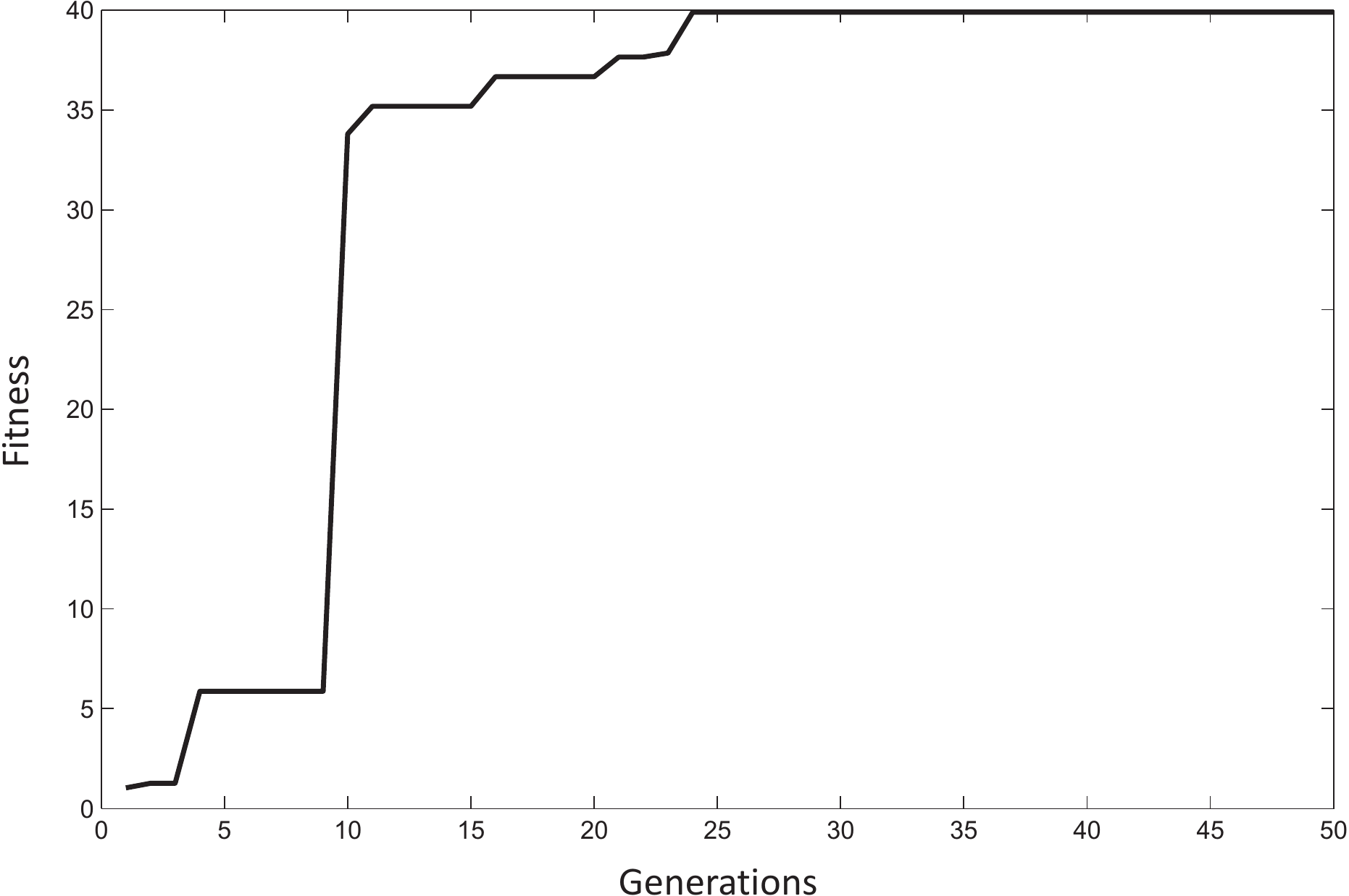}
	\caption{Best coral fitness evolution in the CRO-5SL algorithm.}
	\label{EvoMaxCRO-5SL}
\end{figure}

\begin{figure}[!ht]
\begin{center}
\subfigure[]{\includegraphics[width=0.5\textwidth]{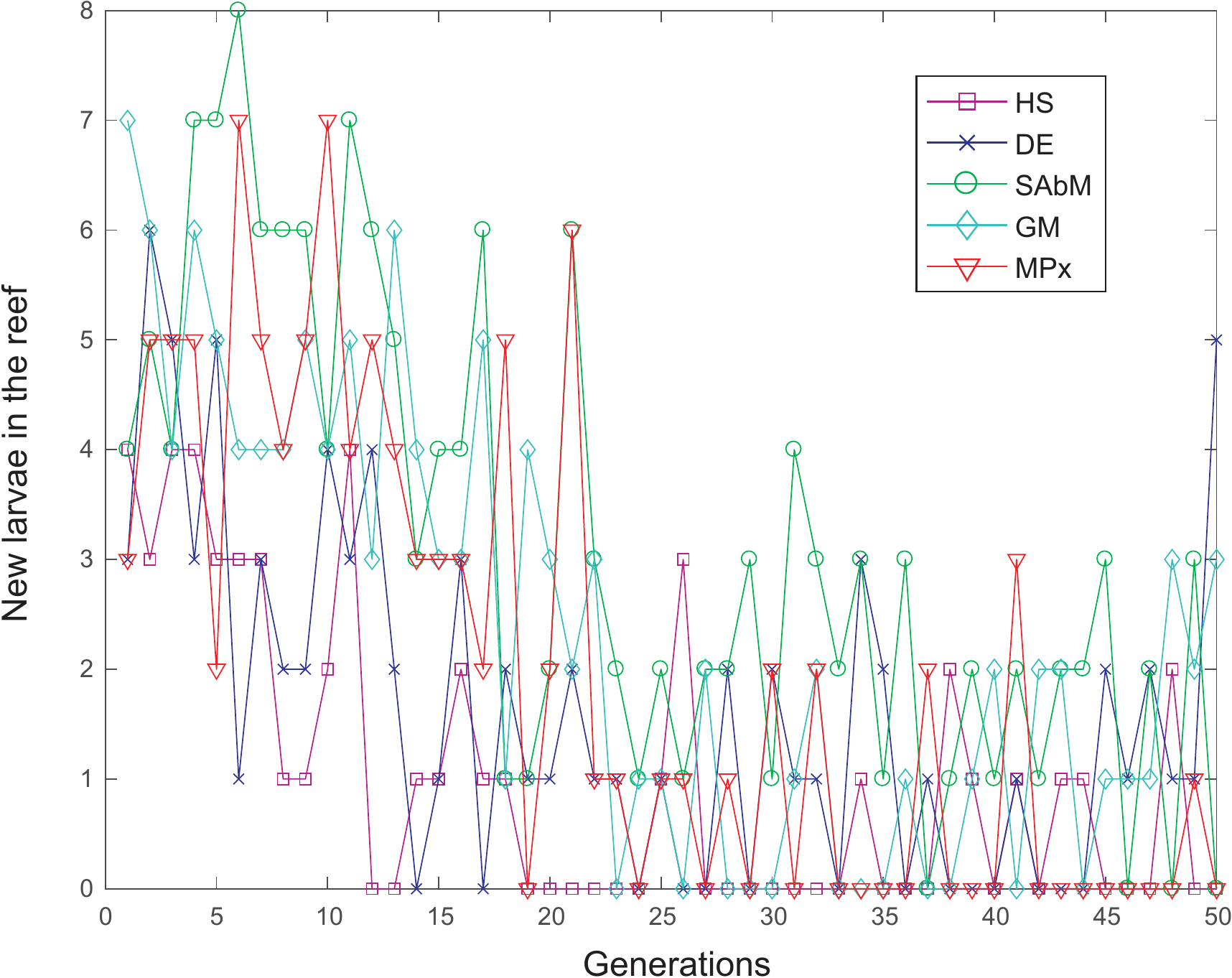}}
\subfigure[]{\includegraphics[width=0.5\textwidth]{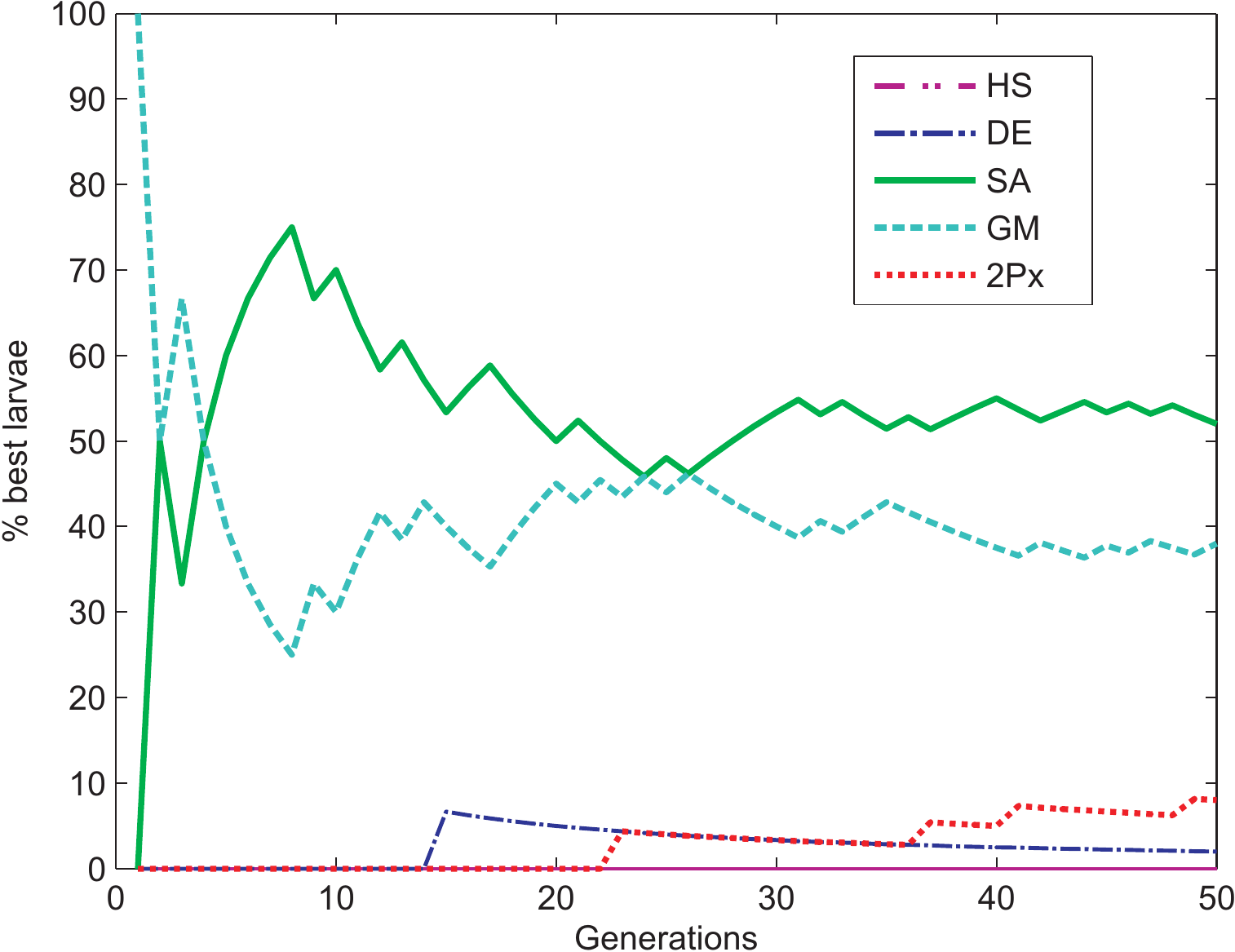}}
\end{center}
\caption{Evolution of the number of new larvae which are able to get into the reef per generation and substrate, and percentage of best larvae obtained from each substrate, in the best run of the CRO-5SL algorithm; (a) Number of new larvae in the reef; (b) Percentage of best larvae formed.}
\label{Substrate_performance}
\end{figure}

Figure \ref{EvoMaxCRO-5SL} shows the fitness evolution of the CRO-5SL when the best antenna prototype was obtained. As can be seen, the CRO-5SL approach is able to quickly get to a high-quality solution, obtaining the algorithm's convergence in less than 25 generations. Figures \ref{Substrate_performance} (a) and (b) illustrate an analysis of the CRO-5SL performance in terms of its different substrates. Figure \ref{Substrate_performance} (a) shows the evolution of the number of new larvae into the reef per generation and substrate. This figure can assist us to evaluate which substrates are able to get more corals (solutions) into the reef. As can be seen, the SA, GM and 2Px operators are the ones which attach a larger number of corals in the reef in each generation. It seems that the other operators are not able to attach so many larvae during the evolution. This feeling is confirmed in Figure \ref{Substrate_performance} (b), where the percentage of best larvae formed during the evolution per substrate is displayed. As can be seen, the SA and GM substrates are able to generate the best larvae (solutions) consistently during the CRO-SL evolution. The 2Px and DE substrates obtain the best larvae in a small percentage of the generations, whereas it seems that the HS is not able to obtain the best larvae at any time. Note, however, that the HS is able to attach some solutions to the reef (as shown in Figure \ref{Substrate_performance} (a), and contributes this way to the CRO-SL good performance. It is specially interesting the good performance of the SA operator in this problem. The SA operator, first introduced in \cite{Salcedo16b}, is able to introduce a kind of highly non-linear search pattern, with fractal structure, which seems to be very effective in obtaining good solutions in the optimization problem at hand.

\section{Conclusions}\label{sec:conclusions}

The CRO-SL algorithm has been applied in this chapter to adjust the resonant frequency and the bandwidth of the wearable antenna for RFID services at 2.45 GHz. The structure of the proposed device is based on an Inverted-F antenna including a meander shape with variable width and spacing, manufactured with a textile materials (FELT and flexible copper). The CRO-SL algorithm has been used to establish the optimal antenna and meander dimensions, which optimized the characteristics of resonant frequency and bandwidth of the antenna. The algorithm has been implemented in Matlab and antenna simulations during the evolution have been carried out using the CST Microwave Studio software, which allows a complete simulation of tentative antenna prototypes, and the fast calculation of the objective function. We have also shown the goodness of the proposed CRO-SL approach by comparing its performance with that of alternative state-of-the-art meta-heuristics, obtaining better results in all cases. We have also shown that the proposed automatic optimization process proposed is able to obtain a high quality RFID antenna, with good properties of bandwidth and resonant frequency, by constructing the best devices obtained from the CRO-SL evolution. They have shown that the constructed antennas fully agreed with the simulations carried out in their design, and moreover, the final prototypes constructed presented reduced dimensions and light weight, which are important characteristics of an RFID antenna to be embedded into protective clothing for different applications.

\part{Conclusions and future research activities}\label{part:conclusiones}
\chapter*{Final Remarks}
As a consequence of the research activity carried out in this Thesis, several results have emerged as well as relevant considerations. The existing results, those of special significance, are outlined below:
\begin{itemize}
\item In this Thesis the CRO-SL algorithm performance in several engineering optimization problems has been analyzed. The problems tackled are characterized by being difficult to address analytically and computationally heavy.
\item In Chapter \ref{cap:scheduling} the algorithm has been successfully applied to the battery scheduling optimization problem, getting better results for every week in the four seasons carried out than the deterministic or expert solution.
\item In Chapter \ref{cap:tmds} the CRO-SL algorithm is able address a problem which till now, it has been performed by hand due to the computational cost.
\item It is also remarkable that in this kind of problem the expert determines the TMD locations which is commonly all in the last floor. However the solution provided by the algorithm challenge the expert knowledge, since it says that must be one by floor.
\item In Chapter \ref{cap:avcs} it is posed the problem of human-induced vibrations via active controllers, in which the positioning and the control gains of the AVC's must be optimize. The CRO-SL shows a high performance beating the algorithm recently used, the ECBO.
\item In Chapter \ref{cap:antena}, it is showed the application of the CRO-SL to a design problem of a textile RFID antenna. The fitness function used in this problem takes information provided by CST software, which takes a long time to perform the antenna simulation. In spite of the low number of iterations the algrithm proposed attain a great solution, fulfilling widely the requirements. Furthermore, it has been experimentally proved that the antenna is able to work in two bands of frequency.
\item The CRO-SL algorithm can be used for give information about which kind of search procedures is acting better. It has been experimentally proved that removing the weaker substrates does not guarantees a higher performance. In many cases, as it showed in this thesis, all substrates by separate works worse than together. \textit{Unity makes strength}.
\end{itemize}

\chapter*{Future research lines}
Despite the promising results which have been obtained this research work, there are still different directions in which subsequent studies could be conducted. Some of these aspects that could be addressed more deeply in the near future are the following:
\begin{itemize}
\item This thesis focuses on the optimization of several mono-objective optimization problems. One of the main extensions of this research would be to adapt the CRO-SL algorithm to work in multi-objective problems. For this, the way in which the corals and the larvae compete for a position on the coral must be redefined and a crowding distance must be defined. The result would be the obtention of a pareto front of the individuals that gives compromise solutions. A first version of Multi-Objective CRO is given in \cite{SalcedoMO15}.
\item Nowadays it is common that new algorithms based on the mutation and evolution of a population appear. So, it would be very interesting for the developing of a new research the addition of substrates based on emerging algorithms into the CRO-SL, so they can be compared with the classical ones in different applications.
\item One of the main drawbacks of the CRO-SL algorithm is that it depends on many parameters. The user is who determines their values and, depending on them, the performance of the algorithm could be different. This is the reason why another extension of this research should be the auto-tune of the parameters. With this modification, the algorithm should be able to increase the number or the ratio of coral depredated if the population are stagnated. A first attempt in the classical CRO was recently carried out in \cite{Duran18}.
\item In this work it has been shown that some of the substrates do not obtain good individuals, however, they help maintain the reef diversity. Following this idea, the algorithm could dispose of a large number os substrates but they would be only used when they were needed, for example in cases of low diversity, etc.
\end{itemize}

\part{Appendix}
\chapter*{List of publications}

This section constitutes a compendium of the different scientific publications produced as a result of the research work developed in connection with this work. In addition to those, other studies carried out during the training process of this Ph.D. were developed in parallel.

\section*{Papers directly to the Ph.D. Thesis.}
\subsection*{Papers in International Journals}

\begin{enumerate}
\item C. Camacho-Gómez, X. Wang, E. Pereira, IM. Díaz, S. Salcedo-Sanz, ``Active vibration control design using the Coral Reefs Optimization with Substrate Layer algorithm,'' \em{Engineering Structures}, vol. 157, pp. 14-26,  2018, (JCR 2016: 2.258).
\item S. Salcedo-Sanz, C. Camacho-Gómez, R. Mallol-Poyato, S. Jiménez-Fernández, J. Del Ser, ``A novel Coral Reefs Optimization algorithm with substrate layers for optimal battery scheduling optimization in micro-grids,'' \em{Soft Computing}, vol. 20, no. 11, pp. 4287-4300, 2016, (JCR 2016: 2.472).
\item S. Salcedo-Sanz, C. Camacho-Gómez, A. Magdaleno, E. Pereira, A. Lorenzana, ``Structures vibration control via Tuned Mass Dampers using a co-evolution Coral Reefs Optimization algorithm,'' \em{Journal of Sound and Vibration}, vol. 393, pp. 62-75, 2017, (JCR 2016: 2.593).
\item S. Salcedo-Sanz, A. Aybar-Ruíz, C. Camacho-Gómez, E. Pereira, ``Efficient fractal-based mutation in evolutionary algorithms from iterated function systems,'' \em{Communications in Nonlinear Science and Numerical Simulation}, vol. 56, pp. 434-446, 2018, (JCR 2016: 2.784).
\item S. Salcedo-Sanz, C. Camacho-Gómez, L. Carro-Calvo, F. Jaume-Santero, E. Alexandre-Cortizo, ``Near-Optimal Selection of Representative Measuring Points for Robust Temperature Field Reconstruction with the CRO-SL and Analogue Methods,'' \em{International Journal of Climatology} accepted with minor revision(JCR 2016: 3.76).
\item C. Camacho-Gómez, I. Marsá-Maestre, J. M. Giménez-Guzmán, S. Salcedo-Sanz, ``A Coral Reefs Optimization Algorithm with Substrate Layer for Robust Wi-Fi Channel Assignment,'' \em{Applied Soft Computing}, under review, (JCR 2016: 3.541).
\item R. Sánchez-Montero, C. Camacho-Gómez, P. López-Espí, S. López-Ruiz, S. Salcedo-Sanz, ``Optimal Design of a Planar Textile Antenna for RFID Systems with the CRO-SL Algorithm,'' \em{IEEE Transaction on Antennas and Propagation}, under review, (JCR 2016: 2.957).
\item S. Salcedo-Sanz, C. Camacho-Gómez, A. Aybar-Ruiz, E. Alexandre-Cortizo, ``Wind Power Field Reconstruction from a Reduced Set of Representative Measuring Points,'' \em{Applied Energy}, under review, (JCR 2016: 7.182).
\end{enumerate} 

\subsection*{Papers in International Conferences}
\begin{enumerate}
\item S. Salcedo-Sanz, C. Camacho-Gómez, D. Molina, F. Herrera, ``A coral reefs optimization algorithm with substrate layers and local search for large scale global optimization,'' in \em{IEEE Congress on Evolutionary Computation (CEC)} 3574-3581, 2016.
\end{enumerate}

\section*{Other results achieved during the Ph.D. training process.}
\subsection*{Papers in International Journals}

\begin{enumerate}
\item S. Salcedo-Sanz, R. Deo, L. Cornejo-Bueno, C. Camacho-Gómez, S. Ghimire, ``An efficient neuro-evolutionary hybrid modelling mechanism for the estimation of daily global solar radiation in the Sunshine State of Australia,'' \em{Applied Energy}, vol. 209, pp. 79-94, 2018, (JCR 2016: 7.182).
\item C. Camacho-Gómez, S. Jiménez-Fernández, R. Mallol-Poyato, Javier Del Ser, S. Salcedo-Sanz, ``Optimal Design of Microgrid’s Network Topology and Location of the Distributed Renewable Energy Resources using the Harmony Search Algorithm,'' \em{Soft Computing}, $1^{st}$ revision submitted to the journal, (JCR 2016: 2.472).
\end{enumerate}

\subsection*{Papers in International Conferences}

\begin{enumerate}
\item C. Camacho-Gómez, R. Mallol-Poyato, S. Jiménez-Fernández, ``Optimal placement of distributed generation in micro-grids with binary and integer-encoding evolutionary algorithms,'' \em{IEEE Congress on Evolutionary Computation (CEC)}, 3630-3637, 2016.
\item L. Cornejo-Bueno, A. Aybar-Ruiz, C. Camacho-Gómez, L. Prieto, A. Barea-Ropero, S. Salcedo-Sanz, ``A Hybrid Neuro-Evolutionary Algorithm for Wind Power Ramp Events Detection,'' \em{International Work-Conference on Artificial Neural Networks} 745-756, 2017.
\end{enumerate}

\subsection*{Papers in National Conferences}
\begin{enumerate}
\item L. Cornejo-Bueno, C. Camacho-Gómez, A. Aybar-Ruiz, L. Prieto, S. Salcedo-Sanz, ``Feature Selection with a Grouping Genetic Algorithm–Extreme Learning Machine Approach for Wind Power Prediction,'' \em{Conference of the Spanish Association for Artificial Intelligence}, 373-382, 2016.
\end{enumerate}

\part*{References}\label{Refs}

%
%
%
%
%
%
%
%


\end{document}